# An LLM-Associated Register Shift in Korean Journal Abstracts

## A morphology-aware excess-vocabulary study of 398,296 abstracts, 2018 to August 2026, with an article-level comparison of the English abstracts of the same articles

Ahron Lee · *INTFRAME Research, Seoul, Republic of Korea* · ORCID 0009-0009-7780-086X

Working paper, version 8 · 4 September 2026 · DOI of version 8: 10.5281/zenodo.22303588 (all versions: 10.5281/zenodo.22102389) · Version history: Appendix J · Code and frequency tables: see Data availability

## Abstract

Excess vocabulary, a word's frequency above its pre-2023 trend, is how the change in scholarly English after 2022 has been measured. We adapt it to Korean with morphological units on 398,296 KCI abstracts (2018–August 2026), with 47,165 Vietnamese abstracts for comparison. Placebo floors are 0.1–2.2 points for the single-word statistic and at most 2.9 for the re-selected split-half set statistic. Korean abstracts show nothing in 2023, onset in late 2024, a rise through 2025 flattening in mid-2026: 시사하다 "suggest" appears in 21.4% of 2026 abstracts against 5.3% expected; plain verbs like 알아보다 "look into" fall to a quarter of trend. Under stated assumptions the single-word conditional lower bound on LLM-processed abstracts is 3.5%, 10.5% and 16.1% for 2024–2026 and a split-half set bound 7.8%, 20.6% and 33.0%. Holzwarth et al.'s estimator under the same discipline gives 41.9% and 72.1% for 2025–2026. Subject-matter controls reduce but do not remove it: restricting the set to lemmas three language-model annotators all call style leaves 14.7 of the 33.0 points, and pairing each 2026 abstract with its journal's closest base-period abstract leaves 34.1. Tested translation routes do not explain it: the surface marks of translated Korean fall as the markers rise. In the same articles' English abstracts the excess appears a year earlier; where the English side carries none, the Korean shift persists at 30 to 66% of the rate where it does. Control abstracts from three providers reproduce the rising words, with marker turnover consistent with model generations; implied prevalences are scenario-dependent.

## 1 Introduction

Since 2023, a growing body of work has measured how large language models (LLMs) changed the vocabulary of scholarly English. The clearest signal is lexical: a small set of words that ChatGPT-era models favour, such as *delve*, *underscore*, *intricate* and *showcase*, appeared in English abstracts far more often than any pre-2023 trend could explain. Kobak et al. formalised this as *excess vocabulary*: for each word, the frequency observed in a given year is compared with the frequency expected from a linear extrapolation of earlier years, and the surplus is attributed to LLM-assisted writing [1]. Applied to more than 15 million PubMed abstracts, the method produced a lower bound of 13.5% LLM-processed abstracts in 2024, and a full-text follow-up estimated that by late 2025 about 89% of biomedical papers carried LLM vocabulary somewhere in the manuscript [2]. Related distributional estimates for arXiv, Dimensions, Scopus, PubMed Central and MDPI point the same way [3–9]. A full-text study of 7.3 million articles from four publishers put the share showing LLM-associated language at 57% in 2025 against 12% in 2023, and found that adoption correlates with institutional and regional variables [19]. The inference from word frequencies to a prevalence figure has itself been contested [20].

Every one of these studies measures English. That is a natural place to start, since English is where the models are strongest and where the corpora are largest, but it leaves a basic question open. Do the same models leave the same kind of fingerprint when researchers write in their own languages? The one multilingual study we know of looked at news text in 34 languages and found convergent "emphasize"-type verbs in most of them, yet it also reported a *decrease* of AI-associated words in Korean news between 2020–21 and 2023–24, which the author attributed to corpus-composition effects and flagged for "dedicated follow-ups" [11]. Vietnamese was not included at all. For Korean scholarly writing one study exists: Koo, Kim and Kim applied an excess-vocabulary comparison to about 880,000 KCI abstracts in the humanities and social sciences from 2004 to 2024, setting the 2024 frequency of each word against a projection from 2021–2022, and found a sharp post-2023 rise of English style words (*additionally*, *underscoring*, *aligning*) in the English abstracts and only modest increases, in evaluative expressions, in the Korean ones [24]. Their series ends in 2024, the year in which, as we show below, the Korean signal had only begun; none of the studies we found covers 2025–2026, validates the Korean markers against model output, reports placebo levels or a prevalence bound, or includes Vietnamese. The English literature has meanwhile moved past the existence question: it has documented the lexical excess at the scale of whole databases [1, 2, 6], shown that LLM polishing homogenises style, amplifying a dominant register while suppressing individual variation [21], and begun to ask whether generative AI is pulling scientific English toward its U.S. variety, on 5.65 million Scopus articles [23]. What is not known is whether either phenomenon exists in a scholarly language that is structurally unlike English, and whether a measurement built on whitespace tokens can be carried to languages in which those tokens are not words. Those are the questions this paper takes up.

The gap matters for three reasons. First, policy: Korean and Vietnamese journals, funders and universities are writing rules about generative AI, and the only prevalence numbers they can cite are English ones. Second, linguistics: Korean is an agglutinative, verb-final language with an elaborate honorific and register system, and Vietnamese is an isolating, tonal language in which written spaces separate syllables rather than words. Neither can be analysed by counting whitespace tokens, so it is not obvious in advance that the English recipe transfers. Third, mechanism: if LLM style words in Korean turn out to be translations of the English ones (*emphasize*, *underscore*, *delve*), the phenomenon reflects a shared model prior; if they are language-specific formal-register items, it reflects something about how the models learned each language.

We adapt the excess-vocabulary method to Korean and Vietnamese and apply it to two large openly harvestable corpora of scholarly abstracts in these languages: 398,296 Korean-language abstracts from 2,282 journals indexed in the Korea Citation Index (KCI), covering 2018 through August 2026, and 48,357 Vietnamese abstracts from Vietnam Journals Online (VJOL), of which 47,165 fall in the analysed years 2019 through 2025. The Korean corpus is the main study. The Vietnamese corpus is about one eighth its size and lacks a usable negative control, so it serves only as an external comparison of the same words in a second language and is used for no bound. The contributions are the following.

- **Method transfer.** We replace whitespace tokens with morphological lemma–part-of-speech units (Korean) and segmented word–part-of-speech units (Vietnamese), fuse Sino-Korean noun + *hada* derivations into single verb lemmas, add adjacent-bigram and regular-expression constructions for multi-word style, and introduce a journal panel and a structured-abstract header normalisation that remove two composition artefacts we found to produce spurious 30-fold "excess" in a pre-ChatGPT control year.
- **Validation.** Placebo target years before ChatGPT, run at every extrapolation horizon the data allow (one to four years ahead, two to five base years), put the false-positive level of the statistic we use at 0.1 to 2.2 pp of excess, of the unrestricted maximum at 0.7 to 6.3 pp, always on a high-frequency unit, and of the split-half set statistic at −0.8 to +2.9 pp; we read only excess above the matched-horizon floor as signal, and a re-run on density-normalised frequencies gives the discount a conservative reader should apply (14% of the 2026 bound). A positive control (1,437 abstracts generated by three OpenAI models, one Anthropic model and one Korean open-weight model in sixteen prompt conditions) shows that once output length is matched to real abstracts, many of the highest-excess markers are reproduced at rates comparable to or above their real 2026 rates (of 27 rising units, nine above, fifteen overlapping and three below, 구조적 the clear exception), and most of the words that fell are barely produced at all; the Korean model, released in late 2024, writes the marker set that peaked in 2024, which is consistent with a turnover by model generation rather than by any one provider (Section 6.2). Measured on the same control corpus, the sensitivity of the set indicator puts the 2026 share at 77–97% in the scenario in which abstracts were processed as our matched-length drafting prompts process them (propagated ranges from 58% to above 100%), at 66–90% on the two headline words if they were rewritten from a draft as the two 2026 API models rewrite them, and shows that polishing of the kind our editing prompt elicits cannot produce the observed vocabulary at any prevalence (Section 5.10).
- **Style or topic.** A three-annotator classification of the marker lemmas, by language models working independently of one another, and a topic-matched comparison within journals test whether the register change is instead a change in what the journals study (Section 5.13); a scan for the surface marks of translated Korean, and a model translation condition, test whether it can be explained by the translation routes examined here (Section 5.14).

- **Within-article comparison.** The English abstracts of the same articles give an article-fixed comparison across languages (Section 5.12): the English excess appears in 2023 and the Korean in late 2024; the fixed English markers of the 2024 model generation recede in 2026 as their Korean counterparts do; and the Korean shift is not confined to articles whose English abstract shows model vocabulary.
- **Robustness.** The result survives equal weighting by journal, a strict panel of journals present in every year, a split of 2026 by half-year, and a selection-aware bootstrap of the bound; and within single 2026 abstracts the rising and falling vocabularies became strongly negatively associated (7.3% of the abstracts that contain a rising marker also contain 살펴보다, against 12.7% of other abstracts, $z = -12.5$), a pattern consistent with a document-level register shift, by a model or by an author who has adopted the register, and harder to obtain from word-by-word diffusion (Section 5.8).
- **Findings.** Korean scholarly writing shows an LLM signature that begins in the second half of 2024, one year after English, rises steeply through 2025 and flattens in the second quarter of 2026 at four times its pre-2023 level. Several style verbs and adjectives exceed their trend by four to eleven times; the verb *sisahada* (시사하다, "suggest, imply") alone appears in 21% of 2026 abstracts against an expected 5.3%. Plain-register expressions such as *araboda* (알아보다, "look into") and *doumi doeda* (도움이 되다, "be helpful") fell to a quarter of their trend. The single-word lower bound on LLM-processed Korean abstracts (units with excess ratio at least 1.5) is at the placebo floor in 2023 and rises to 3.5% in 2024, 10.5% in 2025 and 16.1% in 2026 (January to August). A set-based bound of the kind reported for English, with the set of excess words chosen on one half of the journals and measured on the other, is 7.8% in 2024, 20.6% in 2025 and 33.0% in 2026 (95% CI 30.8 to 35.4). The Korean 2026 floor is above the English 2024 floor, but floors do not order the quantities they bound, and the two studies search sets of different sizes (Section 5.9). The marker set itself turns over between 2025 and 2026 in step with model generations. Both bounds survive placebo target years at every extrapolation horizon the pre-LLM data allow, alternative expectation models, removal of the morphological fusion rule that generates eleven of the thirty largest excesses, and refitting without the thinnest base year; the rising words rose in about nine of every ten of the journals with enough abstracts to tell (Section 5.8). An exploratory Vietnamese comparison shows the same family of words (*nhấn mạnh* "emphasize", *then chốt* "key", *cơ chế* "mechanism") a year later and at lower amplitude. The English abstracts of the same KCI articles (328,358 of the 337,151 panel abstracts have one) move a year earlier than their Korean abstracts, and an article whose English abstract carries the period's English markers is 2.3 times as likely, in odds, to carry the Korean ones in 2026 (baseline 1.6 to 1.9 before 2023), while the Korean shift is also present where the English side carries none, at 70% of the rate among articles whose English side does carry them, and at 30 to 66% of that rate after the English indicator's own sensitivity is corrected (Section 5.12).
- **Resources.** We release the full per-year document-frequency tables for both languages, the analysis code, and a public "Korean AI style dictionary" of the most affected expressions with their trajectories, so that editors, teachers and other researchers can reuse the measurements without re-harvesting the corpora.

## 2 Related work

### 2.1 Corpus-level measurement of LLM-assisted writing

Two families of methods exist. Per-document detectors classify individual texts as human or machine written; they are known to be unreliable on edited or paraphrased text and to be biased against non-native writers of English [17], which makes them a poor instrument for exactly the population studied here. For Korean specifically, Park et al. built a human-versus-LLM benchmark and detector, one of whose three genres is paper abstracts, but a detector trained on a constructed benchmark answers a different question from the one asked here [18]. Corpus-level methods instead estimate the *share* of a collection touched by LLMs from distributional shifts, without labelling any document. Liang et al. fitted a mixture of human and LLM word distributions to arXiv, bioRxiv and Nature-portfolio abstracts and estimated up to 17.8% LLM-modified computer-science abstracts by early 2024 [3]. Geng and Trotta tracked ChatGPT-preferred words in one million arXiv abstracts and estimated about 35% LLM-style abstracts in computer science, a figure that is specific to that field and conditional on taking one model's response to one revision prompt as the reference [4]. In a follow-up they showed that a marker can fall once it becomes publicly known as a marker: the frequency of *delve* dropped after it was widely discussed, while less-discussed markers kept rising [5]. Kobak et al. introduced the excess-vocabulary design used here, which needs no LLM reference distribution at all: it only requires a pre-LLM series to extrapolate from, and it yields a lower bound on prevalence from the excess of a single word or of a set of words, because every surplus occurrence must sit in some processed abstract [1]. Kousha and Thelwall, Thelwall and Kousha, Gray, and Bao et al. applied word-tracking designs to Scopus, Web of Science, PubMed Central, Dimensions, MDPI and arXiv, and consistently found field differences and rapid growth through 2024–2025 [6–9]. Topaz and Bahl object that a lexical shift is not a calibrated measure of prevalence and should not carry the determinants read off it, a caution we take up in Section 5.9 by measuring the statistic's behaviour on pre-LLM years [20]. Holzwarth, González-Márquez and Kobak extended the analysis to 1.19 million full texts and reported that 89% of December 2025 biomedical papers show LLM vocabulary, with usage highest in Discussion sections [2].

### 2.2 Why models prefer particular words

Juzek and Ward examined the sources of lexical over-representation in LLM output and concluded that preference-tuning (RLHF) rather than pre-training data is the most likely origin of words such as *delve* and *explore*, some of which are three to five times more frequent in model output than in human text [10]. This matters for the cross-lingual question: if the preference is induced at the tuning stage, and tuning data are largely English, one would expect the preferred words in other languages to be translations of the English favourites, or formal-register near-equivalents produced when the model writes "carefully".

### 2.3 Beyond English

Juzek compared GPT-4.1 continuations with human text in 34 languages drawn from the WMT news crawl and found cross-lingual convergence on a small set of concepts (an "emphasize"-type verb appears among AI-overused lemmas in 24 of 34 languages), together with diachronic increases in 26 languages [11]. Korean was one of the six languages whose AI-associated words fell relative to baseline words (eight fell in absolute terms), which the paper attributed to corpus composition and flagged for follow-up. Vietnamese was not covered. For Korean scholarly writing there is one published excess-vocabulary study, discussed in the next paragraph; the other Korean work we found is either small-sample or about a different register (student writing, online discourse about generative AI) and is not cited here. For Vietnamese scholarly writing we are not aware of any published excess-vocabulary or distributional study. A parallel line of work places these shifts in a homogenisation frame, reporting a narrowing of linguistic variety in LLM-mediated text [21].

Closest to this paper is the KCI study of Koo, Kim and Kim [24], which we read in full. It covers 882,261 abstracts of KCI humanities and social-science journals from 2004 to 2024, English and Korean, computes document frequencies by year, and takes as the expected 2024 frequency of each word a straight line through its 2021 and 2022 values; excess is reported as a difference and as a ratio, and diffusion as the mean and median number of excess words per abstract. Its Korean units are Kiwi morphemes re-merged into inflected forms, so that 탐구하며, 탐구한다 and 탐구할 are separate units. It finds sharp rises in the English abstracts (*underscores* at 124 times its expectation, *additionally*, *aligning*; *these* from 22.7% to 35.8% of abstracts) and modest ones in the Korean abstracts (top ratios of 4.6 in the social sciences and 8.1 in the humanities, on rare forms; per-abstract means of excess words moving from about 3 to 4.5 with unchanged medians), and offers three explanations: that English abstracts are produced through model translation, that Korean abstracts are written or edited by the authors themselves, and that Korean inflection disperses one expression over many forms. Relative to it, this paper extends the series through August 2026, the years in which the Korean signal actually developed; works on lemma–tag units with suffix fusion, so that 시사하다 and all its inflections are one unit, which removes the dispersion their third explanation invokes; fits five base years and measures the placebo level of the statistic at every extrapolation horizon; chooses marker words on one half of the journals and measures them on the other, with journal-cluster resampling; validates the markers against 1,437 model-written abstracts from three providers and reads the turnover of markers across model generations; converts the excess into a conditional lower bound and prompt- and model-conditioned scenario shares; and adds Vietnamese as an external comparison.

### 2.4 Diffusion into human language

Yakura et al. showed with a synthetic-control design over hundreds of thousands of hours of podcasts and academic talks that ChatGPT-preferred words such as *delve* rose abruptly in spontaneous English speech after the model's release [12]. Excess vocabulary in written abstracts therefore mixes two processes: text that was drafted or edited by a model, and text written by people whose vocabulary has shifted. The lower bound in Section 4.7 does not depend on separating them; Section 6 discusses how the two can be told apart.

## 3 Data

### 3.1 Korean: KCI abstracts

The Korea Citation Index, operated by the National Research Foundation of Korea, exposes article metadata through an OAI-PMH endpoint [15]. Its native `oai_kci` format carries the original-language abstract, the publication year and month, the journal name and the article language. We harvested records by article identifier range (the resumption token encodes the next identifier, which allows deterministic keyset paging at 100 records per request) and kept every record whose original-language abstract contains Hangul and is at least 80 characters long. The corpus is the union of two harvests: a first walk whose identifier ranges under-covered 2018 and 2021, and a completing walk over the missing ranges made the following day. Together they comprise 398,296 unique article records, de-duplicated by article identifier, published from 2018 to August 2026 in 2,282 journals (432,545 harvested rows before de-duplication) (Table 1), and every number in this paper is computed on this completed corpus. Year sizes are still uneven because KCI indexes years at different depths, so the analysis caps each year at a common size and never compares raw counts across years; Appendix H reports the first harvest alone and shows what completing it changed. The article language recorded by KCI is Korean for 99.4% of the records; the rest are articles in English, Chinese or Japanese that carry a Korean-language abstract, which is the text analysed. 102 records (0.03%) share their abstract text verbatim with another record and were left in place.

The source carries two biases. KCI lists journals that passed the Foundation's accreditation, so the corpus over-represents established Korean-language humanities, social-science, education and applied-science journals and under-represents science and engineering journals that publish English-only abstracts. And the abstract is the most polished part of a paper, the part most likely to be run through a model, so prevalence measured on abstracts is not prevalence in full texts.

**English abstracts of the same articles.** KCI records carry the English abstract of each article alongside the Korean one. For the comparison of Section 5.12 we re-walked the identifier ranges of the harvest and stored the English abstract of every record in the harvest store: 418,271 of its 428,131 records (98%) have one of at least 80 Latin characters (79% for 2026, whose records are still being completed). The store is larger than the corpus of Table 1 because it also holds records from 2017 and 1,072 stray records from 2015–2016 caught by the identifier walk, all outside the analysed years; the comparison of Section 5.12 uses the 337,151 panel abstracts of Table 1, of which 328,358 (97%) have an English abstract. English text is tokenised into lower-cased word forms of three letters or more, without lemmatisation, so that *underscore* and *underscores* are separate units as in the English studies [1].

### 3.2 Vietnamese: VJOL abstracts

Vietnam Journals Online is an Open Journal Systems installation hosting several hundred Vietnamese journals with an OAI-PMH interface in Dublin Core [16]. Abstracts are often given in both English and Vietnamese within the same `dc:description`; we kept the longest description containing at least eight Vietnamese diacritic characters and decoded HTML entities, which affect 28% of records. The journal is taken from `dc:publisher`. The collection comprises 48,357 Vietnamese abstracts, of which 47,165 fall in 2019–2025; the 2018 and 2026 records, incomplete years on the platform, are listed in Table 1 but not analysed. Because journals joined the platform at different times, the number of contributing journals grows with time, which makes an uncorrected year-on-year comparison a comparison of different journal sets; Section 4.3 addresses this with a journal panel.

### 3.3 Positive-control corpus

To check that the words rising in real abstracts are the words models actually produce, we generated 1,437 Korean abstracts in sixteen conditions, always from 2019 material so that the subject matter is by construction pre-ChatGPT. *Drafting*: each of three OpenAI models (`gpt-4o-mini`, `gpt-5.6-luna`, `gpt-5.6-terra`) writes 100 abstracts from 100 randomly sampled 2019 titles, instructed to produce 300 to 400 characters. *Matched length*: `gpt-5.6-luna` writes 100 abstracts from 2019 titles with the instruction changed to 600 to 700 characters, the range of real abstracts; this arm draws its titles from articles whose abstracts are 400 to 900 characters long, so it is a different title sample from the drafting arm, and the length effect is not fully separated from the sample difference there. *Editing*: `gpt-4o-mini` and `gpt-5.6-luna` are each given the same 100 real 2019 abstracts and asked to polish them without changing the content, which makes that arm paired with its own originals. *Rewriting*: the same 100 originals are presented to `gpt-5.6-luna` as a rough draft (초고) with the instruction to rewrite it to submission standard, keeping the content and results but free to restructure sentences and change expressions; this condition sits between polishing and drafting and is closer to how a model is typically used on an existing draft than either, and it is paired with its originals in the same way. Sampling parameters are the provider defaults throughout. *Translation*: because KCI articles carry an English abstract as well, three further conditions translate: gpt-5.6-luna and Claude Sonnet 5 on 99 of the same 100 articles (one article has no English abstract in the harvest store) and EXAONE on 59 of its 60 are given the article's own English abstract and asked for a Korean abstract of submission standard. These conditions are what a translation route would look like if a model produced it, and Section 5.14 compares them with the corpus. The same articles also seed six English-output conditions (514 abstracts): the same three models write an English abstract from the article's own English title, and translate the article's Korean abstract into English. These are used in Section 5.12 to measure the sensitivity of the English marker set, not as Korean controls. *Korean open-weight model*: the matched-length and editing conditions were repeated with EXAONE 3.5 7.8B Instruct [22], a Korean-English model released by LG AI Research in December 2024, run on our own CPU server through llama.cpp (4-bit Q4_K_M quantisation, temperature 0.7, top-p 0.95, fixed seed) on the first 60 titles of the matched-length sample and the first 60 of the editing originals. Because the model ignores length instructions when left to itself (a first run produced 1,289-character texts with markdown headings and numbered lists, which is kept in the release but not analysed), the local runs use a system prompt that forbids headings and lists, a completion cap of 430 tokens for drafting, and a cleaning step that strips any remaining heading or list marker and drops a final unfinished sentence; the resulting drafts average 770 characters against 690 for the OpenAI matched-length arm. *Anthropic model*: the matched-length and editing conditions were repeated with Claude Sonnet 5 (`claude-sonnet-5`), called through the Claude Code command-line client (version 2.1.235) in non-interactive mode with no tools, a one-line neutral system prompt, default sampling and extended thinking off, on the same 100 titles and the same 100 originals as the OpenAI arms, and the rewriting condition was run with both Claude Sonnet 5 (100 originals) and EXAONE (first 60); Claude's drafts average 632 characters and its polished abstracts 643 against 630 for their originals. The control corpus therefore spans three providers and two model generations, 2024 and 2026.

### 3.4 Licensing and release

Both endpoints are public and were harvested at a low request rate. We do not redistribute abstracts. The release contains the per-year document-frequency tables (every lemma and bigram that occurs in at least five abstracts of a year), the journal panel lists, the code, and the generated positive-control abstracts.

## 4 Methods

### 4.1 Lexical units

For Korean we use the Kiwi morphological analyser [13] and represent each abstract as the set of lemma–tag units it contains, keeping common nouns (NNG), verbs (VV), adjectives (VA), adverbs (MAG), roots (XR), auxiliary verbs (VX) and the copulas (VCP, VCN). Sino-Korean nouns that take the verbalising or adjectivising suffix (XSV, XSA) are fused with the suffix into one lemma, so 제공하다 "provide" and 다양하다 "be diverse" are single units rather than a noun plus a suffix. A parallel rule fuses a common noun with the derivational suffixes 적 "-al", 화 "-isation" and 성 "-ness" into one unit, so 구조적 "structural" is one lemma rather than 구조 "structure" plus a suffix; Appendix E re-runs the pipeline with this rule switched off. This fusion is what makes the verb layer of Korean visible: most of the style verbs reported below (시사하다, 규명하다, 기여하다) are of this type. Punctuation ends a clause, and every ordered pair of adjacent kept units within a clause is added as a bigram (written a+b), which captures fixed expressions such as 이러한 결과 "these results" and 중요한 역할 "important role". We call a unit a *style word* if it is anything other than a plain common noun (verbs, adjectives, adverbs, roots, fused derivations, and bigrams at least one of whose parts is such a unit; a bigram of two plain common nouns counts as content) and a *content word* if it is a plain common noun. The split is deliberately coarse; it is meant to separate words that a writer could swap without changing what the paper is about from words that name the subject. The negative control of Table 2 searches the slightly larger candidate set that admits every bigram, which can only raise its floor.

For Vietnamese we use the underthesea toolkit for word segmentation and conditional-random-field part-of-speech tagging [14], keep nouns (N), verbs (V), adjectives (A) and adverbs (R), lower-case the surface form, and add adjacent bigrams within a clause. Style words are V, A, R and bigrams; content words are N. Unlike the Korean rule, a bigram of two nouns is not excluded here, which lets topic terms such as công nghệ số into Table 8; the control floor of Section 5.7 is set by a noun–verb bigram and does not depend on this.

### 4.2 Structured-abstract normalisation

A first pass over a pre-ChatGPT control year produced apparent 33-fold excess for items such as 결론 및 제언 "conclusion and suggestions". These are the section labels of structured abstracts (목적: 방법: 결과: 결론:), which some journals adopted in the study period. Since a formatting change has nothing to do with LLMs, a regular expression removes such labels, in Korean and English, when they occur at the start of a clause and are followed by a colon (Appendix C). The labels are removed before tokenisation for every year.

### 4.3 Journal panel and capping

Both repositories change composition over time. To keep the set of sources comparable we use a *loose journal panel*: a journal enters the analysis only if it contributes at least $k = 5$ abstracts in the base period (2018–2022 for Korean, 2019–2022 for Vietnamese) and at least 5 in the target period. This keeps 1,853 of 2,282 Korean journals and 87 Vietnamese journals, and removes journals that exist only before or only after the cut, which are the main source of topical drift. A strict panel (every journal present in every year) was also tried and gives the same ranking with smaller samples. Within the panel each year is capped at a fixed random sample (40,000 Korean, 6,000 Vietnamese abstracts) so that no year dominates the frequency floor of the other years.

### 4.4 Excess frequency

Let $f_w(y)$ be the share of abstracts in year $y$ that contain unit $w$. For each unit a least-squares line is fitted through the base years and extrapolated to the target year $t$, giving the expected share $e_w(t)$, truncated at zero. The excess ratio and the excess frequency are

$$r_w(t) = f_w(t) \,/\, \max(e_w(t),\, \tfrac{1}{2}\,\bar{y}_w,\, 1/N_t), \qquad \delta_w(t) = f_w(t) - e_w(t), \tag{1}$$

where $\bar{y}_w$ is the mean base-year share and $N_t$ the number of abstracts in year $t$. The floor in the denominator, a choice of this paper rather than of Kobak et al., prevents a unit that was nearly absent before 2023 from producing an arbitrarily large ratio; such units are still visible through $\delta$, which is reported in percentage points (pp). Kobak et al. form the expectation differently: they extrapolate from the two years that precede the LLM period (2021 and 2022 for a 2024 target) with the slope truncated at zero. We fit five base years instead, which makes the slope less sensitive to any single year but extrapolates further ahead, four years for 2026. Section 5.8 measures the cost of that horizon directly on pre-LLM years, and the same section repeats the bound under the rule of Kobak et al. and under an expectation model with no trend at all. Only units that occur in at least five abstracts of a year enter the tables.

### 4.5 Uncertainty

For 65 marker units (every unit in Tables 3 and 4, plus the units fixed in a pilot pass over the same corpus), we computed 95% confidence intervals for $r$ and $\delta$ with a cluster bootstrap: whole journals are drawn with replacement once per replicate, each selected journal carrying all of its abstracts in every base and target year, because abstracts from one journal are not independent (1,000 replicates for the per-marker intervals, 500 for the set and maximum statistics, at most 12,000 abstracts per year, the fit and the extrapolation recomputed in every replicate). The bootstrap sample is smaller than the main sample for computational reasons, so point estimates in the tables come from the main run and intervals from the bootstrap, whose own point estimates (Table A1) differ from the main-run values; for 2 of the 75 bracketed values in Tables 3 and 4 the main-run value lies outside its bracket (marked †), so a bracket measures the sampling width of the subsample estimate and does not certify the printed value. One caution applies to every interval attached to a unit that the same data selected: the units in Tables 3 and 4 were chosen on the data the intervals are computed on, so their coverage is conditional on that selection and is not the nominal 95%; the split-half set bound of Section 5.9 is the exception, since its set is chosen on the other half of the journals. Section 5.8 therefore also reports a bootstrap of the selection itself, in which the maximum is recomputed inside every replicate, and a split-half design in which the markers are chosen on one set of journals and measured on another (Appendix F). The marker units were fixed before the final run, but on an earlier pass over the same corpus, which is weaker than pre-registration.

### 4.6 Controls

The negative control repeats the whole pipeline with 2022 as the target year and 2018–2021 as the base. Since ChatGPT was released on 30 November 2022, 2022 abstracts are, apart from a negligible tail, pre-LLM, and any excess found there measures the method's false-positive level: topical drift, tokeniser artefacts, journal formatting. The positive control (Section 3.3) measures, for each marker, the share of model-written abstracts that contain it in each of sixteen conditions, and compares it with the real 2019 and 2026 shares and, for the editing arm, with the same abstracts before editing.

### 4.7 Lower bound on LLM-processed abstracts

Kobak et al. observed that if a unit has excess $\delta_w(t)$, then at least $\delta_w(t)$ of the year's abstracts must have been processed by an LLM, because every surplus occurrence lives in some processed abstract and one abstract contributes at most one occurrence to a document frequency. The maximum of $\delta_w(t)$ over style words is therefore a lower bound on the processed share, regardless of how many other processed abstracts use none of the marker words [1]. The same argument holds for a set $G$ of style words: the share of abstracts containing at least one word of $G$, minus its extrapolated expectation, is also a lower bound, and the headline figure of Kobak et al., 13.5% for 2024, is of this kind, the mean of a ten-word set chosen by hand and of the set of all rare excess style words [1]. We report the single-word bound per year, per field and per language, and the set bound for the whole corpus with the words chosen on one half of the journals and the gap measured on the other (Section 5.9), so that the choice of words cannot inflate it. The single-word bound is conservative in two ways: models do not use any single word in every abstract they write (the positive control shows the most frequent marker in at most 88% of model abstracts), and human writers who have absorbed the style also contribute to it only partially. It is also a maximum over many noisy estimates, so it is biased upward under the null; Section 5.8 measures that bias directly by running the same statistic on pre-LLM target years at every extrapolation horizon the data allow, and we treat only the part of the bound above that placebo level as meaningful. The interval we attach to the bound is a journal-cluster bootstrap: journals, not abstracts, are resampled, each journal entering with all of its abstracts in all years, and the maximum is reselected within each replicate over the fixed list of tracked marker units rather than over the whole vocabulary, so the interval covers the noise in choosing among those units and not the uncertainty of the original search over all units, which the split-half design of Section 5.9 addresses instead. Because the maximum in percentage points is easily attained by a very common function verb whose share drifts by a few percent of itself, the year-by-year comparison restricts the search to units whose excess ratio is at least 1.5; the unrestricted maximum is reported alongside. Because the statistic is a maximum, its value depends on how many units are searched and on the sample size, so bounds computed on different corpora with different unit inventories are not directly comparable.

**Identification assumptions.** The step from an excess to a statement about LLM use rests on assumptions that the arithmetic does not contain, so we state them. *A1, counterfactual trend.* Absent LLMs, the document frequency of each unit in the target year would have followed the least-squares line through its base years, up to the drift and sampling error that the placebo runs of Section 5.8 measure at every extrapolation horizon. *A2, attribution.* The excess above that line is not produced by writing that no model touched, whether through a coincident change in Korean academic style, a change in what journals ask authors to write, or authors imitating a register they read elsewhere. *A3, one abstract at most once.* Document frequency counts each abstract once for each unit, so one abstract cannot supply more than one unit of excess; this holds by construction. Under A1 and A3 the statistic is a lower bound on the share of abstracts whose wording departs from the pre-LLM trend. A2 is what licenses reading that share as LLM processing, and it is an assumption, not a measurement: the timing, the positive control of Section 5.6, the fall of plain expressions, the within-article English comparison of Section 5.12 and the style-or-topic checks of Section 5.13 each test a way A2 could fail, but none of them can separate text a model wrote from text a human wrote in a register learned from models. We therefore call the statistic a conditional lower bound, write "lower bound" for it throughout as shorthand, and read it as a floor on direct LLM processing only under A1 and A2. Section 6.3 states what follows and what does not.

A floor becomes an estimate only under an assumption about how the processed abstracts were produced. If a share $\pi$ of the abstracts of a year were processed, a share $s$ of processed abstracts contains at least one word of $G$ and a share $q$ of unprocessed abstracts does, then $P = \pi\ s + (1 - \pi)\ q$ and $\pi = (P - q) \,/\, (s - q)$; with $q$ estimated by the trend expectation $Q$, the set bound $P - Q$ equals $\pi\ (s - Q)$ and falls short of $\pi$ by the factor $s - Q$. Section 5.10 measures $s$ on the positive-control corpus separately for every prompt condition, $q$ on pre-2023 abstracts, and reports the range of $\pi$ the conditions imply as a set of prompt- and model-conditioned scenarios, not as an estimate of the field's prevalence and not as a replacement for the floor. Because s is measured on 60 or 100 abstracts, each implied share is reported with a propagated range that combines the Wilson interval of s with the bootstrap interval of the gap (Table 13).

### 4.8 Fields and multi-word constructions

KCI's field codes are not exposed in the harvested format, so journals were assigned to eleven groups by keyword rules on their names (medicine and health, engineering and IT, natural science, education, law and public administration, business and economics, social science, humanities, arts and sport, convergence and content studies, other), with “other” holding the 30% of panel journals that no rule matched (Appendix D). Constructions whose parts are not adjacent, such as 단순한 X를 넘어 “beyond a mere X”, cannot be seen by the bigram layer and were counted by regular expressions on the raw text (Table 5).

## 5 Results

**Table 1:** Corpus summary. Korean abstracts come from the KCI OAI-PMH endpoint (Korean-language abstracts of at least 80 characters); Vietnamese abstracts from the VJOL repository. "Panel" columns give the number of abstracts entering the analysis after journal-panel filtering and per-year capping (Section 4.3).

| Year | KCI abstracts | KCI journals | KCI panel | VJOL abstracts | VJOL journals | VJOL panel |
|---|---|---|---|---|---|---|
| 2018 | 49,795 | 1,734 | 40,000 | 265 | 15 | – |
| 2019 | 41,484 | 1,775 | 39,419 | 4,022 | 51 | 3,719 |
| 2020 | 55,197 | 1,828 | 40,000 | 6,005 | 68 | 5,254 |
| 2021 | 38,578 | 1,679 | 37,665 | 7,424 | 81 | 6,000 |
| 2022 | 54,717 | 1,919 | 40,000 | 6,027 | 82 | 5,969 |
| 2023 | 27,672 | 1,746 | 27,437 | 8,232 | 102 | 6,000 |
| 2024 | 50,907 | 2,018 | 40,000 | 6,571 | 106 | 4,746 |
| 2025 | 45,745 | 2,000 | 40,000 | 8,884 | 110 | 5,411 |
| 2026 (Jan–Aug) | 34,201 | 1,829 | 32,630 | 927 | 3 | – |
| Total | 398,296 | 2,282 | 337,151 | 48,357 | 154 | 37,099 |

### 5.1 Negative control: 2022 is quiet

Table 2 lists the style words with the largest excess when the pipeline is run on 2022 with the 2018–2021 trend, and Section 5.8 repeats the exercise with 2020 and 2021 as targets. The largest excess of any style word in 2022 is 0.7 pp, at 1.01 times its expectation, and no style word reaches 1 pp of excess. The words that appear (있다 "be", 이해하다 "understand", 특히 "in particular", 수행하다 "perform", 고려하다 "consider") are ordinary academic vocabulary with slow pre-existing growth. One unit sits outside this floor and is worth naming: the bigram 목적 + 연구, at 1.7× and 1.0 pp, is a residue of the structured-abstract heading "목적 본 연구는 ..." written without a colon, which the pattern of Section 4.2 does not catch; it is excluded from all tables as a formatting artefact rather than a style word, together with every other unit containing a section-heading noun. With that exclusion the method's false-positive level on a pre-LLM year is 0.7 pp of excess here and at most 1.4 pp on the main frequency table across the three placebo years of Section 5.8 (computed as in this table those years reach 1.5 pp), always at a ratio below 1.2. This is the floor against which the following years should be read.

**Table 2:** Negative control. The same pipeline applied to 2022 (the last full year before ChatGPT), with expectations extrapolated from 2018–2021. The twelve style words with the largest excess in percentage points are listed; all are common academic vocabulary with slow pre-existing growth. The largest excess of any style word is 0.7 pp, so no style word reaches 1 pp of excess. Journal panel k = 5, n = 40,000 abstracts in 2022.

| Style word | Observed 2022 (%) | Expected (%) | Ratio | Excess (pp) |
|---|---|---|---|---|
| **있다** *be, exist* | 61.6 | 60.9 | 1.01 | +0.7 |
| **이해하다** | 7.4 | 6.7 | 1.10 | +0.7 |
| **특히** *in particular* | 15.3 | 14.7 | 1.04 | +0.6 |
| **수행하다** *perform* | 10.5 | 9.9 | 1.06 | +0.6 |
| **고려하다** *consider* | 11.1 | 10.6 | 1.05 | +0.5 |
| **지니다** *have, bear* | 5.7 | 5.2 | 1.10 | +0.5 |
| **나아가다** | 6.7 | 6.2 | 1.08 | +0.5 |
| **검토하다** *examine, review* | 8.4 | 7.9 | 1.06 | +0.4 |
| **파악하다** *grasp, identify* | 11.8 | 11.4 | 1.04 | +0.4 |
| **중요하다** *be important* | 15.0 | 14.6 | 1.03 | +0.4 |
| **주목하다** | 5.7 | 5.2 | 1.08 | +0.4 |
| **이론적** *theoretical* | 4.5 | 4.1 | 1.09 | +0.4 |

### 5.2 Time course: a lag year, an onset, and acceleration

Figure 1 shows the yearly share of abstracts containing twelve marker units, and Figure 6 the same series by month. Three phases are visible in every rising panel. In 2023, the first year after ChatGPT, nothing moves: the largest excess of any style word in 2023 is 1.0 pp (및 "and", 1.02×), inside the placebo range of Section 5.8. The ratios in this section are corpus-wide values on the full panel (up to 40,000 abstracts per year); the point estimates of Table A1 come from the 12,000-abstract bootstrap subsample and, for rare units, differ by several tenths. In 2024 a first group rises moderately: 촉진하다 "promote" (2.0×), 탐구하다 "explore" (2.0×), 통합하다 "integrate" (1.7×), 기여하다 "contribute" (1.7×, 8.2% of abstracts), 강화하다 "strengthen" (1.6×), 강조하다 "emphasize" (1.5×), 시사하다 "suggest" (1.4×). In 2025 the amplitude doubles and a second group joins: 단순하다 "mere" (6.3×), 기능하다 "function as" (3.9×), 통합하다 (2.7×), 넘다 "go beyond" (2.8×), 조명하다 "shed light on" (2.7×), 구조적 "structural" (2.7×), 시사하다 (2.5×, +7.8 pp), 기여하다 (2.5×, +7.2 pp). In 2026 the second group is at its highest level while the first falls back: 단순하다 reaches 11.0× (12.3% of abstracts against 1.1% expected), 기능하다 7.0×, 통합하다 4.8×, 구조적 6.0× (15.1% against 2.5%), 넘다 4.0×, 규명하다 "elucidate" 3.9×, 시사하다 4.0× (21.4% against 5.3%), 확인되다 "be confirmed" 2.9× (21.5% against 7.4%), while 기여하다 drops from 11.9% to 7.8% (1.6×), 강조하다 from 9.0% to 5.5% (1.2×), 탐구하다 from 5.1% to 3.1% (1.2×), and 중요한 역할 "important role" falls back toward trend (1.26×).

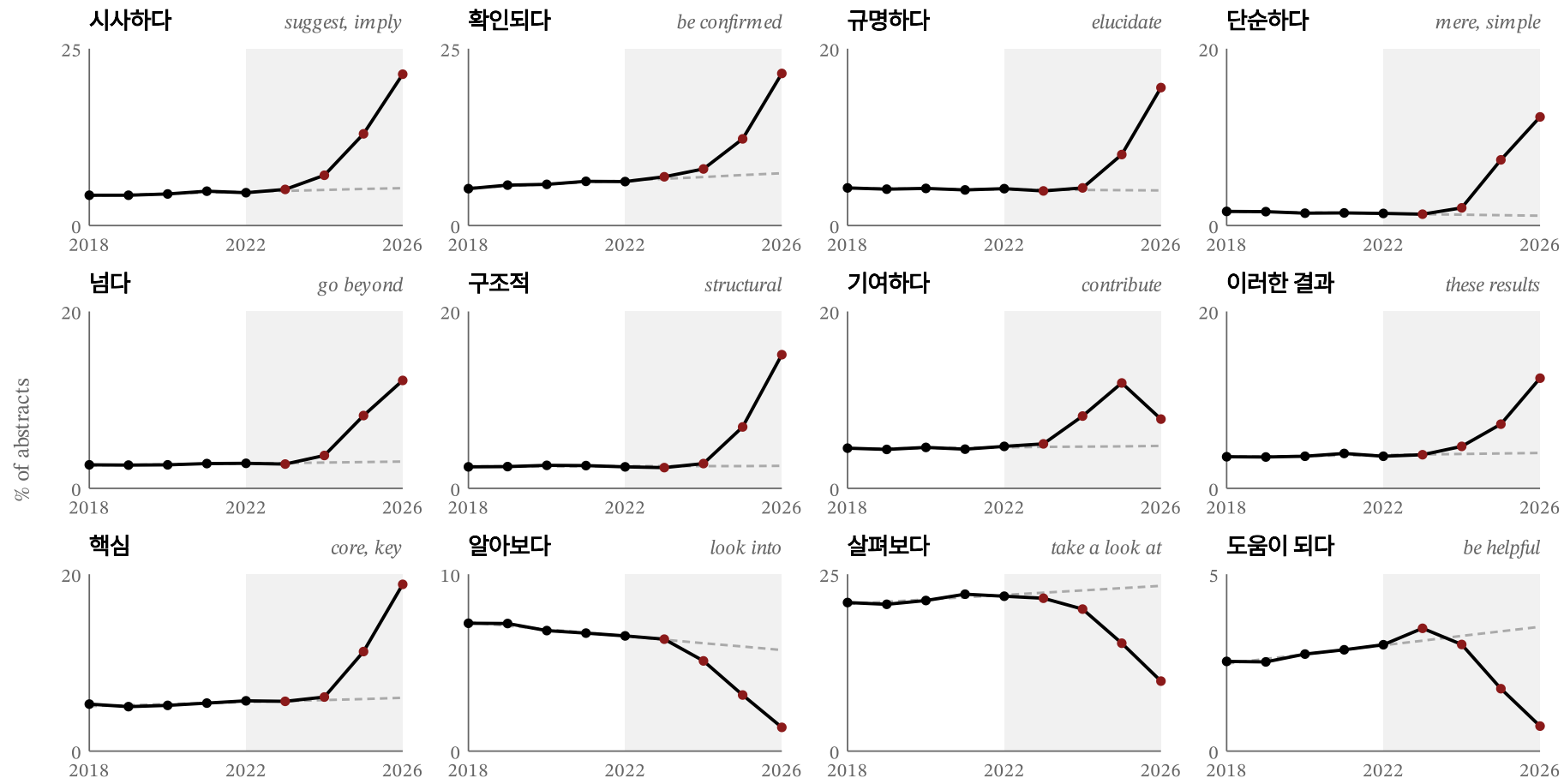

**Figure 1:** Share of Korean abstracts containing twelve marker lemmas, 2018–2026 (journal panel, at most 40,000 abstracts per year). Solid line: observed; dashed line: least-squares trend through 2018–2022 extrapolated forward; shaded area: years after the release of ChatGPT (30 November 2022); red points: 2023–2026. The first nine panels rise, the last three fall. Note the different vertical scales.

The same time course appears in the negative direction. 알아보다 "look into" falls from 6.8% (2020) to 5.1% (2024), 3.2% (2025) and 1.3% (2026), which is 0.24 of its expected 5.7%; 많이 "a lot" to 0.30 of expectation; 도움이 되다 "be helpful" to 0.20; 진행하다 "carry out (a study)" to 0.31; 살펴보다 "take a look at", the most common plain verb of Korean research prose, from 21% to 9.9% (0.42). None of these words moved in 2023.

### 5.3 The dictionary: what rose and what fell

Table 3 lists the thirty style words with the largest 2026 excess in percentage points, restricted to words at least twice their expectation, and Table 4 the fifteen with the largest deficit ratio. Figure 2 places every unit with positive excess on a ratio-by-excess map for 2025 and 2026; Figure 3 shows the two lists side by side as changes from the 2018–2022 mean.

**Table 3:** Korean style words with the largest excess in 2026 (ratio at least 2 and excess at least 1 pp), ordered by excess percentage points. A two-word expression is listed only when its excess is larger than that of either of its parts, so the same shift is not counted twice. Columns give the share of abstracts containing the lemma in selected years, the expected 2026 share from the 2018–2022 linear trend, the excess ratio and the excess in pp. Brackets are 95% bootstrap intervals (1,000 replicates resampling whole journals, at most 12,000 abstracts per year, Section 4.5), computed around the subsample point estimates of Table A1, which differ from the corpus-wide values printed here; † marks a printed value that lies outside its bracket. The intervals are conditional on the selection rule that produced this table and are not nominal 95% intervals (Section 4.5). n = 32,630 abstracts in 2026. The last column subtracts each word's own excess in the 2022 negative control, so it reports the excess above the level that the method produces for that word on a pre-LLM year.

| Style word | 2018 | 2020 | 2022 | 2024 | 2025 | 2026 | Exp. 2026 | Ratio [Table A1 interval] | Excess pp [Table A1 interval] | Adj. pp |
|---|---|---|---|---|---|---|---|---|---|---|
| 시사하다 *suggest, imply* | 4.3 | 4.5 | 4.6 | 7.1 | 13.0 | 21.4 | 5.3 | 4.0 [3.4, 4.7] | +16.1 [14.7, 17.4] | +16.1 |
| 확인되다 *be confirmed* | 5.2 | 5.8 | 6.2 | 8.0 | 12.3 | 21.5 | 7.4 | 2.9 [2.9, 4.0] | +14.1 [13.8, 16.7] | +14.1 |
| 구조적 *structural* | 2.4 | 2.6 | 2.4 | 2.8 | 6.9 | 15.1 | 2.5 | 6.0 [4.7, 7.2] | +12.6 [11.6, 13.5] | +12.6 |
| 규명하다 *elucidate* | 4.3 | 4.2 | 4.2 | 4.3 | 8.0 | 15.6 | 4.0 | 3.9 [3.4, 5.3] | +11.6 [10.5, 12.8] | +11.6 |
| 단순하다 *mere, simple* | 1.6 | 1.4 | 1.4 | 2.0 | 7.4 | 12.3 | 1.1 | 11.0 [6.1, 11.6] | +11.2 [9.8, 11.6] | +11.1 |
| 넘다 *go beyond* | 2.6 | 2.7 | 2.8 | 3.7 | 8.2 | 12.2 | 3.0 | 4.0 [3.3, 5.0] | +9.2 [8.1, 10.0] | +9.2 |
| 실증적 *empirical* | 2.6 | 2.5 | 2.3 | 3.0 | 7.8 | 10.8 | 2.2 | 4.9 [3.6, 6.0] | +8.6 [7.3, 9.2] | +8.6 |
| 지니다 *have, bear* | 6.0 | 5.5 | 5.5 | 5.7 | 7.7 | 11.3 | 4.9 | 2.3 [1.7, 2.4] | +6.4 [4.8, 7.0] | +5.9 |
| 결합하다 *combine* | 1.4 | 1.6 | 1.8 | 2.2 | 3.8 | 8.2 | 2.2 | 3.7 [2.9, 4.3] | +6.0 [5.3, 6.6] | +6.0 |
| 제도적 *institutional* | 2.4 | 2.5 | 2.3 | 2.5 | 5.2 | 8.0 | 2.1 | 3.7 [2.7, 4.5] | +5.8 [4.9, 6.4] | +5.8 |
| 기능하다 *function as* | 0.9 | 0.9 | 1.0 | 1.2 | 3.7 | 6.8 | 1.0 | 7.0 [6.7, 14.7] | +5.8 [5.4, 6.8] | +5.7 |
| 통합적 *integrative* | 1.4 | 1.4 | 1.3 | 1.7 | 4.0 | 6.9 | 1.2 | 5.7 [4.0, 7.6] | +5.7 [4.9, 6.3] | +5.7 |
| 작동하다 *operate* | 1.1 | 1.1 | 1.2 | 1.3 | 2.5 | 6.3 | 1.3 | 5.1 [3.6, 6.7] | +5.1 [4.2, 5.6] | +4.9 |
| 체계적 *systematic* | 3.5 | 3.5 | 3.7 | 4.7 | 7.7 | 8.8 | 3.8 | 2.3 [2.0, 2.8] | +5.0 [4.0, 5.7] | +4.8 |
| 이론적 *theoretical* | 4.6 | 4.3 | 4.5 | 4.5 | 7.2 | 9.1 | 4.3 | 2.1 [1.7, 2.6] | +4.8 [3.8, 5.9] | +4.4 |
| 해석하다 *interpret* | 3.5 | 3.6 | 3.5 | 3.7 | 5.0 | 8.2 | 3.4 | 2.4 [1.9, 2.8] | +4.8 [3.8, 5.5] | +4.8 |
| 형성되다 *be formed* | 3.2 | 2.9 | 3.0 | 2.8 | 3.9 | 7.4 | 2.8 | 2.7 [2.3, 4.0] | +4.6 [4.0, 5.8] | +4.3 |
| 확장하다 *extend* | 1.9 | 2.0 | 2.3 | 3.0 | 4.3 | 7.4 | 2.9 | 2.6 [2.1, 3.0] | +4.6 [3.7, 5.2] | +4.6 |
| 작용하다 *act on* | 3.8 | 3.7 | 3.7 | 4.2 | 7.2 | 7.9 | 3.5 | 2.3 [1.7, 2.6] | +4.4 [3.2, 5.0] | +4.4 |
| 형성하다 *form* | 3.4 | 3.1 | 3.4 | 3.7 | 5.1 | 7.6 | 3.3 | 2.3 [1.8, 2.6] | +4.3 [3.2, 4.9] | +4.1 |
| 실질적 *substantive* | 2.9 | 2.8 | 2.8 | 3.9 | 7.3 | 6.6 | 2.6 | 2.5 [1.9, 2.9] | +4.0 [2.9, 4.4] | +3.7 |
| 강화하다 *strengthen* | 3.2 | 3.1 | 3.5 | 5.6 | 8.2 | 7.6 | 3.7 | 2.1 [1.8, 2.6] | +4.0 [3.2, 4.8] | +3.7 |
| 상대적 *relative* | 4.5 | 4.3 | 4.2 | 4.2 | 5.0 | 7.7 | 3.8 | 2.0 [1.6, 2.4] | +3.9 [3.0, 4.8] | +3.6 |
| 구조화 *structuring* | 1.2 | 1.4 | 1.4 | 1.4 | 2.6 | 5.4 | 1.7 | 3.2 [2.4, 4.4] | +3.7 [2.9, 4.2] | +3.7 |
| 실천적 *practical* | 1.8 | 1.8 | 2.1 | 2.2 | 5.1 | 6.0 | 2.4 | 2.6 [2.1, 3.5] | +3.7 [3.0, 4.3] | +3.7 |
| 통합하다 *integrate* | 1.0 | 0.9 | 1.0 | 1.6 | 2.6 | 4.6 | 0.9 | 4.8 [3.7, 8.6] | +3.6 [3.2, 4.3] | +3.5 |
| 결론적 *conclusive* | 2.0 | 2.0 | 1.8 | 2.3 | 3.0 | 5.2 | 1.7 | 3.1 [2.2, 4.0] | +3.5 [2.6, 4.0] | +3.5 |
| 충분히 *sufficiently* | 1.8 | 1.7 | 1.9 | 2.0 | 2.8 | 5.5 | 2.0 | 2.7 [2.0, 3.2] | +3.5 [2.6, 3.8] | +3.4 |
| 결합되다 *be combined* | 0.9 | 0.7 | 0.7 | 0.8 | 1.3 | 3.9 | 0.5 | 8.2 [4.7, 11.5] | +3.4 [2.9, 4.0] | +3.4 |
| 정서적 *emotional* | 1.6 | 1.6 | 1.8 | 2.1 | 4.3 | 5.6 | 2.1 | 2.6 [2.0, 3.4] | +3.4 [2.7, 4.1] | +3.4 |

First, the top of the list is dominated by verbs that state what a finding means: 시사하다 "suggest, imply" (+16.1 pp, 95% CI 14.7 to 17.4, ratio 3.4 to 4.7), 확인되다 "be confirmed" (+14.1), 규명하다 "elucidate" (+11.6), 검토하다 "examine" (+7.6), 보여주다 "show" (+8.0), together with the frame 이러한 결과는 "these results" (+8.5), all corpus-wide values (Table A1 gives the subsample points). A 2026 abstract that "confirms" a "structural relation" and "suggests" an "implication" uses four of the top ten in one sentence. Second, an adjectival layer that Korean forms with the Sino-Korean suffix 적: 구조적 "structural" rose from 2.4% of abstracts in 2022 to 15.1% (6.0×), 실증적 "empirical" to 10.8% (4.9×), 제도적 "institutional" to 8.0% (3.7×). With them come the evaluative words 단순하다 "mere" (+11.2 pp, 11.0×), 넘다 "go beyond" (+9.2, 4.0×) and 충분히 "sufficiently", and among content nouns 메커니즘 "mechanism" (4.5×), 기제 "mechanism", Sino-Korean (4.6×), 틀 "framework" and 통합 "integration". Third, the intervals are narrow: across the 65 tracked marker units, 19 have a corpus-wide 2026 excess above 5 pp (the values of Tables 3 and 4); 14 of those have a bootstrap ratio interval (Table A1) that excludes 2 and 9 have one that excludes 3. The exceptions are units that were already common before 2023, 특히 "in particular" (1.78×), 보여주다 "show" (1.98×), 검토하다 "examine" (1.91×), 지니다 "have, bear" (2.29×), 체계적 (2.33×), which carry large excess in percentage points precisely because of their high baseline. The last column of Table 3 subtracts each word's own excess in the 2022 placebo year; the largest such correction is 0.5 pp, on 지니다, and no other correction reaches 1 pp, so the ordering of the table is unchanged.

**Table 4:** Korean style words with the largest deficit in 2026 (observed share below the trend expectation), ordered by ratio among words with a deficit of at least 1 pp. Same columns as Table 3 (brackets and † as there), plus a conservative comparison. As in Table 3, a two-word expression is listed only when its deficit exceeds that of either of its parts, which is why 도움이 되다 is absent: its parts 되다 and 도움 fall further. Because a word whose share was still rising through the base period gets a high extrapolated expectation, the deficit against the trend can overstate the fall; the last three columns therefore repeat the comparison against the *lowest* share the word reached in any base year, which no extrapolation can inflate. Only words that fall below that lowest base-year share are listed, the symmetric counterpart of the filter applied to Table 8. Every word in the unfiltered top fifteen passes this condition, so the direction is not an artefact of extrapolation, but three of them (진행하다, 기대하다, 세계적) lose about 70% of their apparent size under the conservative comparison, while 보다 and 알아보다 gain.

| Style word | 2018 | 2020 | 2022 | 2024 | 2025 | 2026 | Exp. 2026 | Ratio [Table A1 interval] | Deficit pp | Base min (%) | Cons. ratio | Cons. pp |
|---|---|---|---|---|---|---|---|---|---|---|---|---|
| 알아보다 *look into* | 7.2 | 6.8 | 6.5 | 5.1 | 3.2 | 1.3 | 5.7 | 0.24 [0.16, 0.26] | −4.4 | 6.5 | 0.21 | −5.2 |
| 연구하다 | 4.4 | 4.1 | 4.1 | 3.8 | 2.5 | 1.0 | 3.9 | 0.24 [0.25, 0.41]† | −3.0 | 4.1 | 0.23 | −3.2 |
| 노력하다 | 1.7 | 1.7 | 1.8 | 1.6 | 1.0 | 0.5 | 1.8 | 0.27 [0.30, 0.68]† | −1.3 | 1.6 | 0.30 | −1.2 |
| 끼치다 | 1.2 | 1.3 | 1.3 | 1.1 | 0.7 | 0.4 | 1.4 | 0.28 [0.20, 0.46] | −1.0 | 1.2 | 0.32 | −0.8 |
| 많이 *a lot* | 4.8 | 4.7 | 4.4 | 3.7 | 2.5 | 1.2 | 4.0 | 0.30 [0.28, 0.49] | −2.8 | 4.4 | 0.27 | −3.2 |
| 진행하다 *carry out* | 6.0 | 7.6 | 8.8 | 9.1 | 6.9 | 3.6 | 11.7 | 0.31 [0.26, 0.33] | −8.1 | 6.0 | 0.61 | −2.4 |
| 기대하다 | 4.2 | 5.2 | 6.1 | 6.3 | 4.7 | 2.7 | 8.1 | 0.33 [0.28, 0.39] | −5.4 | 4.2 | 0.63 | −1.6 |
| 소개하다 | 2.2 | 2.1 | 2.3 | 2.0 | 1.4 | 0.7 | 2.2 | 0.33 [0.26, 0.60] | −1.4 | 2.0 | 0.35 | −1.3 |
| 세계적 | 1.4 | 1.7 | 1.9 | 1.6 | 1.4 | 0.8 | 2.4 | 0.34 [0.28, 0.54] | −1.6 | 1.4 | 0.59 | −0.6 |
| 변화되다 | 1.3 | 1.5 | 1.4 | 1.1 | 0.8 | 0.5 | 1.6 | 0.34 [0.33, 0.68] | −1.0 | 1.3 | 0.41 | −0.8 |
| 보다 | 13.1 | 12.7 | 11.9 | 9.9 | 7.0 | 3.8 | 10.9 | 0.35 [0.32, 0.42] | −7.0 | 11.9 | 0.32 | −8.1 |
| 알다 | 8.9 | 8.4 | 8.1 | 7.1 | 4.9 | 2.7 | 7.4 | 0.36 [0.33, 0.48] | −4.7 | 8.1 | 0.33 | −5.4 |
| 생각하다 | 3.7 | 3.5 | 3.1 | 2.7 | 1.9 | 0.9 | 2.5 | 0.37 [0.31, 0.59] | −1.6 | 3.1 | 0.30 | −2.2 |
| 일어나다 | 2.1 | 2.3 | 2.1 | 1.8 | 1.1 | 0.7 | 2.0 | 0.37 [0.33, 0.70] | −1.3 | 2.1 | 0.35 | −1.3 |
| 많다 | 14.3 | 13.7 | 13.1 | 11.4 | 8.1 | 4.5 | 12.2 | 0.37 [0.31, 0.42] | −7.7 | 13.1 | 0.34 | −8.6 |

Table 4 is the complement. The words that fell are the everyday verbs and adverbs of Korean academic prose: 알아보다, 살펴보다, 도움이 되다, 많이, 진행하다, 가장 많이 "the most", and the classic title frame ~에 관한 연구 "a study on …". The topic words of the pandemic (대면 "face-to-face", 온라인 수업 "online class", 방역 "quarantine") also fell, as they should; the panel does not remove topic change, and the style/content split is what separates the two kinds of decline.

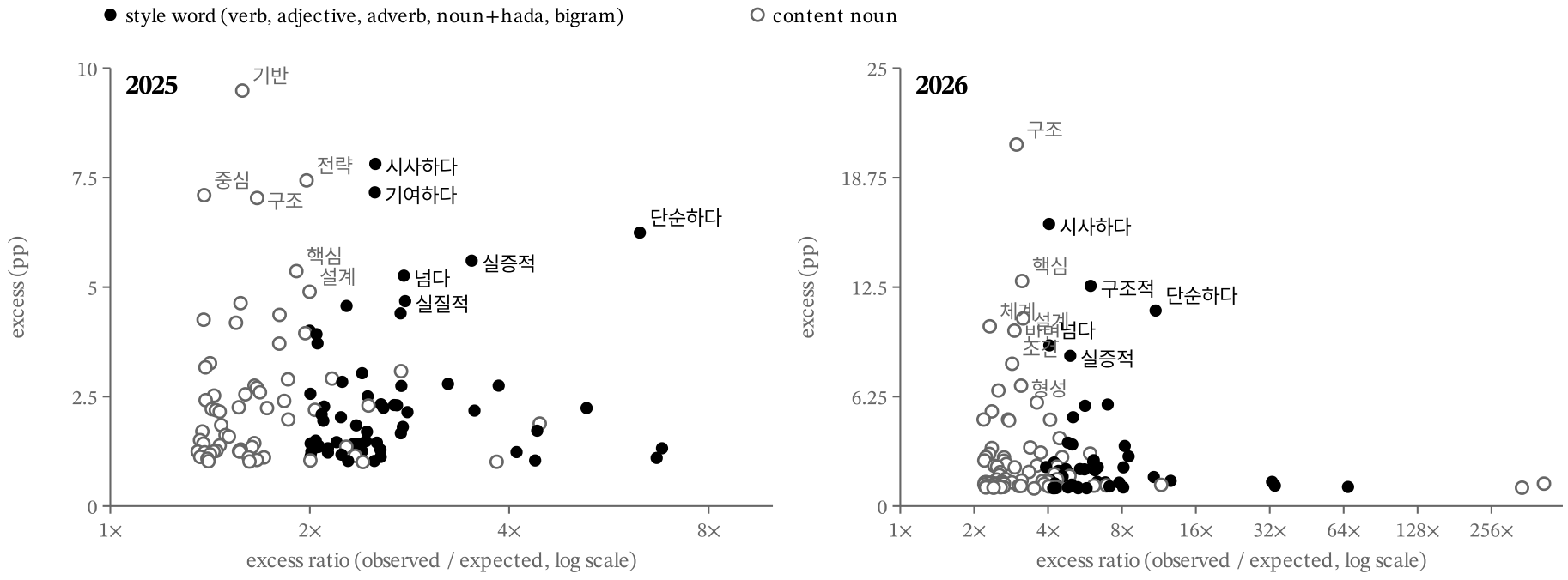

**Figure 2:** Excess map for 2025 (left) and 2026 (right): every unit with at least 1 pp excess, plotted by excess ratio (log scale) against excess in percentage points. Filled: style words (verbs, adjectives, adverbs, fused derivations, bigrams containing one); hollow: content nouns and noun–noun bigrams. Units with a near-zero base frequency, such as the generative-AI topic bigrams at the far right of the 2026 panel, obtain large ratios but small excess; the words that matter for prevalence sit high on the vertical axis.

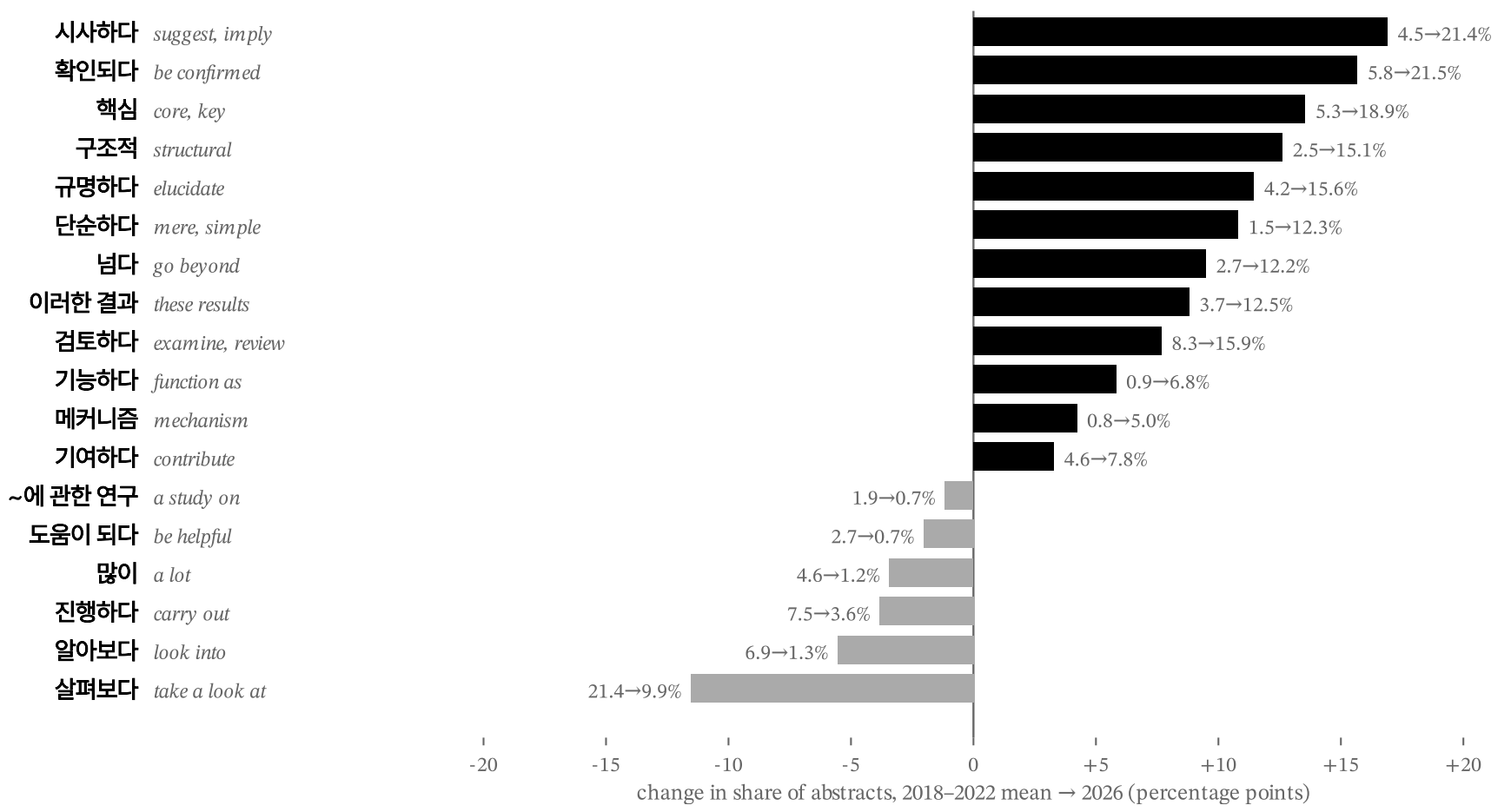

**Figure 3:** Change in the share of abstracts containing selected words between the 2018–2022 mean and 2026, in percentage points. Black: rising style words; grey: falling ones. Labels give the two shares.

### 5.4 Constructions

Bigrams see only adjacent units. Table 5 counts, by regular expression on the raw text and without per-year capping, constructions whose parts may be separated. The largest ratio of the period belongs to 단순한 X를 넘어 "beyond a mere X": 0.12% of abstracts in 2019, 0.68% in 2024, 3.71% in 2025 and 4.93% in 2026, a ratio of 61 against trend. Its adverbial variant 단순히 X를 넘어 rose only 6×, so the models prefer the adjectival form. The predicate X을 시사한다 "suggests X" closes 20.8% of 2026 abstracts (3.4% in 2019); 이러한 결과는 "these results" opens 10.6% (1.8%); 본 연구는 … 규명 "this study elucidates …" 6.6% (1.1%); 구조적 "structural" 15.1% (2.5%); 확인되었다 "was confirmed" 16.1% (3.5%); 체계적으로 "systematically" 5.1% (1.2%). The same table shows the turnover of Section 5.2 at the level of phrases: 중요한 역할 "important role" went 1.1% → 2.7% (2024) → 1.3% (2026), 기여할 것으로 기대 "is expected to contribute" 0.2% → 1.1% (2025) → 0.6% (2026), 실질적인 "substantive" 1.4% → 3.5% (2025) → 2.5% (2026).

**Table 5:** Multi-word constructions counted by regular expression on the raw text (share of abstracts, %), on the same journal panel without per-year capping (n = 32,630 in 2026). The morphological pipeline cannot see non-adjacent constructions such as "단순한 X를 넘어" (beyond a mere X), so these are counted separately. Ratios are against the 2018–2022 linear trend. These are surface strings counted on the uncapped panel, so they do not coincide with the lemma-level values of Table 3: "X을/를 시사한다" reaches 4.7× here where the lemma 시사하다 reaches 4.1× there, because the lemma also covers forms the construction does not match and the two are measured on different samples. The lemma value of Table 3 is the one quoted in the text.

| Construction | 2019 | 2022 | 2024 | 2025 | 2026 | Ratio 2025 | Ratio 2026 | Excess 2026 (pp) |
|---|---|---|---|---|---|---|---|---|
| X을/를 시사한다 | 3.40 | 3.86 | 6.37 | 12.25 | 20.77 | 2.8 | 4.6 | +16.2 |
| 구조적 | 2.45 | 2.44 | 2.79 | 6.87 | 15.08 | 2.7 | 5.9 | +12.5 |
| 확인되었다 | 3.50 | 4.01 | 5.26 | 8.80 | 16.22 | 1.9 | 3.3 | +11.3 |
| 특히 | 15.63 | 15.28 | 18.37 | 24.68 | 24.88 | 1.7 | 1.8 | +10.7 |
| 이러한 결과는 | 1.77 | 1.83 | 2.65 | 5.28 | 10.61 | 2.8 | 5.5 | +8.7 |
| 본 연구는 ~을/를 규명 | 1.12 | 1.09 | 1.13 | 3.02 | 6.61 | 2.9 | 6.5 | +5.6 |
| 단순한 X를/을 넘어 | 0.12 | 0.12 | 0.68 | 3.71 | 4.93 | 41.4 | 60.8 | +4.8 |
| 구조화 | 1.18 | 1.37 | 1.41 | 2.58 | 5.39 | 1.7 | 3.4 | +3.8 |
| 체계적으로 | 1.20 | 1.33 | 2.10 | 3.96 | 5.05 | 3.0 | 3.8 | +3.7 |
| 나아가 | 5.77 | 6.04 | 5.94 | 6.73 | 9.49 | 1.1 | 1.5 | +3.3 |
| 시사점을 제공한다 | 0.82 | 0.95 | 1.65 | 2.87 | 3.21 | 2.6 | 2.9 | +2.1 |
| 종합적으로 | 1.06 | 1.31 | 1.77 | 2.47 | 3.09 | 1.6 | 2.0 | +1.5 |
| 실질적인 | 1.39 | 1.36 | 2.04 | 3.48 | 2.49 | 2.7 | 1.9 | +1.2 |
| 핵심적인 | 0.91 | 0.87 | 1.00 | 1.48 | 1.52 | 1.8 | 1.9 | +0.7 |
| 메커니즘을/기제를 규명 | 0.04 | 0.04 | 0.03 | 0.20 | 0.70 | 4.3 | 14.9 | +0.7 |
| 단순히 X를/을 넘어 | 0.11 | 0.11 | 0.37 | 0.58 | 0.51 | 4.8 | 4.1 | +0.4 |
| 기여할 것으로 기대 | 0.17 | 0.27 | 0.81 | 1.11 | 0.64 | 3.7 | 2.0 | +0.3 |
| 중요한 역할 | 1.14 | 1.14 | 2.68 | 2.51 | 1.27 | 2.4 | 1.2 | +0.2 |
| 많은 도움 | 0.13 | 0.08 | 0.08 | 0.04 | 0.01 | 0.4 | 0.2 | −0.1 |
| 구체적으로 | 4.31 | 4.46 | 4.68 | 4.78 | 4.31 | 1.1 | 0.9 | −0.3 |
| 도움이 될 것 | 0.73 | 0.82 | 0.80 | 0.43 | 0.13 | 0.5 | 0.2 | −0.7 |
| 알아보고자 | 2.27 | 1.82 | 1.39 | 0.90 | 0.40 | 0.5 | 0.3 | −1.2 |
| 살펴보고자 | 2.55 | 2.77 | 2.35 | 1.86 | 1.17 | 0.7 | 0.4 | −1.7 |

### 5.5 Fields

Table 6 and Figure 4 stratify by field. The composite index (the share of abstracts containing at least one of eleven rising style verbs and adjectives) roughly doubles to quadruples across fields between 2019 and 2026, but the fields start from different levels and move at different speeds. In absolute terms the most affected fields in 2026 are business and economics (71.8% of abstracts, from 30.9% in 2019), arts and sport (71.1%, from 33.0%) and education (64.5%, from 29.5%); the least affected are engineering and IT (46.9%, from 15.1%) and law and public administration (54.7%, from 28.6%). In relative terms the ordering reverses: natural science quadruples its index (13.2% to 51.6%), engineering and IT more than triples it (15.1% to 46.9%) and medicine and health triples it (19.5% to 59.2%), while the humanities, which started highest, only double. The fields whose Korean-language abstracts were plainest in 2019 moved the most.

The field bound, computed within each field over the tracked style markers, shows the same spread: 26.1% for medicine and health, 26.2% for business and economics and 22.3% for natural science in 2026, against 16.2% for engineering and IT and 12.7% for the unclassified group. The word attaining the bound is 시사하다 "suggest" in most fields, 단순하다 "mere" in the humanities and 구조적 "structural" in law and public administration (Section 6.1). Two cautions apply: this bound searches only the tracked markers, so it is not the same statistic as the corpus-wide bound of Table 9, and the smaller fields (natural science n = 866, medicine n = 954 in 2026) have a higher placebo floor than the corpus as a whole, so the ordering among the lower half of the table should not be read as significant. The onset is common: no field moves before 2023, and in 2023 itself the largest within-field excess is 2.9 pp (social science), against a placebo floor that is higher inside a field than in the whole corpus because the samples are smaller. From 2024 every field turns; only the amplitude is field-specific. One caution applies to the medical and natural-science figures: these fields publish a smaller share of their work with Korean-language abstracts, so the Korean-abstract population there is not representative of the field as a whole.

**Table 6:** Field-stratified results. Fields are assigned from journal names by keyword rules (Appendix D); "other" collects journals no rule matched. The composite index is the share of abstracts containing at least one of eleven rising style words (시사하다, 기여하다, 단순하다, 넘다, 기능하다, 규명하다, 통합하다, 조명하다, 강화하다, 강조하다, 탐구하다). Field sizes differ, so the maximum-based bound is noisier in the smaller fields; the placebo floor of Section 5.8 grows as the sample shrinks. Ratios are 2026 excess ratios of three marker words; LB is the lower bound on LLM-processed abstracts computed within the field, as the maximum excess in pp over the tracked *style* markers only (a smaller search set than the corpus-wide bound of Table 9, so the two are not interchangeable). At most 3,000 abstracts per field and year.

| Field | n 2026 | Comp. 2019 | Comp. 2024 | Comp. 2025 | Comp. 2026 | 시사하다 × | 단순하다 × | 규명하다 × | LB 2025 (%) | LB 2026 (%) |
|---|---|---|---|---|---|---|---|---|---|---|
| 경영·경제 | 1,571 | 31 | 44 | 59 | 72 | 3.0 | 16.4 | 3.6 | 13.1 | 26.2 |
| 예술·체육 | 1,620 | 33 | 49 | 66 | 71 | 7.8 | 13.0 | 3.9 | 14.2 | 19.7 |
| 교육 | 3,000 | 30 | 37 | 48 | 64 | 3.6 | 14.6 | 3.4 | 7.7 | 21.1 |
| 융합·콘텐츠 | 3,000 | 21 | 35 | 54 | 61 | 5.5 | 11.0 | 3.5 | 12.2 | 21.5 |
| 사회과학 | 3,000 | 29 | 38 | 52 | 61 | 3.4 | 11.6 | 5.8 | 7.6 | 16.8 |
| 인문학 | 3,000 | 32 | 39 | 51 | 60 | 3.7 | 10.0 | 2.9 | 9.0 | 15.8 |
| 의약학 | 954 | 20 | 29 | 43 | 59 | 4.7 | 30.7 | 3.7 | 12.6 | 26.1 |
| 기타 | 3,000 | 27 | 34 | 47 | 56 | 3.7 | 7.5 | 3.6 | 6.1 | 12.7 |
| 법·행정 | 1,567 | 29 | 35 | 44 | 55 | 2.8 | 8.5 | 5.8 | 7.3 | 18.0 |
| 자연과학 | 866 | 13 | 24 | 39 | 52 | 6.3 | 11.6 | 4.5 | 8.9 | 22.3 |
| 공학·IT | 3,000 | 15 | 27 | 37 | 47 | 7.8 | 7.1 | 5.6 | 8.6 | 16.2 |

| | 2018 | 2019 | 2020 | 2021 | 2022 | 2023 | 2024 | 2025 | 2026 | |
|---|---|---|---|---|---|---|---|---|---|---|
| 경영·경제 | 31 | 31 | 33 | 35 | 34 | 36 | 44 | 59 | 72 | n=1,571 · LB 26% |
| 예술·체육 | 33 | 33 | 33 | 33 | 32 | 35 | 49 | 66 | 71 | n=1,620 · LB 20% |
| 교육 | 28 | 30 | 30 | 29 | 30 | 31 | 37 | 48 | 64 | n=3,000 · LB 21% |
| 융합·콘텐츠 | 22 | 21 | 21 | 23 | 23 | 23 | 35 | 54 | 61 | n=3,000 · LB 22% |
| 사회과학 | 30 | 29 | 30 | 28 | 28 | 31 | 38 | 52 | 61 | n=3,000 · LB 17% |
| 인문학 | 32 | 32 | 32 | 32 | 31 | 34 | 39 | 51 | 60 | n=3,000 · LB 16% |
| 의약학 | 21 | 20 | 20 | 21 | 22 | 22 | 29 | 43 | 59 | n=954 · LB 26% |
| 기타 | 25 | 27 | 27 | 28 | 26 | 27 | 34 | 47 | 56 | n=3,000 · LB 13% |
| 법·행정 | 31 | 29 | 28 | 27 | 30 | 30 | 35 | 44 | 55 | n=1,567 · LB 18% |
| 자연과학 | 14 | 13 | 14 | 16 | 16 | 19 | 24 | 39 | 52 | n=866 · LB 22% |
| 공학·IT | 14 | 15 | 14 | 16 | 16 | 17 | 27 | 37 | 47 | n=3,000 · LB 16% |

**Figure 4:** Composite index by field and year: the share of abstracts containing at least one of eleven rising style verbs and adjectives (시사하다, 기여하다, 단순하다, 넘다, 기능하다, 규명하다, 통합하다, 조명하다, 강화하다, 강조하다, 탐구하다). Darker is higher. The right-hand column gives the number of 2026 abstracts and the 2026 lower bound on LLM-processed abstracts for the field. At most 3,000 abstracts per field and year.

### 5.6 Positive control: the models write the rising words

Table 7 and Figure 5 compare the share of real abstracts containing each marker with the share of model-written abstracts containing it, across sixteen Korean-output conditions spanning drafting, matched-length drafting, polishing, translation and rewriting, from five models, with the mean text length of every condition reported. Length matters: document frequency grows with length, and the first three conditions produced abstracts of 313 to 509 characters against about 690 for real ones, which depresses their rates. The condition that matches real length (600 to 700 characters requested, 690 produced) is therefore the one to read against the 2026 column.

Under that condition the correspondence is close and in both directions, within the limits of 100 abstracts per condition, for which a 95% interval on a rate near 30% is about ±9 pp (Wilson intervals in brackets). The model writes 시사하다 into 27% [19, 36] of its abstracts against 22% [21, 24] of real 2026 abstracts, 확인되다 41% [32, 51] against 21% [20, 23], 단순하다 31% [23, 41] against 12% [11, 14], 검토하다 29% against 16%, 이러한 결과 22% against 12%, 규명하다 21% against 14%, 핵심 21% against 19%, 넘다 18% against 12%. Of the 27 rising units in Table 7, the length-matched model's interval lies at or above the real 2026 rate for nine (확인되다, 단순하다, 결합하다, 조명하다, 촉진하다, 강화하다, 검토하다, 특히, 이러한 결과), below it for three (구조적, 기제, 보여주다) and overlaps it for the remaining fifteen, which at this sample size cannot be separated from their real rates; 구조적 is the largest gap (2% [1, 7] against 15%). The words carrying most of the measured excess are in the first two groups, which is what the excess-vocabulary account requires, while the three below set a limit on how much of the change these models explain. The falling markers go the other way: 알아보다 0%, 도움이 되다 0%, 진행하다 0%, 살펴보다 4%, against 1%, 1%, 4% and 11% in real 2026 abstracts and 7%, 2%, 7% and 20% in 2019. The exception is the 2024 model, which writes 살펴보다 into 23% of its abstracts, slightly more than real 2019 text, and the editing arm, which preserves it at 20 to 21%.

The generational pattern of Section 5.2 appears in the shorter drafting conditions. `gpt-4o-mini`, a 2024 model, writes 기여하다 into 77% of its abstracts, 강조하다 into 35% and 탐구하다 into 33%, which is the group that rose in 2024 and receded by 2026; the two 2026 models write those words at 1 to 6% and instead produce 시사하다, 규명하다 and 검토하다. Since the receding group was selected as the 2024 maximum, this is consistent with a generation effect but does not by itself establish one (Section 6.2).

The editing condition, in which the model is given a real 2019 abstract and asked to polish it without changing the content, moves the vocabulary far less than drafting does. Against the same 100 originals, 시사하다 goes from 9% to 10 and 11%, 확인되다 from 3% to 9 and 16%, and 알아보다 falls from 9% to 2 and 3%, while 살펴보다 is preserved at 20 and 21% against 19% in the originals. Light editing therefore reproduces the direction of the observed change but only a fraction of its size; the pattern seen in 2026 abstracts is closer to what drafting or heavy rewriting produces.

**A Korean open-weight model.** The matched-length arm repeated with EXAONE 3.5 7.8B, a Korean model released in December 2024 and run on our own hardware (Section 3.3), removes the objection that the control rests on one provider, and it adds a second data point on model generation. On the words that carry the 2026 excess the Korean model agrees with the 2026 OpenAI model: 시사하다 in 27% of its abstracts (27% for the OpenAI matched-length arm, 22% in real 2026 abstracts), 확인되다 20%, 단순하다 12%, 넘다 30%, 핵심 20%, and it leaves the falling words alone (알아보다 0%, 도움이 되다 2%, 많이 2%, 진행하다 5%), with 살펴보다 at 12%, the real 2026 rate. On the words that peaked in 2024 and receded it looks like the 2024 OpenAI model, only more so: 기여하다 in 72% of its abstracts (77% for `gpt-4o-mini`, 6% for the 2026 OpenAI model, 8% in real 2026 text), 탐구하다 72% (33%, 2%, 3%), 강조하다 53% (35%, 4%, 6%), 중요한 역할 27% (11%, 1%, 1%) and 특히 in 88%. Two of the 2026 verbs it does not write, 규명하다 (2%) and 검토하다 (5%), and 구조적 it writes into 3% of its abstracts, the same 2 to 6% band as every OpenAI condition. Its drafts are 770 characters long against 690 for the OpenAI arm, so its rates carry a slight upward bias that does not affect these contrasts. Asked only to polish, the Korean model moves the vocabulary further than the OpenAI editors did: against its own 60 originals, 확인되다 goes from 3% to 28%, 규명하다 from 2% to 17%, 특히 from 13% to 50%, 탐구하다 from 0% to 22% and 강조하다 from 3% to 20%, while 살펴보다 falls from 17% to 7% and 알아보다 from 7% to 0%; the edits are also 11% shorter than the originals (566 against 637 characters), so part of the loss of the plain words is compression. The 2024 group again enters through polishing as it does through drafting. A model trained in Korea in 2024 therefore reproduces the 2024 marker set and a 2024-vintage model from a different provider does the same, which is what Section 6.2 needs: the turnover of markers between 2024 and 2026 follows the generation of the model rather than its maker in the five models tested.

**A third provider.** Claude Sonnet 5, a 2026 model from Anthropic run at matched length on the same 100 titles (Section 3.3), writes the 2026 group at or above the real 2026 rates: 시사하다 in 41% of its abstracts (22% in real 2026 text), 규명하다 35% (14%), 확인되다 33% (21%), 단순하다 20% (12%) and 넘다 13% (12%). The receding 2024 group it writes at 6 to 20% (기여하다 20%, 강조하다 12%, 탐구하다 6%), between the 2026 OpenAI model (1 to 6%) and the two 2024-generation models (33 to 77%), and 구조적 at 6%. It differs from the OpenAI model in keeping 살펴보다, a falling word, at 18% (4% for the OpenAI arm, 11% in real 2026 abstracts). Asked only to polish, it behaves like the OpenAI editors: against the same originals, 시사하다 goes from 9% to 16%, 규명하다 from 3% to 10%, 확인되다 from 3% to 9%, and 살펴보다 is preserved at 25%.

**Rewriting a draft.** The condition between polishing and drafting, in which the model receives the real 2019 abstract as a rough draft and is asked to rewrite it to submission standard, produces the two headline words at or above their real 2026 rates and the rest of the 2026 group below them. Against the same 100 originals (시사하다 9%, 확인되다 3%, 규명하다 3%, 이러한 결과 8%), the rewritten versions contain 시사하다 in 23% (OpenAI) and 27% (Claude) of abstracts, 확인되다 in 25% and 35%, 규명하다 in 21% and 32% and 이러한 결과 in 25% and 25%, at or above the real 2026 rates of 22, 21, 14 and 12%; but 단순하다 stays at 5 to 7% (12% in real 2026 text), 넘다 at 5 to 7% (12%) and 구조적 at 2 to 4% (15%), and the receding 2024 group is not written (기여하다 12%, 강조하다 2%, 탐구하다 1 to 2%). The falling words go down without disappearing: 살펴보다 from 19% in the originals to 7% (OpenAI) and 16% (Claude), 알아보다 from 9% to 0%, 진행하다 from 8% to 2 and 5%. Rewriting therefore moves a text about halfway from the 2019 to the 2026 register on the frequent markers and much less on the rare ones, at unchanged length (635 and 687 characters against 630 for the originals). EXAONE, rewriting the first 60 originals, writes 시사하다 into 10%, 확인되다 into 27% and 규명하다 into 12% of them, brings back the receding 2024 group of its own generation (기여하다 28%, 강조하다 32%, 탐구하다 43%) and keeps 살펴보다 at 7%, so the rewriting condition sorts the receding group by model generation as the drafting condition did.

One marker resists every condition. 구조적 "structural" rose from 2% of real abstracts in 2019 to 15% in 2026, but the models produce it in only 2 to 6% of their output in all sixteen conditions, the Korean and the Anthropic model included. Whatever drives that word is not captured by these five models under these prompts, and Section 6 returns to it.

**Table 7:** Positive control. Share of abstracts (%) containing each word in real KCI abstracts (random samples of 3,000 per year), in the 100 real 2019 abstracts that the editing control was run on, and in model-written abstracts (100 per condition, 60 for the EXAONE conditions). Conditions: *draft* writes an abstract from a 2019 title (300–400 characters requested); *long* is the same instruction with the real length range (600–700 characters); *edit* asks the model to polish a real 2019 abstract without changing its content; *rewrite* presents the same real 2019 abstract as a rough draft and asks for a rewrite to submission standard, keeping the content and results but free to restructure sentences and change expressions. OpenAI models were called through their API with default sampling; EXAONE 3.5 7.8B is a Korean open-weight model run locally through llama.cpp (llama-cpp-python 0.3.35, Q4_K_M quantisation, temperature 0.7, top-p 0.95, fixed seed) on the first 60 titles of the matched-length sample and the first 60 of the editing and rewriting originals, so its columns rest on 60 abstracts each; its drafting output was capped at 430 tokens and stripped of headings, which still leaves it 12% longer than the OpenAI matched-length arm. Claude Sonnet 5 (Anthropic) was run on the same 100 editing and rewriting originals and the same 100 matched-length titles through the Claude Code command-line client (version 2.1.235, non-interactive mode, no tools, a one-line neutral system prompt, default sampling, extended thinking off) on 26 August 2026. Document frequency grows with text length, so the last row gives mean length and rates should be compared within comparable lengths. Rates are percentages of 100 (or 60) abstracts: a 95% Wilson interval is about ±6 pp at 10%, ±9 pp at 30% and ±10 pp at 50% for n = 100, and ±8, ±12 and ±13 pp for n = 60, so differences smaller than that between model columns are not resolved; the real-year columns rest on 3,000 abstracts and have intervals of about ±1.5 pp.

| Word | Real 2019 | Real 2025 | Real 2026 | Paired 2019 | draft gpt-4o-mini | draft gpt-5.6-luna | draft gpt-5.6-terra | edit Claude Sonnet 5 | edit EXAONE 7.8B | edit gpt-4o-mini | edit gpt-5.6-luna | en2ko Claude Sonnet 5 | en2ko EXAONE 7.8B | en2ko gpt-5.6-luna | long Claude Sonnet 5 | long EXAONE 7.8B | long gpt-5.6-luna | re-write Claude Sonnet 5 | rewrite EXAONE 7.8B | re-write gpt-5.6-luna |
|---|---|---|---|---|---|---|---|---|---|---|---|---|---|---|---|---|---|---|---|---|
| 시사하다 *suggest, imply* | 5 | 13 | 22 | 9 | 16 | 19 | 14 | 16 | 13 | 10 | 11 | 12 | 12 | 8 | 41 | 27 | 27 | 27 | 10 | 23 |
| 규명하다 *elucidate* | 4 | 8 | 14 | 3 | 13 | 20 | 14 | 10 | 17 | 4 | 10 | 21 | 10 | 22 | 35 | 2 | 21 | 32 | 12 | 21 |
| 확인되다 *be confirmed* | 6 | 13 | 21 | 3 | 11 | 21 | 8 | 9 | 28 | 16 | 9 | 10 | 14 | 14 | 33 | 20 | 41 | 35 | 27 | 25 |
| 기여하다 *contribute* | 4 | 12 | 8 | 6 | 77 | 5 | 6 | 8 | 12 | 8 | 8 | 12 | 17 | 15 | 20 | 72 | 6 | 12 | 28 | 12 |
| 단순하다 *mere, simple* | 2 | 8 | 12 | 1 | 11 | 9 | 7 | 2 | 5 | 2 | 1 | 1 | 3 | 1 | 20 | 12 | 31 | 7 | 5 | 5 |
| 넘다 *go beyond* | 3 | 8 | 12 | 2 | 6 | 7 | 6 | 2 | 2 | 0 | 1 | 0 | 3 | 4 | 13 | 30 | 18 | 5 | 8 | 7 |
| 기능하다 *function as* | 1 | 3 | 7 | 0 | 6 | 6 | 2 | 1 | 0 | 2 | 1 | 1 | 0 | 2 | 7 | 0 | 8 | 5 | 0 | 6 |
| 통합하다 *integrate* | 1 | 3 | 5 | 1 | 3 | 1 | 2 | 1 | 2 | 0 | 1 | 1 | 7 | 3 | 1 | 10 | 6 | 1 | 3 | 0 |
| 결합하다 *combine* | 2 | 4 | 8 | 4 | 1 | 7 | 7 | 4 | 3 | 4 | 5 | 7 | 3 | 4 | 3 | 5 | 17 | 4 | 3 | 7 |
| 작동하다 *operate* | 1 | 2 | 6 | 4 | 0 | 1 | 1 | 4 | 0 | 3 | 4 | 2 | 0 | 1 | 2 | 0 | 3 | 6 | 0 | 2 |
| 조명하다 *shed light on* | 1 | 4 | 3 | 3 | 17 | 0 | 1 | 4 | 8 | 3 | 3 | 4 | 3 | 6 | 8 | 10 | 7 | 10 | 3 | 4 |
| 촉진하다 *promote* | 1 | 3 | 4 | 2 | 4 | 4 | 1 | 4 | 12 | 4 | 6 | 4 | 0 | 2 | 1 | 27 | 9 | 2 | 13 | 8 |
| 강화하다 *strengthen* | 4 | 8 | 7 | 3 | 8 | 10 | 8 | 5 | 5 | 4 | 5 | 3 | 5 | 5 | 12 | 30 | 20 | 6 | 13 | 7 |
| 강조하다 *emphasize* | 4 | 9 | 6 | 2 | 35 | 3 | 2 | 2 | 20 | 5 | 3 | 3 | 10 | 3 | 12 | 53 | 4 | 2 | 32 | 2 |
| 탐구하다 *explore* | 2 | 5 | 3 | 0 | 33 | 3 | 1 | 0 | 22 | 9 | 1 | 4 | 19 | 1 | 6 | 72 | 2 | 1 | 43 | 2 |

| Word | Real 2019 | Real 2025 | Real 2026 | Paired 2019 | draft gpt-4o-mini | draft gpt-5.6-luna | draft gpt-5.6-terra | edit Claude Sonnet 5 | edit EXAONE 7.8B | edit gpt-4o-mini | edit gpt-5.6-luna | en2ko Claude Sonnet 5 | en2ko EXAONE 7.8B | en2ko gpt-5.6-luna | long Claude Sonnet 5 | long EXAONE 7.8B | long gpt-5.6-luna | re-write Claude Sonnet 5 | rewrite EXAONE 7.8B | re-write gpt-5.6-luna |
|---|---|---|---|---|---|---|---|---|---|---|---|---|---|---|---|---|---|---|---|---|
| 작용하다 *act on* | 4 | 7 | 8 | 2 | 10 | 4 | 5 | 4 | 3 | 3 | 3 | 2 | 8 | 4 | 13 | 25 | 8 | 10 | 7 | 8 |
| 검토하다 *examine, review* | 9 | 10 | 16 | 6 | 12 | 36 | 38 | 11 | 5 | 8 | 12 | 12 | 7 | 19 | 26 | 5 | 29 | 15 | 5 | 25 |
| 구조적 *structural* | 3 | 6 | 15 | 3 | 2 | 6 | 3 | 3 | 0 | 3 | 3 | 2 | 2 | 3 | 6 | 3 | 2 | 4 | 5 | 2 |
| 핵심적 *core, key* | 1 | 2 | 3 | 1 | 2 | 0 | 1 | 1 | 3 | 3 | 1 | 2 | 3 | 1 | 3 | 17 | 1 | 9 | 3 | 4 |
| 실질적 *substan-tive* | 3 | 7 | 6 | 1 | 9 | 6 | 4 | 2 | 17 | 2 | 1 | 2 | 17 | 1 | 16 | 33 | 5 | 11 | 17 | 5 |
| 핵심 *core, key* | 4 | 12 | 19 | 7 | 2 | 11 | 10 | 9 | 20 | 6 | 7 | 8 | 12 | 10 | 8 | 20 | 21 | 15 | 25 | 11 |
| 메커니즘 *mecha-nism* | 1 | 2 | 5 | 1 | 2 | 2 | 1 | 1 | 2 | 1 | 1 | 2 | 3 | 2 | 4 | 2 | 3 | 1 | 3 | 1 |
| 기제 *mecha-nism (Sino-Korean)* | 1 | 1 | 4 | 2 | 0 | 0 | 0 | 2 | 2 | 2 | 2 | 1 | 0 | 1 | 3 | 0 | 0 | 4 | 3 | 2 |
| 특히 *in particular* | 17 | 25 | 25 | 13 | 43 | 22 | 15 | 14 | 50 | 14 | 13 | 14 | 31 | 17 | 37 | 88 | 47 | 19 | 57 | 27 |
| 이러한 결과 *these results* | 4 | 8 | 12 | 8 | 18 | 19 | 0 | 14 | 8 | 15 | 13 | 9 | 7 | 13 | 31 | 8 | 22 | 25 | 13 | 25 |
| 중요한 역할 *impor-tant role* | 1 | 2 | 1 | 1 | 11 | 1 | 0 | 1 | 2 | 1 | 1 | 2 | 5 | 3 | 6 | 27 | 1 | 1 | 3 | 0 |
| 보여주다 *show* | 8 | 13 | 16 | 4 | 6 | 1 | 3 | 6 | 8 | 7 | 7 | 7 | 7 | 8 | 6 | 17 | 9 | 11 | 22 | 9 |
| 알아보다 *look into* | 7 | 3 | 1 | 9 | 0 | 1 | 0 | 2 | 0 | 2 | 3 | 2 | 0 | 0 | 0 | 0 | 0 | 0 | 0 | 0 |
| 살펴보다 *take a look at* | 20 | 16 | 11 | 19 | 23 | 3 | 0 | 25 | 7 | 20 | 21 | 14 | 8 | 7 | 18 | 12 | 4 | 16 | 7 | 7 |
| 많이 *a lot* | 5 | 3 | 1 | 3 | 1 | 0 | 0 | 0 | 0 | 2 | 2 | 0 | 0 | 0 | 0 | 2 | 0 | 0 | 0 | 0 |
| 도움이 되다 *be helpful* | 2 | 2 | 1 | 1 | 0 | 0 | 0 | 1 | 0 | 1 | 1 | 2 | 0 | 2 | 0 | 2 | 0 | 1 | 0 | 1 |
| 진행하다 *carry out* | 7 | 7 | 4 | 8 | 4 | 1 | 0 | 8 | 12 | 11 | 5 | 4 | 5 | 3 | 2 | 5 | 0 | 5 | 17 | 2 |
| ~에 관한 연구 *a study on* | 2 | 1 | 1 | 2 | 0 | 0 | 0 | 1 | 0 | 1 | 2 | 1 | 0 | 3 | 0 | 0 | 0 | 0 | 0 | 1 |
| *mean length (charac-ters)* | *682* | *681* | *693* | *630* | *509* | *420* | *313* | *643* | *566* | *634* | *640* | *592* | *588* | *579* | *632* | *770* | *690* | *687* | *570* | *635* |

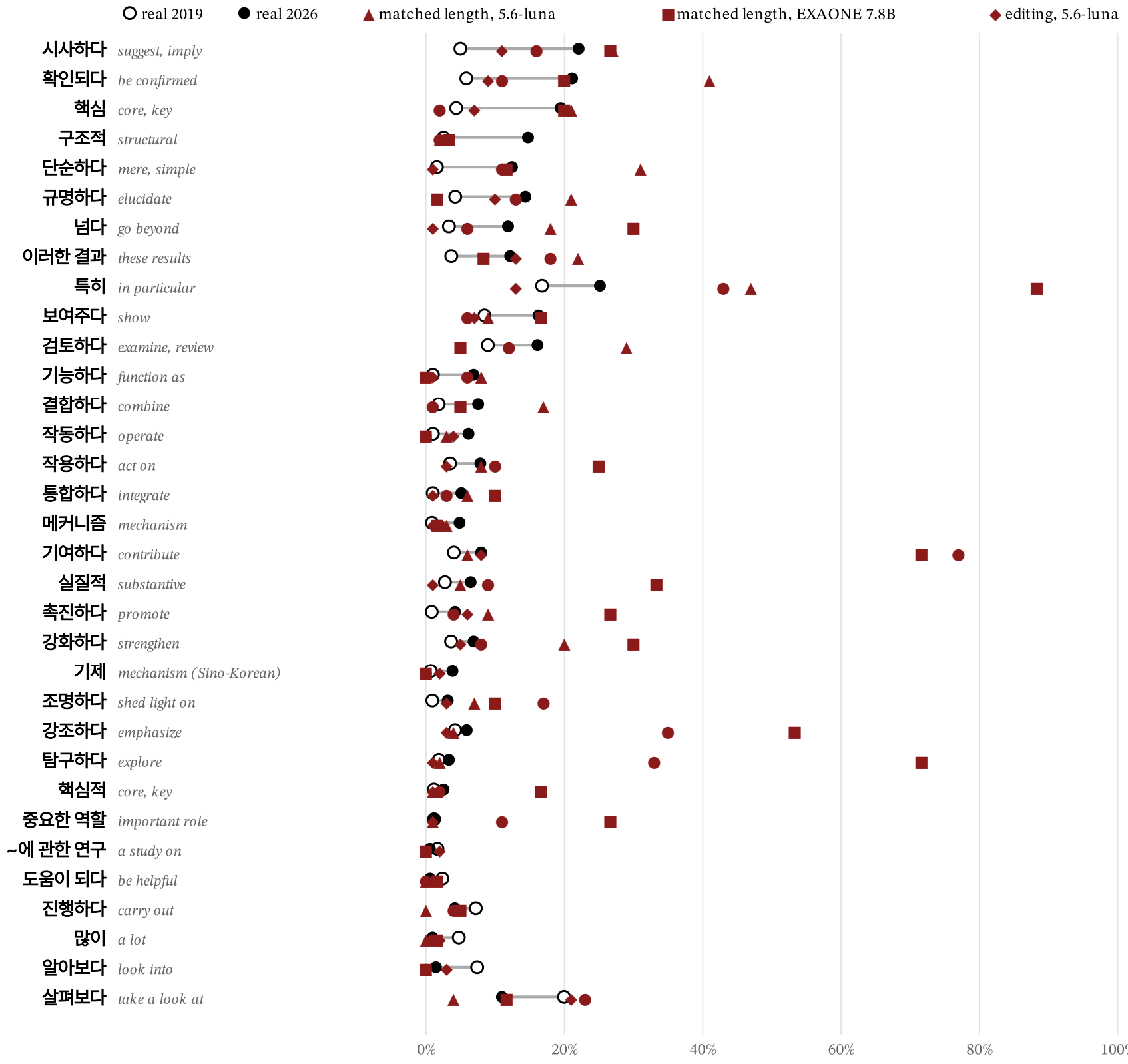

**Figure 5:** Positive control. For each marker, the share of real abstracts containing it in 2019 (hollow) and 2026 (filled), and the share of model-written abstracts containing it in four of the sixteen generation conditions (red symbols): the 2026 OpenAI model drafting at matched length, the Korean open-weight model EXAONE 3.5 7.8B drafting at matched length, the 2026 OpenAI model editing a real 2019 abstract, and the 2024 OpenAI model drafting freely. Markers are ordered by the 2019 to 2026 change.

### 5.7 External comparison: Vietnamese

Table 8 gives the Vietnamese results for 2025 on the VJOL journal panel. The corpus is about one eighth of the Korean one and its composition moved more, and both facts show. Many of the top units are topic words that the coarse part-of-speech split lets through (giảng dạy "teaching", chuyển đổi số "digital transformation", trí tuệ nhân tạo "artificial intelligence", khoa học công nghệ "science and technology", chế tạo "manufacture"), reflecting a growth of education and technology journals on the platform. Setting those aside, a register group remains that mirrors the Korean and English lists: nhấn mạnh "emphasize" (0.4% expected, 4.0% observed, 11.3×, +3.7 pp), tích hợp "integrate" (4.1×), tối ưu hóa "optimize" (9.5×), toàn diện "comprehensive" (2.4×), linh hoạt "flexible" (3.6×), thách thức "challenge" (2.9×), phản ánh "reflect" (2.8×), mở ra "open up" (4.4×), and among nouns then chốt "key, pivotal" (8.3×), kỷ nguyên "era" (4.4×) and cơ chế "mechanism" (2.7×); the nouns come from the content-word list of the same run, since Table 8 lists verbs, adjectives, adverbs and bigrams only. In 2024 3 of these words are already above trend (nhấn mạnh 3.7×, tối ưu hóa 4.9×, toàn diện 1.6×, with 1 to 2 pp of excess) and the rest are at or below it, and in 2023 none is above trend, so the Vietnamese onset is 2025, one year after the Korean one and two after the English one. The 2022 negative control on this corpus is far noisier than the Korean one: its largest style-word excess is 9.1 pp on the bigram bài viết "this article" (1.5×) and 9.0 pp on viết "write", with ratios up to 4.0× on lower-frequency units. In percentage points, therefore, the register words of 2025 do not clear the control floor: nhấn mạnh reaches 3.7 pp against a floor of 9.1 pp, and no unit in the 2025 list exceeds that floor (the largest excess is 3.7 pp). What is outside the control range is the *ratio* of nhấn mạnh, 11.3× against a control maximum of 4.0×. We therefore report no prevalence bound for Vietnamese. What the corpus supports is the weaker but still informative statement that the register group appears in 2025, not before, and that it is the same group of concepts found in Korean and English.

**Table 8:** Vietnamese words tagged as verbs (V), adjectives (A) or adverbs (R), and adjacent bigrams containing one, with the largest excess in 2025 on the VJOL journal panel (k = 5, n = 5,411), expectation from the 2019–2022 trend. Share of abstracts (%). Only units whose 2025 share exceeds their highest base-year share are listed, so that no listed unit merely regained ground lost during the base years; where the base-year trend fell, the expectation lies below every base-year value shown and the ratio still overstates the rise. Topic words (teaching, digital transformation, artificial intelligence) pass the coarse part-of-speech split and are marked in the text. In the 2022 negative control (trend from 2019–2021) the largest style-word excess is 9.1 pp at 1.5×, and the largest ratio is 4.0×.

| Word | POS | 2019 | 2021 | 2023 | 2024 | 2025 | Exp. 2025 | Ratio | Excess (pp) |
|---|---|---|---|---|---|---|---|---|---|
| **nhấn mạnh** *emphasize* | V | 0.8 | 0.5 | 0.9 | 1.6 | 4.0 | 0.4 | 11.3 | +3.7 |
| **giảng dạy** | V | 2.9 | 2.5 | 3.2 | 4.4 | 6.2 | 2.7 | 2.3 | +3.6 |
| **tích hợp** *integrate* | V | 1.1 | 1.4 | 1.0 | 1.8 | 4.6 | 1.1 | 4.1 | +3.4 |
| **chuyển đổi + số** *digital transformation* | bigram | 0.0 | 0.4 | 1.6 | 1.6 | 5.2 | 1.8 | 2.9 | +3.4 |
| **tối ưu hóa** | V | 1.1 | 0.7 | 0.9 | 2.1 | 3.6 | 0.3 | 9.5 | +3.3 |
| **toàn diện** *comprehensive* | A | 2.7 | 2.0 | 2.9 | 3.5 | 5.3 | 2.2 | 2.4 | +3.1 |
| **nhân tạo** | A | 0.9 | 0.9 | 1.2 | 2.1 | 4.1 | 1.1 | 3.7 | +3.0 |
| **mở** | V | 1.7 | 1.7 | 1.9 | 2.1 | 4.3 | 1.6 | 2.7 | +2.7 |
| **quốc tế** | A | 4.4 | 2.6 | 3.3 | 3.2 | 4.6 | 2.0 | 2.3 | +2.6 |
| **trí tuệ + nhân tạo** *artificial intelligence* | bigram | 0.3 | 0.3 | 0.6 | 1.4 | 2.9 | 0.4 | 7.6 | +2.5 |
| **hợp tác** | V | 1.4 | 1.4 | 1.5 | 1.7 | 3.5 | 1.0 | 3.5 | +2.5 |
| **nghiên cứu + tập trung** | bigram | 0.8 | 0.8 | 1.0 | 1.6 | 3.2 | 0.7 | 4.5 | +2.5 |
| **thách thức** | V | 1.3 | 1.1 | 1.5 | 2.1 | 3.7 | 1.3 | 2.9 | +2.4 |

| Word | POS | 2019 | 2021 | 2023 | 2024 | 2025 | Exp. 2025 | Ratio | Excess (pp) |
|---|---|---|---|---|---|---|---|---|---|
| **giao tiếp** | V | 1.5 | 0.9 | 1.3 | 1.3 | 2.7 | 0.3 | 4.5 | +2.4 |
| **viết + đề xuất** | bigram | 1.5 | 1.2 | 1.5 | 1.7 | 3.0 | 0.9 | 3.3 | +2.1 |
| **mở + ra** | bigram | 0.8 | 0.7 | 0.8 | 0.9 | 2.7 | 0.6 | 4.4 | +2.1 |
| **phản ánh** | V | 1.7 | 1.4 | 1.7 | 1.3 | 3.3 | 1.1 | 2.8 | +2.1 |
| **cắt** | V | 3.5 | 2.2 | 3.9 | 3.8 | 3.6 | 1.6 | 2.3 | +2.0 |
| **công nghệ + số** | bigram | 0.1 | 0.1 | 0.6 | 0.5 | 2.6 | 0.6 | 4.4 | +2.0 |
| **bối cảnh + chuyển đổi** | bigram | 0.0 | 0.0 | 0.2 | 0.3 | 2.0 | 0.1 | 15.0 | +1.9 |

### 5.8 Robustness

The design has three exposures: the linear extrapolation, the journal panel, and the fact that the reported bound is a maximum over many noisy estimates. We test each.

**Placebo target years.** Running the whole pipeline with a pre-LLM year as the target measures what the method produces when the answer is known to be zero. One year ahead, the largest style-word excess is 1.5 pp at 1.02× with 2020 as target (base 2018–2019), 1.5 pp at 1.03× with 2021 (base 2018–2020) and 0.7 pp at 1.01× with 2022 (base 2018–2021) (Table 2), and on the main frequency table the same one-year statistic returns 0.7 to 1.4 pp (Table 10). The words attaining these maxima are high-frequency function verbs (위하다 "for" and 있다 "be") whose share drifts by a few percent of itself between years; at a base share of 50%, a ratio of 1.04 is already 2 pp. The 2023 value of 1.0 pp sits inside this band and we do not interpret it.

**Extrapolation horizon.** The placebo above extrapolates one year beyond the last base year, whereas the 2026 estimate extrapolates four, and extrapolation error grows with distance. Table 10 therefore repeats the placebo at every horizon the pre-LLM data allow, from one to four years ahead and with two to five base years, every cell computed on the completed corpus, the rows with pre-2018 base years using its extended store (Appendix H). Searched over all style units, the floor grows with the horizon and shrinks with the number of base years: with two base years it is 0.9–1.4 pp at one year, 1.4–3.7 at two, 3.1–5.5 at three and 6.3 at four, and it is almost always a high-share function word that attains it (같다 at a ratio of 1.34 gives the 6.3), the one exception being the fused noun 가능성 in a single cell; with three base years it is 0.9–1.2 pp at one year, 1.2–2.2 at two and 2.5 at three; with four base years 0.7–1.2 at one year and 1.3 at two; and with five base years, the length of the main fit, the one-year floor is 0.9 pp. That is not what a marker of model text looks like, and a ratio requirement removes it: restricted to units whose excess ratio is at least 1.5, the floors over all base lengths are 0.1–0.5 pp at one year, 0.4–1.5 at two, 0.6–1.8 at three and 2.2 at four (two base years), while the bounds for the real target years become 0.1 pp in 2023, 3.5 in 2024 (기여하다 "contribute"), 10.5 in 2025 (특히 "in particular") and 16.1 in 2026 (시사하다, whose ratio is 4.0). We use the ratio-restricted statistic for the year-by-year comparison of Table 9; it is the same kind of restriction Kobak et al. apply when they define excess words. Under it, 2024 is 2.3 times the largest floor at its horizon, 2025 5.8 times, and 2026 7.3 times the measured four-year floor, which was obtained from the least favourable fit the data allow (two base years). The set statistic of Section 5.9 was run on the same grid (last two columns of Table 10). With two base years its placebo value lies between −0.8 and +2.2 pp at one to three years ahead and reaches 2.9 pp at four (2017–2018 fitted, 2022 measured, the rare set; 2.1 pp for the top-k set); with three base years it is 0.6 pp on the one configuration where any word passes selection, and with four or five base years no word passes the selection criteria on half A, so the set is empty and the statistic is zero by construction. These columns admit bigrams, which enlarges the candidate inventory; the rare-lemma set statistic behind the bound column of Table 9 has smaller floors, at most 1.5 pp at any horizon and −1.2 pp at four years ahead (Table A10, whose top-k column is higher at three years ahead). The set bounds of 7.8, 20.6 and 33.0 pp for 2024–2026 are 4.3, 7.1 and 11.4 times the largest set floor at their own horizon, the largest chosen-cell placebo value across every set construction of Tables 9, 10 and A10 at that horizon (1.8, 2.9 and 2.9 pp; each multiple divides the rounded values shown). The floors are point values from single placebo constructions and carry sampling error of their own, so the multiples are descriptive, not test statistics. The exact main configuration, five base years and four years ahead, has no pre-LLM counterpart, because a 2022 target would need base years from 2014 and the corpus begins in 2017; the grid brackets it between two base years at four years ahead and five base years at one. The unrestricted 2024 and 2025 values, 4.8 and 11.0 pp, are attained by common predicates with ratios of 1.33 and 1.27 and sit within about a factor of two of the matched two-base-year floors, which is why they are not the headline. The rule of Kobak et al., which adds the last one-year change to the last base value once for every year of horizon and never extrapolates a fall, has floors of at most 3.3 pp at any horizon (0.7 pp with the ratio rule) and gives 4.4, 10.8 and 16.8 pp for 2024–2026; a constant expectation equal to the base mean, which involves no extrapolation at all, gives 7.5, 15.3 and 19.5. The 2026 value is the same or larger under every alternative, because the least-squares line credits 시사하다 with a small upward pre-2023 trend that the other rules do not.

**Breadth across journals.** The rise is not carried by a few journals. Among the 482 journals with at least 20 abstracts in both the base period and 2026, the share of abstracts containing 시사하다 rose in 91% (median journal ratio 5.9, interquartile range 3.0 to 13.3), 확인되다 in 91%, 구조적 in 95%, 규명하다 in 91% and 단순하다 in 88%, while 살펴보다 fell in 94% and 알아보다 in 88%. The tenth of journals with the largest excess accounts for 41 to 46% of the total excess of each rising word, so the change is uneven but general: most of the excess lies outside the journals that changed most.

**Composition weighting.** Weighting every journal equally instead of every document, which removes any effect of journals contributing different numbers of abstracts in different years, changes the 2026 excess ratio of every rising marker by less than 11% in either direction and moves the falling markers slightly toward 1: 시사하다 3.3× against 3.7×, 확인되다 3.1 against 3.1, 규명하다 4.2 against 3.9, 단순하다 9.6 against 9.5, 넘다 3.6 against 4.1, 핵심 3.0 against 3.0, and on the falling side 살펴보다 0.49 against 0.44 and 알아보다 0.30 against 0.26. Restricting to the strict panel of 1,238 journals that appear in every year from 2018 to 2026 gives 시사하다 4.7× (+17.1 pp), 확인되다 2.7× (+13.3), 규명하다 3.4× (+11.1), 단순하다 10.6× (+10.6), against 3.7×, 3.1×, 3.9×, 9.5× on the same sample of the loose panel. Journal composition is not driving the result.

**Within-year timing.** Splitting 2026 into January to April and May to August gives 시사하다 21.6% against 21.7%, 확인되다 20.7 against 22.6, 규명하다 16.4 against 14.3, 단순하다 11.7 against 13.1, 핵심 19.3 against 17.8. The year is internally stable, so the 2026 figure is not an artefact of which months have been indexed so far. Figure 6 gives the monthly series, which is the sharper view because Korean journals publish on fixed months and the month composition of a year is therefore also a journal composition: 2021 in the first harvest was 94% January to March, which the completing harvest evens out, 2022 is weighted to the second half, and August 2026 is only partly indexed (329 abstracts in the panel against a monthly average near 4,600) and is excluded from the figure. The monthly curve for 시사하다 moves between 3.4 and 7.5% from 2018 through mid-2024 with no trend, turns in the second half of 2024 (6.2% in August, 8.4% in December), rises in every quarter of 2025 (8.5% in January to 16.4% in December, with one dip in September), reaches 23.4% in April 2026 and then holds between 20 and 23% through July. 살펴보다 falls over the same stretch from 22% to 9%, with small reversals inside 2024. The rise is therefore not still accelerating in 2026; it flattens in the second quarter at about four times its pre-2023 level.

Restricting every year to January–July and weighting each month equally, which removes the composition difference entirely, leaves the year values almost unchanged: 시사하다 21.0% against 21.5% in 2026, 11.7 against 12.0 in 2025, 6.1 against 6.1 in 2024, and 3.9 against 4.3 in 2019; 구조적 14.8 against 15.1 in 2026; 살펴보다 9.7 against 9.9.

**Density normalisation.** The rise in distinct units per abstract documented above (Appendix B) inflates every document frequency mechanically, and a word whose expected share is large absorbs a large part of that inflation in percentage points: for 분석하다 "analyse", expected in 41% of 2025 abstracts, a 4.9% rise in density alone is worth about 2.0 pp, roughly a fifth of the unrestricted 2025 bound that this verb attains. Rather than correct individual words after the fact, we re-ran the whole excess calculation on frequencies divided by each year's density and multiplied by the base-year mean (87.5 units), which scales 2025 by 0.953 and 2026 by 0.895 and leaves the base years within 0.4% of themselves. Table 12 gives the bounds and Table A3 the word-level excess. The normalised bounds are 13.8 pp in 2026 (시사하다, ratio 3.61 instead of 4.04), 9.3 in 2025 (now attained by 특히 under both statistics, since 분석하다 falls from 11.0 to 8.5 pp), 4.6 and 3.4 in 2024, and in 2023 unchanged under the ratio restriction (Table 12); the falling words fall slightly further (살펴보다 0.42 to 0.38, Table A3). The normalisation is conservative: part of the density rise is the very phenomenon under study, since a rewriting pass that varies wording raises the count of distinct units, and dividing it out removes some signal along with the shared component. The figures we quote elsewhere are the unnormalised ones, and Table 12 is the amount by which a reader who prefers the conservative reading should discount them: 14% of the 2026 bound and 15% of the unrestricted 2025 bound.

**Selection in the bound.** Because the bound is a maximum, a bootstrap interval attached to the winning word after the fact understates its uncertainty, and because abstracts from one journal are not independent, a document-level resample understates it further. Resampling whole journals with replacement and recomputing the maximum inside every replicate, over 27 tracked markers and 12,000 abstracts per year, gives 1.5 pp (95% CI 0.9 to 2.6) for 2023, 4.6 (3.6 to 5.9) for 2024, 11.0 (9.7 to 12.5) for 2025 and 15.9 (14.6 to 17.3) for 2026; with the ratio restriction of the horizon paragraph applied inside every replicate it gives 3.5 (2.9 to 4.3) for 2024, 11.0 (9.4 to 12.5) for 2025, 15.9 (14.7 to 17.1) for 2026 (Table 9). The same computation with abstracts rather than journals resampled gives intervals about 19% narrower, which is the measure of how much the journal clustering matters. Only the 2023 value sits inside its matched-horizon placebo floor.

**Substitution within documents.** A general drift in Korean academic style need not pair rising and falling words within one abstract. If instead a single pass, human or machine, rewrites an abstract into the newer register, then within one abstract the presence of a rising word should predict the absence of a falling one. In 20,000 abstracts per year: in 2019 an abstract containing at least one of five rising markers was no less likely than other abstracts to contain a falling one (알아보다 8.1% against 6.9%, 살펴보다 20.8 against 20.3, z between −0.7 and +2.5), and 2022 is the same. In 2026 the relation is strongly negative: 살펴보다 7.3% against 12.7% (z = −12.5), 알아보다 1.0 against 1.7 (z = −4.0), 도움이 되다 0.5 against 1.0 (z = −3.7), 많이 0.9 against 1.6 (z = −4.3), 진행하다 3.1 against 4.1 (z = −3.5). The two vocabularies became strongly negatively associated inside single documents in exactly the period when the aggregate curves crossed. The association is not exclusion: 7.3% of the abstracts with a rising marker still contain 살펴보다 and 3.1% still contain 진행하다. But it appeared from nothing in the year the curves crossed, and it is the pattern expected from a document-level register shift rather than a word-level drift: it is what whole-abstract rewriting produces, by a model or by an author who had adopted the register, and it is harder to obtain from words diffusing one at a time, although a community adopting a register wholesale would produce it too (Section 6.3). It does not by itself separate machine from human rewriting; Section 6.3 says what does.

**Suffix fusion.** Of the thirty largest 2026 excesses in Table 3, eleven are units that the noun + 적/화/성 fusion rule of Section 4.1 creates (구조적, 실증적, 제도적, 통합적, 체계적, 이론적, 실질적, 상대적, 구조화, 실천적, 결론적), so the rule deserves its own test. Re-running the pipeline from the raw text with fusion switched off leaves the bound unchanged in every year, to the first decimal and with the same word attaining it (1.0, 4.8, 11.0 and 16.1 pp for 2023 to 2026), because a fused unit never attains it. Nor does switching it off remove the underlying rise: without the rule the same increase reappears as excess of the corresponding content noun, 구조 at +27.1 pp instead of 구조적 +12.6 and 구조 +20.6, and 실증 at +8.6 pp instead of +0.8. The rule therefore does not create excess; it assigns it, moving a formal-register derivation out of the content noun that would otherwise absorb it. What it does change is the composition of Table 3: without it, eleven style units are replaced by as many others (결합되다, 분석 틀, 의의를 지니다, 전환되다 and so on) and the remaining nineteen keep their places in almost the same order (Appendix E).

**Selection and estimation on different journals.** Markers are chosen and measured on the same corpus, which the intervals of Section 4.5 acknowledge but do not remove. Splitting the 1,853 panel journals by a hash of their names into halves of 929 and 924, ranking markers on half A alone and then measuring on half B, gives a bound that no selection has touched: 16.3 pp for 2026, 11.2 for 2025, 5.0 for 2024, all slightly above the corresponding whole-corpus values, and 1.0 pp for 2023, at the placebo level of that horizon. Selecting and estimating inside half B gives the same three numbers to one decimal. Selection is therefore not what produces the bound (Appendix F).

**Base years.** The 2021 base year was the weakest point of the first harvest, which held 12,975 abstracts for it, 94% of them from January to March; the completed corpus of the main analysis holds 37,665 panel abstracts for it (Appendix H), and the exclusion is kept as a stress test. Refitting the trend on 2018–2020 and 2022 only moves the 2026 bound from 16.1 to 16.4 pp and the 2025 bound from 11.0 to 10.9; dropping 2018 as well gives 16.3 and 11.0. The fit does not depend on the thin year (Appendix G), and a second harvest that completed 2021 to 37,665 abstracts and added 2017–2018 leaves the 2026 bound at 16.1 pp on the same five base years and 16.4 pp on six (Appendix H).

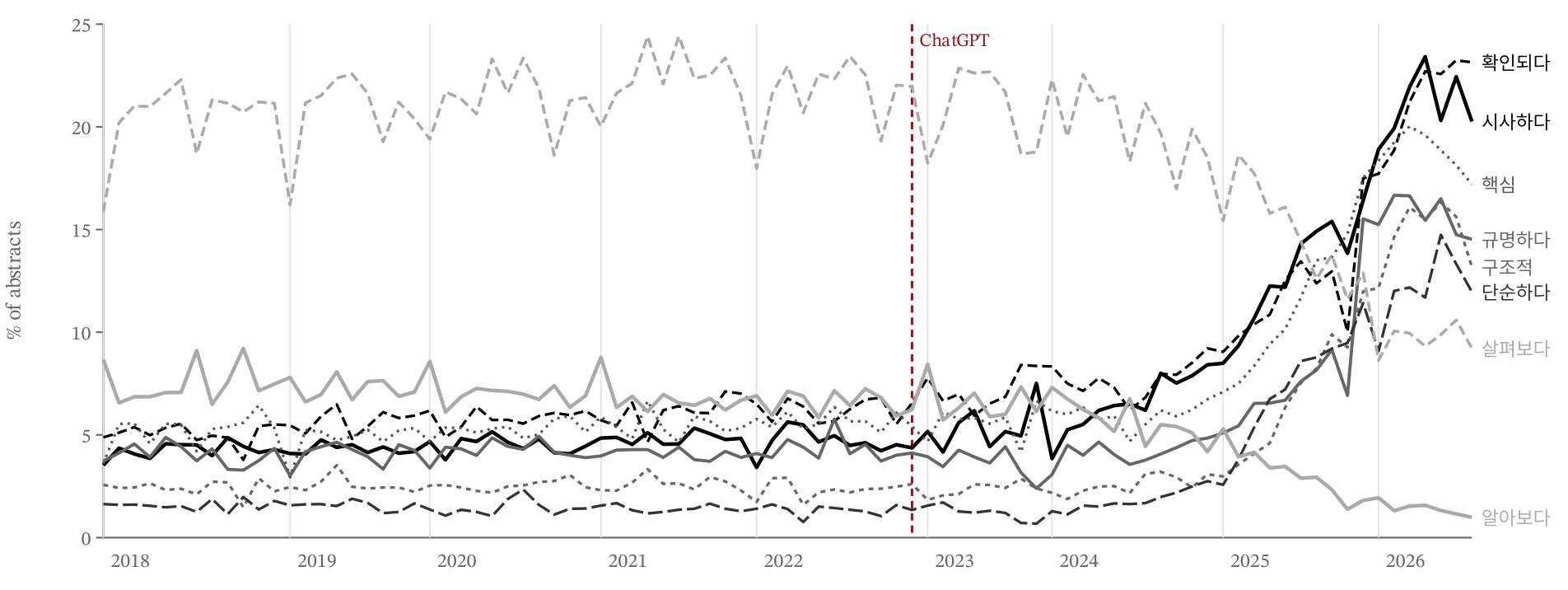

publication month (months with at least 500 abstracts in the journal panel)

**Figure 6:** Monthly series for eight markers, restricted to publication months with at least 500 abstracts in the journal panel. Korean journals publish on fixed months, so month composition varies within a year; the monthly view removes the ambiguity that yearly averages introduce when the within-year slope is steep. The dashed vertical line marks the release of ChatGPT.

### 5.9 Lower bound on LLM-processed abstracts

Table 9 and Figure 7 give the bounds by year. Two statistics are reported. The single-word bound of Section 4.7, searched over style units with an excess ratio of at least 1.5, is 0.1 pp in 2023, inside the placebo floor, 3.5 pp in 2024 (기여하다 "contribute"), 10.5 in 2025 (특히 "in particular") and 16.1 in 2026 (시사하다 "suggest", 21.4% observed against 5.3% expected). Searched over all style units the 2024 and 2025 values are 4.8 and 11.0, attained by 중요하다 "be important" and 분석하다 "analyse", two very common predicates with ratios of 1.33 and 1.27 that sit close to the horizon floor of Section 5.8, and 2026 is unchanged. Recomputing the ratio-restricted maximum inside every replicate of a bootstrap that resamples whole journals gives the intervals in Table 9; the 2026 interval is 14.6 to 17.3 for the unrestricted statistic, which 시사하다 attains either way.

The set bound is larger, as it is in English. Kobak et al. obtain their 13.5% not from a single word but from the share of abstracts containing at least one word of a set, minus that share's own trend expectation [1]. We form such sets from the excess style units, and, unlike Kobak et al., who chose the set on the corpus they measured it on, choose it on one half of the journals and measure it on the other (Section 4.7, Table 11). For 2026 the set of all clearly excess style lemmas with a base-period share below 1% (a threshold itself chosen on half A and different by year: 5% in 2024 and 2% in 2025, Table 11; because the threshold is re-chosen each year, the three bounds are not one statistic under one fixed rule, and Table A9 shows how the 2026 value moves with the threshold), 47 words listed in Table A8, admitted by the part-of-speech filter of Section 4.1 alone and with no topic judgement (Section 5.13 tests their composition directly), appears in 60.1% of the abstracts of the measurement half against 27.0% expected from the 2018–2022 trend of that share: a gap of 33.0 pp (95% CI 30.8 to 35.4 under journal resampling). The two most excess words alone, 시사하다 and 확인되다, give 34.6% against 12.2%, a gap of 22.3 pp (20.4 to 24.1); admitting bigrams to the rare set gives 36.3 pp (34.2 to 38.7) with 88 units. By year the lemma-set bound is 7.8 pp in 2024 (6 lemmas), 20.6 in 2025 (38) and 33.0 in 2026 (47); in 2023 no unit passes the excess criteria on the selection half, so there is no set and no bound. The set statistic on placebo years, searched over the larger all-unit inventory that yields the most conservative floors (Table 9), lies between −0.8 and +2.9 pp over every horizon and base length the pre-LLM data allow, the 2.9 pp at four years ahead from two base years (Table 10, Section 5.8), so the 2026 value is 11.4 times the largest set floor at its own horizon; the right panel of Figure 7 shows why the 2026 value is not an artefact of the fit: the share of abstracts of half B containing any of the 47 lemmas (the series of Figure 7) moves from 25.7% to 26.5% across the five base years, continues at 26.6% in 2023, and then breaks upward to 30.0, 42.4 and 60.1%.

The set bound is a lower bound in the same sense as the single-word one and a tighter one, because it counts every abstract that used any marker rather than only those that used the single best; it is still a floor, since an abstract processed by a model that used none of the 47 lemmas does not count. The English figure of 13.5% for 2024 is a statistic of this kind [1], obtained on a different corpus with a different base period and, unlike ours, with the set chosen on the data it was measured on. Both are floors, and floors do not order the quantities they bound; what the comparison supports is only that the Korean floor for 2026 is well above the English floor for 2024. The field bounds of Table 6, which are single-word bounds over the tracked markers, show that the aggregate hides a roughly two-to-one spread across fields. All of these are floors. The length-matched GPT condition writes its most frequent marker, 특히 "in particular", into 47% of its own abstracts and 확인되다 into 41%, and the EXAONE condition reaches 88%, so even a corpus written entirely by such a model would yield a single-word bound near that model's top rate rather than 100; the set bound closes part of that distance, and the rest remains unmeasured.

**Table 9:** Lower bound on the share of LLM-processed abstracts by year, with the noise level of each statistic measured on pre-LLM years at the same extrapolation horizon $h$ (years between the last base year and the target; Table 10 gives the full placebo grid). Single-word bounds are the largest excess in pp over style units, either over all of them or only over units whose excess ratio is at least 1.5, so that a small relative drift of a very common function verb cannot attain the maximum (a common adverb such as 특히 can still clear it, as in 2025); the cluster column recomputes the ratio-restricted maximum inside every replicate of a bootstrap that resamples whole journals (500 replicates, 27 tracked markers, 12,000 abstracts per year). The set bound is the share of abstracts containing at least one of the clearly excess style lemmas (ratio at least 1.5, excess at least 1 pp, no bigrams) chosen on half A of the journals, minus its trend expectation, both measured on half B; brackets are 95% intervals from resampling the journals of half B. Placebo floors are maxima (single-word columns) or the full range (set column) over the base-year combinations available at that horizon, including the rows of Table 10 that start from 2017 on the completed harvest of Appendix H; the $h = 4$ floors, for the single-word and the set statistic alike, come from a fit on two base years (2017–2018 fitted, 2022 measured), the least favourable configuration the data allow. The main configuration itself, five base years and a target four years ahead, cannot be run on pre-LLM data: a 2022 target would need base years from 2014, and the corpus begins in 2017. Table 10 brackets it with two base years at $h = 4$ and five base years at $h = 1$. The set floors are those of the all-unit statistic (bigrams admitted), which has the larger candidate inventory; the rare-lemma set statistic of the last column has placebo values of at most 1.5 pp at any horizon and −1.2 pp at four years ahead (Table A10; the top-k column of that table is higher at three years ahead). All values in percentage points.

| Year | h | Floor, all style | Floor, ratio ≥ 1.5 | Single word, all style | Single word, ratio ≥ 1.5 | Cluster bootstrap, ratio ≥ 1.5 [95% CI] | Set-statistic floor | Set bound, split-half [95% CI] |
|---|---|---|---|---|---|---|---|---|
| 2023 | 1 | 1.4 | 0.5 | 1.0 (및) | 0.1 (통하다 사용자) | 0.0 [0.0, 0.0] | no excess word selected | no excess lemma |
| 2024 | 2 | 3.7 | 1.5 | 4.8 (중요하다) | 3.5 (기여하다) | 3.5 [2.9, 4.3] | −0.8 to 1.8 | 7.8 [6.7, 9.0] (6 lemmas) |
| 2025 | 3 | 5.5 | 1.8 | 11.0 (분석하다) | 10.5 (특히) | 11.0 [9.4, 12.5] | 0.6 to 2.2 | 20.6 [18.5, 22.8] (38 lemmas) |
| 2026 (Jan–Aug) | 4 | 6.3 | 2.2 | 16.1 (시사하다) | 16.1 (시사하다) | 15.9 [14.7, 17.1] | 2.1 to 2.9 | 33.0 [30.8, 35.4] (47 lemmas) |

**Table 10:** Placebo grid by extrapolation horizon. Every row runs the full statistic with a pre-ChatGPT target year, for which the true LLM share is zero, from the base years shown; $h$ is the distance in years from the last base year to the target. Single-word columns give the largest excess in pp over style units under three expectation models: the five-year least-squares line of the main analysis (here fitted on the base years shown), the rule of Kobak et al. (last base value plus the last one-year change, if positive, once for every year of horizon), and a constant expectation equal to the base mean; "all" searches every style unit, "ratio ≥ 1.5" only units with excess ratio at least 1.5. The last two columns give the split-half set statistic of Section 5.9 for the same placebo, with the set chosen on half A (top-k by excess, and all rare excess words) and measured on half B; a negative value means the union fell below its trend. The 2021 target contains the pandemic-year vocabulary shift. Rows marked * start from 2017 and come from the completed harvest of Appendix H (its own journal panel and split, at most 40,000 abstracts per year); they extend the grid to four years ahead and to five base years. "no excess word" means that no unit passed the selection criteria on half A, so the set is empty and the statistic is zero by construction.

| Base years | Target | h | Linear, all | Linear, ratio ≥ 1.5 | Kobak rule, all | Kobak rule, ratio ≥ 1.5 | Constant, ratio ≥ 1.5 | Set, top-k | Set, rare |
|---|---|---|---|---|---|---|---|---|---|
| 2018, 2019 * | 2020 | 1 | 1.4 (위하다, 1.02×) | 0.4 (극복하다 위하다) | 1.2 | 0.2 | 0.2 | no excess word | no excess word |
| 2017, 2018, 2019 * | 2020 | 1 | 1.2 (위하다, 1.02×) | 0.2 (적용 가능성) | 1.2 | 0.2 | 0.2 | no excess word | no excess word |
| 2017, 2018 * | 2020 | 2 | 3.7 (같다, 1.17×) | 0.7 (~에 관한 연구) | 1.7 | 0.2 | 0.3 | −0.8 | −0.6 |
| 2019, 2020 * | 2021 | 1 | 1.1 (확인하다, 1.05×) | 0.5 (도덕적) | 1.1 | 0.2 | 0.4 | no excess word | no excess word |
| 2018, 2019, 2020 * | 2021 | 1 | 1.0 (이러하다, 1.04×) | 0.3 (거리 두다) | 1.1 | 0.2 | 0.4 | no excess word | no excess word |
| 2017, 2018, 2019, 2020 * | 2021 | 1 | 1.2 (인하다, 1.09×) | 0.4 (거리 두다) | 1.0 | 0.2 | 0.4 | no excess word | no excess word |
| 2018, 2019 * | 2021 | 2 | 2.7 (위하다, 1.04×) | 1.5 (대응하다) | 2.3 | 0.5 | 0.5 | 1.3 | 1.8 |
| 2017, 2018, 2019 * | 2021 | 2 | 2.2 (위하다, 1.03×) | 0.6 (거리 두다) | 2.3 | 0.5 | 0.5 | no excess word | no excess word |
| 2017, 2018 * | 2021 | 3 | 5.5 (같다, 1.28×) | 1.1 (~에 관한 연구) | 2.5 | 0.6 | 0.5 | 1.2 | 1.5 |
| 2020, 2021 * | 2022 | 1 | 0.9 (이해하다, 1.15×) | 0.4 (활용되다 있다) | 0.7 | 0.2 | 0.1 | no excess word | no excess word |
| 2019, 2020, 2021 * | 2022 | 1 | 0.9 (이해하다, 1.14×) | 0.2 (신뢰하다) | 0.7 | 0.2 | 0.2 | no excess word | no excess word |
| 2018, 2019, 2020, 2021 * | 2022 | 1 | 0.7 (이해하다, 1.10×) | 0.2 (신뢰하다) | 0.7 | 0.2 | 0.3 | no excess word | no excess word |
| 2017, 2018, 2019, 2020, 2021 * | 2022 | 1 | 0.9 (있다, 1.01×) | 0.1 (중요성 강조하다) | 0.7 | 0.1 | 0.4 | no excess word | no excess word |
| 2019, 2020 * | 2022 | 2 | 1.4 (아니, 1.08×) | 0.7 (살다) | 1.0 | 0.2 | 0.4 | −0.6 | −0.6 |
| 2018, 2019, 2020 * | 2022 | 2 | 1.2 (가능성, 1.11×) | 0.4 (늘다) | 1.0 | 0.2 | 0.4 | no excess word | no excess word |
| 2017, 2018, 2019, 2020 * | 2022 | 2 | 1.3 (있다, 1.02×) | 0.4 (가속화) | 1.2 | 0.2 | 0.5 | no excess word | no excess word |
| 2018, 2019 * | 2022 | 3 | 3.1 (통하다, 1.06×) | 1.8 (대응하다) | 2.6 | 0.4 | 0.5 | 2.2 | 1.7 |
| 2017, 2018, 2019 * | 2022 | 3 | 2.5 (통하다, 1.05×) | 0.6 (실효성) | 2.6 | 0.4 | 0.9 | 0.6 | 0.6 |
| 2017, 2018 * | 2022 | 4 | 6.3 (같다, 1.34×) | 2.2 (기대하다) | 3.3 | 0.7 | 3.0 | 2.1 | 2.9 |

**Table 11:** Set-based lower bound with selection and estimation on different journals. On half A of the journal panel (929 journals) the excess style units for the target year are ranked; two kinds of set are formed as in Kobak et al.: the top-k units by excess, with k chosen on half A as the value that maximises the gap there, and the set of all excess units whose base-period share is below a threshold T, with T chosen on half A in the same way. Each set is then carried unchanged to half B (924 journals), where P is the share of abstracts containing at least one unit of the set, Q the expectation of that share from its own 2018–2022 trend, and P − Q the bound. A unit is excess when its ratio is at least 1.5 and its excess at least 1 pp; the "lemmas only" variants exclude two-unit expressions so that no topic noun can enter through a bigram. Brackets are 95% intervals from 500 resamples of the journals of half B. Table A9 gives the full k and T curves, and Table A8 lists the 2026 lemma set.

| Year | Set | Selection (on A) | Units | P on B (%) | Q on B (%) | P − Q (pp) [95% CI] | Units (first ten) |
|---|---|---|---|---|---|---|---|
| 2024 | Top-k set, all style units | k = 5 | 5 | 21.1 | 13.7 | **7.5** [6.3, 8.6] | 기여하다, 강화하다, 탐구하다, 중요한 역할, 기대되다 |
| 2024 | Top-k set, lemmas only | k = 5 | 5 | 21.8 | 14.4 | **7.4** [6.2, 8.6] | 기여하다, 강화하다, 탐구하다, 기대되다, 효율성 |
| 2024 | Rare excess set, all style units | base share < 5% | 8 | 24.5 | 16.2 | **8.3** [6.9, 9.4] | 기여하다, 강화하다, 탐구하다, 중요한 역할, 기대되다, 효율성, 촉진하다, 기여하다 있다 |

| Year | Set | Selection (on A) | Units | P on B (%) | Q on B (%) | P − Q (pp) [95% CI] | Units (first ten) |
|---|---|---|---|---|---|---|---|
| 2024 | Rare excess set, lemmas only | base share < 5% | 6 | 23.3 | 15.4 | **7.8** [6.7, 9.0] | 기여하다, 강화하다, 탐구하다, 기대되다, 효율성, 촉진하다 |
| 2025 | Top-k set, all style units | k = 10 | 10 | 65.1 | 44.6 | **20.4** [18.1, 22.4] | 특히, 시사하다, 가능성, 기여하다, 확인되다, 단순하다, 실증적, 넘다, 효과적, 강화하다 |
| 2025 | Top-k set, lemmas only | k = 10 | 10 | 65.1 | 44.6 | **20.4** [18.2, 22.2] | 특히, 시사하다, 가능성, 기여하다, 확인되다, 단순하다, 실증적, 넘다, 효과적, 강화하다 |
| 2025 | Rare excess set, all style units | base share < 2% | 60 | 70.4 | 50.1 | **20.3** [18.1, 22.2] | 단순하다, 실천적, 통합적, 기능하다, 기여하다 있다, 탐구하다, 전략적, 실증적 분석하다, 정서적, 영향 분석하다 ... |
| 2025 | Rare excess set, lemmas only | base share < 2% | 38 | 59.7 | 39.1 | **20.6** [18.5, 22.8] | 단순하다, 실천적, 통합적, 기능하다, 탐구하다, 전략적, 정서적, 촉진하다, 복합적, 조명하다 ... |
| 2026 | Top-k set, all style units | k = 5 | 5 | 75.4 | 50.8 | **24.6** [21.5, 27.9] | 시사하다, 보이다, 확인되다, 이러하다, 구조적 |
| 2026 | Top-k set, lemmas only | k = 5 | 5 | 75.4 | 50.8 | **24.6** [21.4, 27.7] | 시사하다, 보이다, 확인되다, 이러하다, 구조적 |
| 2026 | Rare excess set, all style units | base share < 1% | 88 | 76.3 | 40.0 | **36.3** [34.2, 38.7] | 기능하다, 결합되다, 통합하다, 있음을 시사하다, 반복적, 전환되다, 실증적 분석하다, 의의를 지니다, 전환하다, 제시하다 점 ... |
| 2026 | Rare excess set, lemmas only | base share < 1% | 47 | 60.1 | 27.0 | **33.0** [30.8, 35.4] | Table A8 |

**Table 12:** Single-word bounds before and after density normalisation. Every document frequency of year $t$ is multiplied by the ratio of the base-year mean density (87.5 distinct units per abstract, Appendix B) to the density of year $t$; the scale column gives that factor. Bounds are the largest excess in pp over all style units and over units with excess ratio at least 1.5, with the attaining word in brackets; the 2026 pair coincide under both statistics. Table A3 gives the same re-run word by word.

| Year | Density scale | All style, raw | All style, normalised | Ratio ≥ 1.5, raw | Ratio ≥ 1.5, normalised |
|---|---|---|---|---|---|
| 2023 | 1.004 | 1.0 (및) | 1.1 (및) | 0.1 (통하다 사용자) | 0.1 (통하다 사용자) |
| 2024 | 0.989 | 4.8 (중요하다) | 4.6 (중요하다) | 3.5 (기여하다) | 3.4 (기여하다) |
| 2025 | 0.953 | 11.0 (분석하다) | 9.3 (특히) | 10.5 (특히) | 9.3 (특히) |
| 2026 (Jan–Aug) | 0.895 | 16.1 (시사하다) | 13.8 (시사하다) | 16.1 (시사하다) | 13.8 (시사하다) |

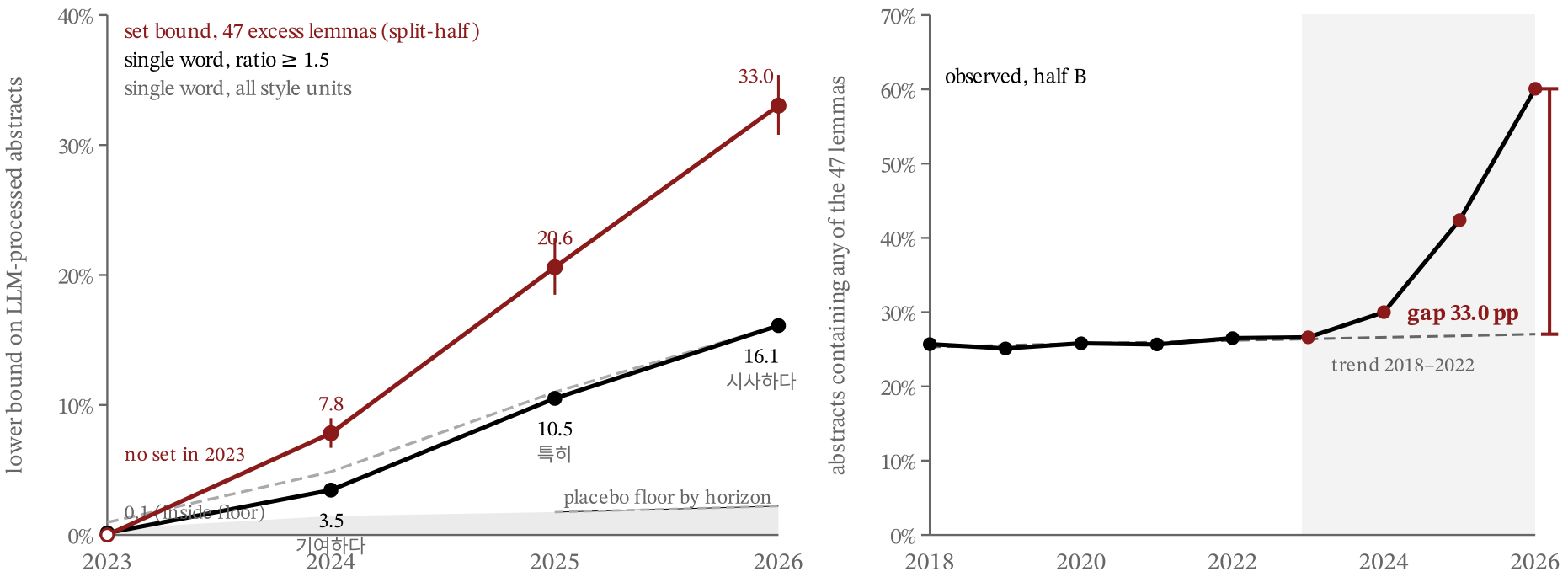

**Figure 7:** Lower bounds on the share of LLM-processed Korean abstracts. Left: by year, the single-word bound searched over style units with excess ratio at least 1.5 (black, word named at each point), the same statistic over all style units (grey dashed), and the split-half set bound for the set of clearly excess style lemmas (red, with 95% journal-resampling intervals; no set exists for 2023). The shaded band is the placebo floor of the ratio-restricted statistic measured on pre-LLM years at the same horizon (Section 5.8); at four years ahead it is the measured two-base-year floor of 2.2 pp (2017–2018 fitted, 2022 measured; Table 10). Right: the share of abstracts of the measurement half that contain at least one of the 47 lemmas chosen on the other half for 2026, 2018–2026, with the least-squares trend through 2018–2022 extrapolated forward; the bracket is the 2026 gap.

### 5.10 Sensitivity, specificity and the scenarios the floor allows

A reader who wants a prevalence rather than a floor needs the sensitivity $s$ of the set indicator on processed abstracts and its false-positive rate $q$ on unprocessed ones (Section 4.7). Table 13 measures both on the material of Sections 3.3 and 5.6, none of which was used to choose the set. On the pre-2023 abstracts of half B the 47-lemma indicator fires in 25.1 to 26.5% of abstracts, and on the 100 real 2019 abstracts that seed the editing conditions in 24%; as a document classifier its specificity is therefore about 76%, which is why it is used only at corpus level, where the expectation $Q$ removes the false positives, and never to label single abstracts. Its sensitivity depends on what the model was asked to do. Asked to write an abstract of 600 to 700 characters from a pre-2023 title, the models produce at least one of the 47 lemmas in 61% (Claude Sonnet 5), 68% (OpenAI) and 70% (EXAONE) of their abstracts; asked to rewrite a real abstract presented as a rough draft, in 40% (OpenAI), 47% (Claude) and 48% (EXAONE), a range of 40–48% across the three rewriting conditions; asked for a short abstract, in 30 to 53%; asked to polish a real abstract, in 26 to 35%, against 24% for the same abstracts before polishing. Polishing moves the indicator so little because it removes rare style lemmas about as often as it inserts them: across the four editing conditions, 5 to 18% of the originals that lacked a set word gained one and 0 to 36% of those that had one lost it; rewriting inserts more (25 to 34% gained for the two API models) and removes about as much (12% lost) (Table A11). Converting the 2026 gap of 33.0 pp with these sensitivities (Table 13, last columns) gives an implied share of 77% (EXAONE), 81% (OpenAI) and 97% (Claude) under the three matched-length drafting conditions, with propagated ranges, from the Wilson interval of each sensitivity and the bootstrap interval of the gap, of 58% to above 100, 62% to above 100 and 72% to above 100. The short-drafting conditions imply more than 100% or are not identifiable, the rewriting conditions imply more than 100%, and under the polishing conditions the sensitivity lies between 26 and 35% against a $Q$ of 27.0% (Table 13), so the implied share is either not identifiable or above 100%, and no share of polished abstracts reproduces the observed 2026 vocabulary. The two-word set (시사하다, 확인되다), whose expected 2026 rate is lower (12.2%), gives 58 to 86% under the matched-length conditions, 66% (Claude) and 90% (OpenAI) under rewriting by the two API models, more than 100% under the Korean model's rewriting, which writes 시사하다 into only 10% of its rewrites, and more than 100% under the others. These are scenario values, not an estimate of prevalence: each is what the 2026 gap would mean if every processed abstract had been processed the way one prompt and one model process, and the field's mixture of models, prompts and habits is unknown. The rewriting condition shows how much the scenario depends on that mixture. Rewriting reproduces the two headline words at their 2026 rates but not the rare lemmas of the 47-word set (Section 5.6), so on the narrow set it implies, for the two API models, that 66–90% of 2026 abstracts were rewritten, and on the broad set that no share of such rewriting suffices: the breadth of the 2026 vocabulary is a fact about the field that our rewriting prompt does not reproduce. If 2026 abstracts were processed the way our matched-length prompts process them, the gap corresponds to 77–97% of them; if they were mostly polished the way our editing prompt polishes, the observed vocabulary could not have arisen at any prevalence, so either the processing in the field is heavier than that prompt or the models in use write these words more often than the ones we tested. Neither statement identifies the share: a sensitivity outside the range we could measure, or a mixture of processing types, moves it in either direction, and only the floor of Section 5.9 survives without a sensitivity assumption. Under A1–A2 and any feasible sensitivity, the 33% floor remains below the implied processing share; without A2, the direct-processing share remains unidentified (Section 6.3). The earlier years behave the same way: the 2025 gap implies 34 to 52% under the matched-length conditions and 79% under rewriting by the two API models, and the 2024 set, which is the vocabulary of the 2024 model generation, implies 10% under drafting by the 2024-generation model (gpt-4o-mini, sensitivity 90%), against the floor of 7.8.

**Table 13:** Sensitivity and specificity of the set indicator, and the share each sensitivity implies. Columns 4–7 give the share of abstracts (%) containing at least one word of the 2024 set (6 lemmas), the 2025 set (38), the 2026 set (47, Table A8) and the two-word set (시사하다, 확인되다), for every positive-control condition of Table 7, for the 100 real 2019 abstracts the edit conditions started from, and for the real abstracts of half B in 2022 (pre-LLM) and 2026. On processed text the rate is the sensitivity of the indicator; on unprocessed text it is one minus its specificity. The last two columns convert the 2026 gap of the measurement half (33.0 pp for the 47-lemma set against an expected 27.0%; 22.3 pp against 12.2% for the two-word set) into the share of processed abstracts that would produce it if every processed abstract had been produced as in that condition, $\pi = (P - Q)/(s - Q)$ (Section 4.7); "> 100" means that no share reproduces the observed gap under that condition, and "n.i." (not identifiable) that the sensitivity is within five points of the expected human rate. EXAONE conditions have 60 abstracts and all others 100. The range in brackets after an implied share propagates the 95% Wilson interval of the sensitivity and the journal-cluster bootstrap interval of the gap (30.8 to 35.4 pp for the 47-lemma set, 20.4 to 24.1 for the two-word set) monotonically, with Q held fixed: it is a propagated range, not a confidence interval with exact coverage, and ">100" at the upper end means that the lower limit of the sensitivity is within five points of Q.

| Condition | n | Length | 2024 set | 2025 set | 2026 set | Two-word set | Implied 2026 share, 47-lemma set (%) | Implied 2026 share, two words (%) |
|---|---|---|---|---|---|---|---|---|
| draft, gpt-4o-mini | 100 | 509 | 90 | 80 | 30 | 23 | n.i. | > 100 (>100–>100) |
| draft, gpt-5.6-luna | 100 | 420 | 22 | 67 | 53 | 30 | > 100 (87–>100) | > 100 (74–>100) |
| draft, gpt-5.6-terra | 100 | 313 | 16 | 59 | 43 | 21 | > 100 (>100–>100) | > 100 (>100–>100) |
| long, Claude Sonnet 5 | 100 | 632 | 54 | 79 | 61 | 51 | 97 (72–>100) | 58 (42–83) |
| long, EXAONE 3.5 7.8B | 60 | 770 | 95 | 100 | 70 | 38 | 77 (58–>100) | 86 (52–>100) |
| long, gpt-5.6-luna | 100 | 690 | 32 | 87 | 68 | 50 | 81 (62–>100) | 59 (43–86) |
| rewrite, Claude Sonnet 5 | 100 | 687 | 29 | 65 | 47 | 46 | > 100 (>100–>100) | 66 (47–99) |
| rewrite, EXAONE 3.5 7.8B | 60 | 570 | 73 | 83 | 48 | 32 | > 100 (91–>100) | > 100 (64–>100) |
| rewrite, gpt-5.6-luna | 100 | 635 | 28 | 65 | 40 | 37 | > 100 (>100–>100) | 90 (59–>100) |
| edit, Claude Sonnet 5 | 100 | 643 | 17 | 50 | 35 | 20 | > 100 (>100–>100) | > 100 (>100–>100) |
| edit, EXAONE 3.5 7.8B | 60 | 566 | 43 | 72 | 27 | 38 | n.i. | 86 (52–>100) |
| edit, gpt-4o-mini | 100 | 634 | 24 | 58 | 32 | 24 | n.i. | > 100 (97–>100) |
| edit, gpt-5.6-luna | 100 | 640 | 21 | 54 | 26 | 17 | n.i. | n.i. |
| en2ko, Claude Sonnet 5 | 99 | 592 | 22 | 46 | 32 | 18 | > 100 (>100–>100) | > 100 (>100–>100) |
| en2ko, EXAONE 3.5 7.8B | 59 | 588 | 34 | 53 | 31 | 22 | n.i. | > 100 (93–>100) |
| en2ko, gpt-5.6-luna | 99 | 579 | 26 | 53 | 34 | 20 | > 100 (>100–>100) | > 100 (>100–>100) |
| Real 2019 abstracts, the 100 originals of the edit conditions | 100 | 630 | 12 | 45 | 24 | 10 | – | – |
| Real 2022 abstracts, half B | 19,828 | – | 15 | 38 | 27 | 10 | – | – |
| Real 2026 abstracts, half B | 16,784 | – | 25 | 71 | 60 | 35 | – | – |

### 5.11 Lexical diversification

Every document frequency depends on how many distinct units an abstract contains, so the density of distinct units is both a nuisance for the excess statistic and, in its own right, a signature. Counted directly, with no minimum-count filter, on a fixed sample of 12,000 abstracts per year, the number of distinct lemmas per abstract is flat at 87.2–87.8 from 2018 to 2023 and then rises to 88.5, 91.9 and 97.8 in 2024, 2025 and 2026, while mean length moves only from 682 to 698 characters (Table A2). Per 100 characters the density is 13.2–13.3 through 2024 and 13.8 and 14.4 in 2025 and 2026; restricted to abstracts of 600–800 characters, which holds length nearly constant by construction, it is 89.1 lemmas in 2022 and 98.0 in 2026. A 2026 abstract of a given length carries about a tenth more distinct lemmas than a 2022 abstract of that length.

First, the shift is uniform across the distribution: the 10th, 50th and 90th percentiles of density per 100 characters all move up by 1.0–1.1 between 2022 and 2026 and the standard deviation is unchanged (2.30 against 2.32), so the added diversity is spread over many abstracts rather than concentrated in a small, heavily rewritten subset. Second, the abstracts that contain the five main rising markers had lower density than the rest before 2024 (12.8 against 13.3 per 100 characters in 2019 and 2022 alike, $z = -15$ and $-19$) and had caught up by 2026 (14.4 against 14.3): the marker-bearing abstracts gained about 1.6 units and the others about 1.0, so the change extends beyond the abstracts that carry the named markers. Third, the positive control separates the two ways a model can touch a text. When the models polish a real 2019 abstract, density does not change: the paired difference is −0.01 and −0.09 per 100 characters for the two OpenAI editing conditions, with 95% intervals inside ±0.25, and the edited version is denser than the original in 47% of pairs. When they draft, density is far above the human level: 16.2 per 100 characters for the length-matched draft against 13.2 for real 2019 abstracts, and 15.1–19.4 for the shorter drafts. When they rewrite a draft, density rises by 0.71 (OpenAI) and 0.89 (Claude) per 100 characters in paired comparison, with 95% intervals of 0.47 to 0.93 and 0.67 to 1.09, and the rewritten version is denser than its original in 74 and 81% of pairs. The corpus-wide rise of 1.1 per 100 characters between 2022 and 2026 (the mean over abstracts of distinct lemmas per 100 characters on the main sample, not the ratio of the Table A2 column means, which are on a fixed 12,000-abstract sample) therefore cannot come from polishing of the kind our prompt elicits, which adds nothing; it is what a substantial share of drafted or rewritten abstracts would produce, and the rewriting condition puts a number on the second: a corpus in which every abstract had been rewritten as our prompt rewrites would have gained 0.7 to 0.9, so the observed rise needs either a larger share of drafting or heavier rewriting than that prompt elicits. We do not turn this into a share, because the arithmetic would depend on the generating model, but the direction is unambiguous: the corpus is moving toward the density of model-drafted text.

### 5.12 The English abstracts of the same articles

KCI carries an English abstract for most of the articles in the Korean corpus. We harvested it for the same identifiers (Section 3.1), which gives 418,271 of the 428,131 harvested articles an English abstract (coverage 98% overall and 79% in 2026, where indexing is still incomplete), and ran the excess computation of Section 4 on it with lower-cased word forms of at least three characters as units (English section headings are not stripped, unlike the Korean side in Section 4.2, and AI-topic words are excluded from marker candidates as in Table 15), with the same journal panel, the same base years and the same per-year cap (between 25,820 and 40,000 abstracts per year). The pairing answers what the comparison with the English literature in Section 5.2 cannot: whether the one-year lag holds with author, journal and article held fixed, and whether the Korean shift is confined to articles whose English abstract shows model vocabulary. Koo, Kim and Kim compared the two languages at the level of journals and years [24]; we are not aware of an earlier article-level pairing.

**Timing.** In the English abstracts of these articles the excess appears in 2023: *insights* in 1.9% of abstracts against 0.5% expected, *crucial* 2.2 against 0.7, *address* 2.3 against 1.0, *additionally* 4.3 against 1.9 (from the run behind Table 14, which lists the 2026 words). By 2024 it is large (*underscores* in 1.2% of abstracts against 0.0% expected, *advancements* 1.1 against 0.0, *foundational* 2.3 against 0.1, *additionally* 11.2 against 2.1), and the 2024 English set is the vocabulary reported for English journals in that year [1, 24]. The Korean abstracts of the same articles show nothing in 2023 and begin in the second half of 2024 (Section 5.2). The lag is therefore not an artefact of the corpora being compared: the English abstracts of the same articles changed about a year be-fore their Korean abstracts. The English single-word bound, computed as in Section 5.9 on words with an excess ratio of at least 1.5, is 13.2 pp in 2024, 23.1 in 2025 and 34.3 in 2026 (attained by *these*, *findings*, *findings*), against 3.5, 10.5 and 16.1 for Korean (Table 9); in 2023 it is already 3.3 pp, against a placebo level of 1.1 pp one year ahead and 1.6 pp two years ahead on pre-ChatGPT targets, while the Korean statistic in 2023 sits at its own floor.

**Turnover in English.** A fixed list of 33 English words widely reported as markers of the 2024 model generation (delve, underscore, showcase, intricate, pivotal and their kin) rises in these articles from 10% of abstracts before 2023 to 41% in 2025 and falls back to 32% in 2026, while the English set of 2026 (*reconfigured*, *foundational*, *mere*, *predominantly*, *reconstructs*, *frameworks*) keeps rising. The marker turnover of Section 6.2 therefore has an English counterpart in the same journals, on the same timetable.

**Pairing.** Table 15 and Figure 8 cross the two flags article by article on half B of the journals, with both sets chosen on half A. Before 2023 an article whose English abstract contained a word of the 2026 English set was 1.6 to 1.9 times as likely, in odds, to contain a lemma of the Korean 2026 set as an article that did not; that is the baseline association that style-heavy authors and journals produce in both languages. The odds ratio rises to 2.2 in 2025 and 2.3 in 2026: the two abstracts of an article moved together more than they used to. But the Korean rate among articles whose English abstract contains none of the period's English markers also rises, from 24% before 2023 to 47% in 2026, against 67% among articles whose English abstract does. The Korean change is therefore not a by-product of a model-written English abstract; it appears, at 70% of the rate read directly from Table 15 and 30 to 66% of it once the English indicator's own sensitivity is allowed for, in articles whose English side shows no model vocabulary at all. With the fixed 2024-generation list the odds ratio does not rise in 2026 (1.2), because that vocabulary recedes in English at the same time, which is why a period-matched set is needed for this test.

**How much the English indicator misses.** The statement that the Korean shift is present where the English abstract carries no marker assumes that an English abstract a model touched would carry one. Table 19 measures that sensitivity the way Section 5.10 measures the Korean one: on English abstracts written from the article's own English title, the 47-word set fires in 76 to 95% of them; on model translations of the article's Korean abstract, in 39 to 71%. Solving the two-by-two mixture for each sensitivity, under the assumption that the English flag and the Korean marker are independent once it is known whether a model touched the English side (author, journal and field effects that move both would violate it), gives the Korean marker rate among articles whose English side a model did not touch: 21 to 45% against the 47% read directly off Table 15, that is 30 to 66% of the rate among articles whose English side was touched rather than the uncorrected 70%. The direction of the claim survives every configuration; its size does not. A sensitivity below 64.0%, the observed 2026 English marker rate, would require more than every article to have been processed, so the low sensitivities of the translation conditions cannot describe 2026 English abstracts as a class, and the lowest sensitivity consistent with a non-negative Korean rate in the untouched group is 72%. And the corrected rate is a floor on the untouched group only in the sense that it inherits A1 and A2 of Section 4.7: it is not a measurement of how many Korean abstracts were written without a model.

What the pairing cannot separate is a model that wrote both abstracts from a model that translated one into the other, and the reading of Koo et al. that the 2024 English rise came from model translation of author-written Korean [24] remains open for the articles in which only the English side changed. What it establishes is the order of events and the partial independence: English first, Korean a year later, and a Korean shift that is present, at 30 to 66% of the rate, where the English abstract carries none of the period's markers.

**Table 14:** Excess vocabulary in the English abstracts of the same KCI articles. The English abstract that KCI carries for each article in the Korean corpus was harvested for the same identifiers and run through the same pipeline (journal panel with at least five abstracts in the base and in the target period, at most 40,000 abstracts per year, base years 2018–2022, least-squares trend), with lower-cased word forms as units. Rows are the twenty English words with the largest 2026 excess ratio among those with at least 1 pp of excess; cells give observed / expected document frequency (%) in 2024, 2025 and 2026, then the 2026 ratio and excess. Year sizes: 2018: 40,000, 2019: 39,274, 2020: 40,000, 2021: 36,843, 2022: 40,000, 2023: 27,173, 2024: 40,000, 2025: 40,000, 2026: 25,820.

| Word | 2024 obs / exp | 2025 obs / exp | 2026 obs / exp | Ratio 2026 | Excess 2026 (pp) |
|---|---|---|---|---|---|
| underscore | – | 1.3 / 0.0 | 1.0 / 0.0 | 103.8 | 1.0 |
| reconfigured | – | – | 1.1 / 0.0 | 98.8 | 1.1 |
| foundational | 2.3 / 0.1 | 4.2 / 0.1 | 4.8 / 0.1 | 53.3 | 4.7 |
| conceptualizes | – | – | 1.5 / 0.0 | 45.4 | 1.4 |
| mere | – | 2.2 / 0.1 | 3.8 / 0.1 | 32.9 | 3.7 |
| merely | 1.2 / 0.1 | 4.0 / 0.0 | 7.1 / 0.0 | 26.9 | 7.1 |
| t-tests | – | 1.1 / 0.1 | 1.6 / 0.0 | 26.4 | 1.5 |
| frameworks | – | 3.0 / 0.1 | 3.6 / 0.1 | 26.2 | 3.4 |
| offers | 2.2 / 0.3 | 5.7 / 0.3 | 5.7 / 0.2 | 24.3 | 5.4 |
| shaped | – | 2.1 / 0.1 | 3.8 / 0.1 | 23.2 | 3.7 |
| sustained | – | 1.1 / 0.1 | 2.2 / 0.0 | 22.5 | 2.1 |
| reliance | – | – | 1.1 / 0.0 | 22.3 | 1.0 |
| shifts | – | 1.7 / 0.1 | 2.7 / 0.1 | 20.1 | 2.6 |
| broader | 1.5 / 0.2 | 3.2 / 0.2 | 3.2 / 0.2 | 18.9 | 3.1 |
| addressing | 2.2 / 0.2 | 3.1 / 0.2 | 2.4 / 0.1 | 16.4 | 2.2 |
| notably | 1.8 / 0.2 | 3.6 / 0.2 | 3.2 / 0.2 | 16.4 | 3.0 |
| alongside | – | 1.6 / 0.2 | 2.7 / 0.2 | 16.4 | 2.5 |
| primarily | 2.7 / 0.5 | 4.4 / 0.4 | 6.6 / 0.4 | 15.9 | 6.1 |
| demonstrating | – | 2.4 / 0.3 | 3.7 / 0.2 | 14.8 | 3.5 |
| employs | – | 1.6 / 0.2 | 2.2 / 0.2 | 14.4 | 2.0 |

**Table 15:** The same articles in two languages. For every article of the measurement half (half B of the journals) that has both a Korean and an English abstract, the share whose English abstract contains at least one word of an English marker set, the share whose Korean abstract contains at least one lemma of the Korean 2026 set (47 lemmas, Table A8), the share with both, the Korean rate conditional on the English flag being present or absent, and the odds ratio between the two flags. The first block uses an English set built like the Korean one: on half A of the journals, words with a base-period document frequency of at most 1%, an excess ratio of at least 1.5 and at least 1 pp of excess in 2026, of which the 47 with the largest ratios are kept so that the set has the size of the Korean set (reconfigured, foundational, conceptualizes, alongside, mere, notably, merely, shaped, frameworks, offers, reliance, sustained, addressing, employs, ...; AI-topic words such as chatgpt, llm and generative excluded). The second block uses a fixed list of 33 English words widely reported as LLM markers (delve, underscore, showcase, intricate, pivotal, meticulous, notably, additionally, crucial, comprehensive, insights, highlight, realm, garner, leverage, foster, tapestry, landscape, navigate, emphasize and their inflections), which is the vocabulary of the 2024 model generation. Both sets are measured out of sample; the odds ratio in the pre-LLM years is the baseline association between style-heavy English and style-heavy Korean writing that author and journal effects produce, and the 2024–2026 values should be read against it.

| Year | Articles with both abstracts | English set (%) | Korean set (%) | Both (%) | Korean given English (%) | Korean given no English (%) | Odds ratio |
|---|---|---|---|---|---|---|---|
| *English set: the 47 rare-excess words of 2026 with the largest ratios (chosen on half A)* | | | | | | | |
| 2018 | 19,822 | 12.0 | 25.7 | 4.1 | 34.2 | 24.5 | 1.60 |
| 2019 | 19,398 | 11.4 | 25.1 | 4.0 | 34.6 | 23.9 | 1.68 |

| Year | Articles with both abstracts | English set (%) | Korean set (%) | Both (%) | Korean given English (%) | Korean given no English (%) | Odds ratio |
|---|---|---|---|---|---|---|---|
| 2020 | 19,686 | 10.5 | 25.8 | 3.8 | 36.3 | 24.6 | 1.75 |
| 2021 | 18,338 | 10.1 | 25.6 | 3.8 | 37.6 | 24.3 | 1.88 |
| 2022 | 19,754 | 9.9 | 26.5 | 3.8 | 38.3 | 25.2 | 1.84 |
| 2023 | 13,361 | 15.5 | 26.6 | 5.5 | 35.5 | 25.0 | 1.65 |
| 2024 | 19,474 | 29.4 | 30.1 | 11.0 | 37.4 | 27.0 | 1.62 |
| 2025 | 19,912 | 48.5 | 42.4 | 25.2 | 52.0 | 33.4 | 2.16 |
| 2026 | 13,121 | 64.0 | 59.6 | 42.7 | 66.7 | 46.9 | 2.27 |
| *English set: a fixed list of 33 widely reported English markers (delve, underscore, showcase, …)* | | | | | | | |
| 2018 | 19,822 | 9.6 | 25.7 | 3.2 | 33.2 | 24.9 | 1.50 |
| 2019 | 19,398 | 9.9 | 25.1 | 3.1 | 31.3 | 24.4 | 1.41 |
| 2020 | 19,686 | 10.4 | 25.8 | 3.5 | 33.3 | 24.9 | 1.50 |
| 2021 | 18,338 | 9.9 | 25.6 | 3.1 | 31.4 | 25.0 | 1.37 |
| 2022 | 19,754 | 10.1 | 26.5 | 3.4 | 33.5 | 25.7 | 1.45 |
| 2023 | 13,361 | 17.8 | 26.6 | 5.8 | 32.3 | 25.4 | 1.40 |
| 2024 | 19,474 | 32.8 | 30.1 | 11.4 | 34.9 | 27.8 | 1.40 |
| 2025 | 19,912 | 40.9 | 42.4 | 19.6 | 47.8 | 38.7 | 1.45 |
| 2026 | 13,121 | 31.7 | 59.6 | 20.0 | 63.1 | 58.0 | 1.24 |

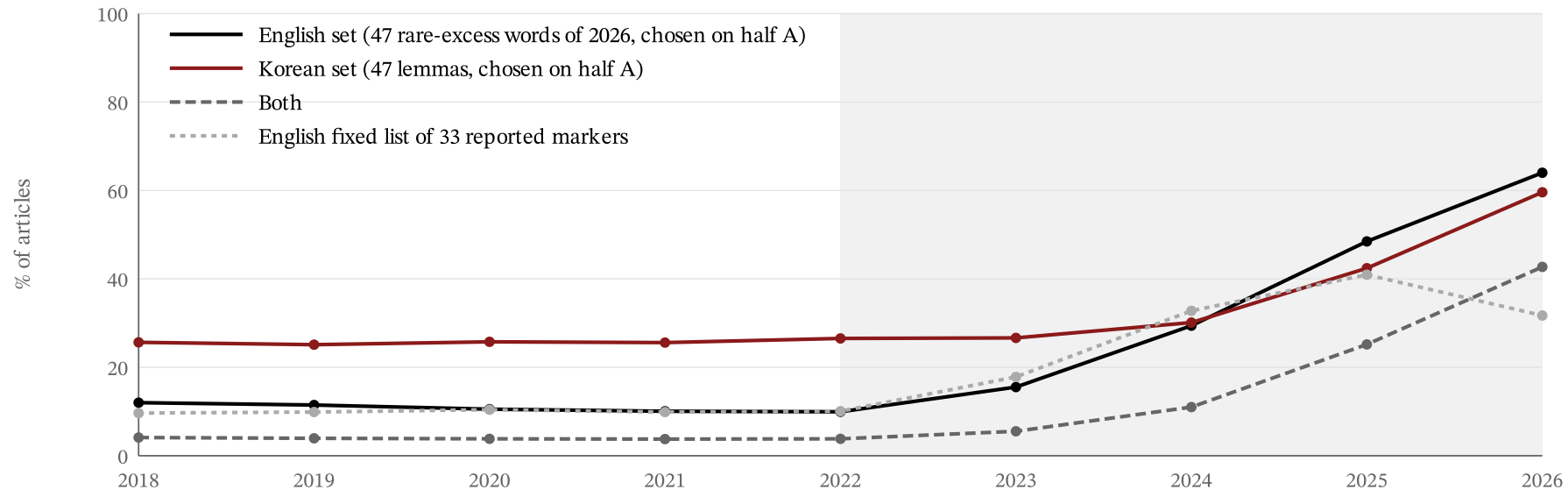

**Figure 8:** The same articles in two languages. Share of articles in half B with both abstracts whose English abstract contains a word of the 2026 English set (47 rare-excess words chosen on half A) or of the fixed list of 33 reported markers, whose Korean abstract contains a lemma of the Korean 2026 set (47 lemmas), and both. Shaded: the years after ChatGPT. The English set moves in 2023, the Korean set in the second half of 2024; the fixed list of 2024-generation words peaks in 2025 and recedes in 2026.

### 5.13 Style or topic: what the marker set is made of

The set bound counts abstracts that carry one of the rare lemmas that rose. If some of those lemmas rose because the field's subject matter moved rather than its wording, the bound absorbs a change that has nothing to do with models. Two checks bound that possibility, one on the words and one on the documents.

**Independent annotation.** The 82 lemmas that make up the rare-excess sets of 2024, 2025 and 2026 were classified as style, topic or ambiguous by three language-model annotators working independently from the same rubric and from the lemma alone, one OpenAI and two Anthropic models (Table 16). Agreement is fair rather than good, Fleiss' $\kappa = 0.24$, and the disagreement is itself informative: the Korean suffixes 적, 화 and 성 build words such as 정량적 "quantitative" and 접근성 "accessibility" that sit exactly on the boundary between a framing device and a subject matter, and the annotators split on 41 of the 82 lemmas, on which at least one of them voted topic. The bound is therefore recomputed on three restricted sets. Dropping every lemma that any annotator called topic leaves 22 lemmas and a 2026 bound of 24.9 pp; keeping only those a majority called style leaves 27 and 27.9 pp; keeping only the 12 lemmas all three called style leaves 14.7 pp, against 33.0 pp for the full set. The placebo of the same restricted sets on pre-LLM target years stays within 2.6 pp. These fixed-set placebos, the same lemmas re-measured on a pre-LLM target, are a different quantity from the re-selection placebo floor of Section 5.8, where the marker set is selected afresh for each placebo target. Part of the fall is mechanical, since a set statistic shrinks when the set does, and the strictest restriction keeps 12 of 47 lemmas; what the restriction establishes is that the sign and most of the magnitude are carried by lemmas no annotator reads as a subject matter, and that even the smallest of these sets stands 6 times the largest placebo value that any restricted set shows (Table 16), its own placebo values being smaller still. Topic words inside the set do not account for the bound.

**Topic-matched abstracts.** The second check works at the level of the document rather than the word. Within each journal of the measurement half, every 2026 abstract is paired with the base-period abstract of the same journal whose title is closest in character-bigram space, which holds journal fixed and matches subject only loosely, as the title similarities of Table 17 show; 16,687 abstracts in 810 journals have such a partner (Table 17). The 2026 abstracts carry a marker in 60.1% of cases, their topic-matched partners in 26.0%, and all base abstracts of the same journals in 26.1%. Matching on topic moves the comparison by 0.1 pp: the difference is 34.1 pp matched against 34.0 pp unmatched. In the quarter of pairs with the closest titles the matched difference is 31.3 pp against 34.8 pp in the quarter with the furthest titles: closeness of topic barely moves it. The two topic controls tried here, neither of them strong, do not account for most of this pattern; the residual is more consistent with a change in how the journals write than in what they study.

**Table 16:** The set bound when the marker set is restricted by a separate style/topic annotation, three language models classifying independently of one another. The 82 lemmas that make up the rare-excess sets of 2024, 2025 and 2026 were classified as style, topic or ambiguous by three annotators working independently from the same written rubric and from the lemma alone: one OpenAI model (gpt-5.6-luna) and two Anthropic models (Claude Sonnet 5, Claude Opus 5). Pairwise agreement is gpt-5.6-luna vs claude-sonnet-5 56%, gpt-5.6-luna vs claude-opus-5 50%, claude-sonnet-5 vs claude-opus-5 63% and Fleiss' $\kappa$ is 0.24, which is fair rather than good agreement and is itself a result: the boundary between a framing verb and a content word is not sharp in Korean derivational morphology. Each restricted set is measured on the same half B of the journals as the full set, with the same base years, so the rows differ only by which lemmas are counted. The placebo column gives the same statistic for the same set on pre-LLM target years (2018–2020 → 2022 / 2018–2019 → 2022). These fixed-set placebo values are a different quantity from the re-selection placebo floor of Tables 9 and 10, where the marker set is selected afresh for each placebo target.

| Year | Marker set | Lemmas | Observed (%) | Expected (%) | Bound (pp) | Placebo (pp) |
|---|---|---|---|---|---|---|

| Year | Marker set | Lemmas | Observed (%) | Expected (%) | Bound (pp) | Placebo (pp) |
|---|---|---|---|---|---|---|
| 2024 | All rare-excess lemmas of the year | 6 | 23.3 | 15.4 | **7.8** | −0.1 / 1.4 |
| | Majority of annotators call it style | 5 | 21.2 | 13.8 | **7.4** | 0.2 / 1.9 |
| | All three annotators call it style | 3 | 15.7 | 9.6 | **6.1** | −0.4 / 0.4 |
| | No annotator calls it topic | 5 | 21.2 | 13.8 | **7.4** | 0.2 / 1.9 |
| 2025 | All rare-excess lemmas of the year | 38 | 59.7 | 39.1 | **20.6** | 0.7 / 3.0 |
| | Majority of annotators call it style | 22 | 43.8 | 24.0 | **19.8** | 0.0 / 0.9 |
| | All three annotators call it style | 14 | 34.1 | 17.1 | **17.0** | −0.5 / -0.5 |
| | No annotator calls it topic | 19 | 40.7 | 21.7 | **19.0** | 0.3 / 1.0 |
| 2026 | All rare-excess lemmas of the year | 47 | 60.1 | 27.0 | **33.0** | 0.8 / 3.2 |
| | Majority of annotators call it style | 27 | 42.9 | 15.1 | **27.9** | 0.5 / 2.6 |
| | All three annotators call it style | 12 | 21.7 | 6.9 | **14.7** | −0.3 / 0.2 |
| | No annotator calls it topic | 22 | 37.8 | 12.9 | **24.9** | 0.4 / 2.2 |

Annotator labels and the rubric are in the reproducibility package (results_marker_annotation.json).

**Table 17:** Marker rate in title-matched pairs within journals. Within each journal of the measurement half, every 2026 abstract is paired with the base-period (2018–2022) abstract of the same journal whose title is most similar, by cosine similarity of character-bigram vectors weighted by inverse document frequency; 16,687 abstracts in 810 journals have a match. The column "matched base" gives the marker rate of the paired base abstracts, the column "all base" the rate over all base abstracts of the same journals. If the 2026 rise were topic diffusion, the topic-matched base abstracts would already carry the markers and the matched difference would be much smaller than the unmatched one. Median title similarity is 0.19 (quartiles 0.13 and 0.27).

| Group | Pairs | 2026 (%) | Matched base (%) | All base (%) | Matched difference (pp) | Unmatched difference (pp) |
|---|---|---|---|---|---|---|
| All matched 2026 abstracts | 16,687 | 60.1 | 26.0 | 26.1 | **34.1** | 34.0 |
| Closest quarter of title matches | 4,172 | 54.1 | 22.8 | – | **31.3** | – |
| Furthest quarter of title matches | 4,172 | 62.6 | 27.8 | – | **34.8** | – |

### 5.14 Translation as a route

Section 5.12 leaves one route open: a model that produced the English abstract and a translation, by machine or by hand, into the Korean one. Two measurements narrow it.

**The Korean abstracts do not carry the marks of translation.** If Korean abstracts were increasingly rendered from an English original, the surface marks of translated Korean should rise with the marker vocabulary. They fall (Table 18). Between 2019 and 2026 the share of abstracts using the agentive 에 의해 goes from 7.9% to 5.7%, the sentence-final 것이다 from 28.1% to 12.9%, the analytic 에 대한 from 52.6% to 36.1%, and the overt plural 들 from 57.8% to 32.9%; the double passive 되어지다, long treated as a translation artefact in Korean style guides, all but disappears. Only Latin-script words rise, from 27.8% to 39.6%, which is a different phenomenon: terminology, not syntax. The model conditions place these numbers: a model asked to translate the article's own English abstract into Korean produces 에 의해 in 11.1% of its output, well above the 2026 corpus rate, while a model asked to polish a real 2019 abstract leaves it near the 2019 rate. The 2026 corpus looks like the polishing and drafting conditions, not like the translation condition.

**Nor can translation describe the English side of 2026 as a class.** The same arithmetic that corrects the 70% in Section 5.12 rules out translation as the description of 2026 English abstracts as a class: the English marker set fires on 39 to 71% of model translations of Korean abstracts, and any sensitivity below 64.0% would require more than every 2026 article to have been processed to explain the observed English marker rate. English abstracts written from the title, by contrast, fire the set in 76 to 95% of cases, which is consistent with the observed rate at a plausible prevalence. This does not exclude translation for individual articles, and the reading of Koo et al. remains available for the 2023–2024 English rise; what it excludes is translation as the general account of the 2026 English vocabulary.

**Table 18:** Marks of a translation route in the Korean abstracts. Document frequency (%) of six features that Korean-English translation tends to leave behind, measured with regular expressions on the whole corpus after removing parenthesised glosses. If the Korean abstracts of 2025 and 2026 were increasingly translations of an English original, these features should rise with the marker vocabulary. The last rows give the same features for the model conditions of Section 3.3, including a condition that translates the article's own English abstract into Korean.

| Group | Abstracts | Latin-script word | 에 의해 *by* (agentive) | 것이다 *it is that* | 에 대한 *about* | 들 (overt plural) | 되어지다 (double passive) |
|---|---|---|---|---|---|---|---|
| 2019 | 41,484 | 27.8 | 7.9 | 28.1 | 52.6 | 57.8 | 0.6 |
| 2022 | 54,717 | 30.1 | 6.5 | 27.1 | 52.1 | 55.4 | 0.3 |
| 2024 | 50,907 | 31.9 | 5.6 | 25.6 | 50.5 | 53.4 | 0.2 |
| 2025 | 45,745 | 33.9 | 4.9 | 21.0 | 45.4 | 46.8 | 0.1 |
| 2026 | 34,201 | 39.6 | 5.7 | 12.9 | 36.1 | 32.9 | 0.0 |
| 2026, English abstract carries the markers | 17,268 | 37.8 | 6.3 | 12.5 | 37.3 | 34.0 | 0.1 |
| 2026, English abstract carries none | 9,717 | 41.9 | 4.8 | 14.2 | 35.1 | 32.3 | 0.0 |
| Model translation of the English abstract (OpenAI) | 99 | 24.2 | 11.1 | 8.1 | 36.4 | 37.4 | 0.0 |
| Model translation of the English abstract (Claude) | 99 | 23.2 | 13.1 | 19.2 | 49.5 | 45.5 | 0.0 |
| Model polishing of a real abstract (OpenAI) | 100 | 20.0 | 8.0 | 14.0 | 39.0 | 34.0 | 0.0 |

The features are not a translation detector; they are surface marks whose direction is informative, and their base rates differ by field.

**Table 19:** Sensitivity of the English marker set, and what it does to the comparison of Section 5.12. The same 2019 articles that seed the Korean control conditions were used to generate English abstracts from their English titles and English translations of their Korean abstracts; each output is scored with the two English sets of Table 15. The last three columns solve the two-by-two mixture of Section 5.12 for a given sensitivity: $\pi$ is the implied share of 2026 English abstracts that a model touched, and the last two columns give the Korean marker rate among articles whose English was not touched and its ratio to the rate among those that were, against the uncorrected 70%. A sensitivity below the observed English marker rate of 64.0% would require more than every article to have been processed, and one below 72% would give the untouched group a negative Korean rate; conditions that fail either test are marked not possible.

| Condition | n | Words | 47-word set (%) | 33-word list (%) | Implied $\pi$ | Korean rate if English untouched (%) | Ratio to touched (%) |
|---|---|---|---|---|---|---|---|
| English abstract from the title, claude-sonnet-5 | 99 | 230 | 77.8 | 40.4 | 0.80 | 26.4 | 39 |
| English abstract from the title, exaone-3.5-7.8b | 59 | 160 | 94.9 | 100.0 | 0.64 | 44.8 | 66 |
| English abstract from the title, gpt-5.6-luna | 99 | 208 | 75.8 | 26.3 | 0.82 | 20.7 | 30 |
| English translation of the Korean abstract, claude-sonnet-5 | 99 | 232 | 39.4 | 13.1 | 1.81 | not possible | – |

| Condition | n | Words | 47-word set (%) | 33-word list (%) | Implied π | Korean rate if English untouched (%) | Ratio to touched (%) |
|---|---|---|---|---|---|---|---|
| English translation of the Korean abstract, exaone-3.5-7.8b | 59 | 196 | 71.2 | 78.0 | 0.88 | not possible | – |
| English translation of the Korean abstract, gpt-5.6-luna | 99 | 220 | 43.4 | 18.2 | 1.60 | not possible | – |
| *Real English abstracts, 2019* | 3,000 | 218 | 11.7 | 9.3 | – | – | – |
| *Real English abstracts, 2022* | 3,000 | 224 | 9.1 | 9.7 | – | – | – |
| *Real English abstracts, 2024* | 3,000 | 225 | 28.5 | 33.0 | – | – | – |
| *Real English abstracts, 2025* | 3,000 | 219 | 48.6 | 40.4 | – | – | – |
| *Real English abstracts, 2026* | 3,000 | 217 | 64.0 | 30.3 | – | – | – |

False-positive rate q taken from real 2022 English abstracts (9.1%).

### 5.15 An alternative estimator

The bound of Section 5.9 is a frequency gap. Holzwarth, González-Márquez and Kobak have since proposed a second estimator of the same quantity [2]. They write the observed indicator rate as a mixture, $q = (1 - \beta)p_{\text{human}} + \beta p_{\text{LLM}}$, replace the unobservable $p_{\text{LLM}}$ by its upper bound of one, and take $\beta = \max_T (q - \hat{p}_{\text{human}})/(1 - \hat{p}_{\text{human}})$ over a grid of 19 thresholds $T$ that define the word set (0.0002, 0.0005, 0.001, 0.002, 0.005, 0.01, 0.02, 0.03, 0.04, 0.05, 0.07, 0.1, 0.15, 0.2, 0.3, 0.4, 0.5, 0.7 and 1 here; their paper describes twelve log-spaced values with several added between 0.03 and 0.7), discarding any threshold whose standard error exceeds 0.025 or whose $\hat{p}_{\text{human}}$ reaches 0.999. The denominator is below one, so the result is never below the gap, and they report it as an estimate rather than a floor. Applied to the Section 5.9 set as it stands, with $\hat{p}_{\text{human}} = 27.0\%$ on the measurement half, the ratio turns the 33.0-point gap of 2026 into 45.3%; the rest of this section chooses the set for the ratio, as [2] does.

We reimplemented it on the Korean corpus, taking the variance from the form in their released code [25] rather than from the expression printed in their methods section, in which the two partial derivatives appear unsquared and which, evaluated at the December 2025 values of their released series for the word *these* ($q = 50.3\%$, $\hat{p}_{\text{human}} = 33.5\%$, prediction standard deviation 0.60 points), is negative for any sample above 9,448 documents. Their code implements the squared form and asserts that the variance is non-negative, so this is a slip in the printed equation and not an error in their reported standard errors. The implementation is checked against the monthly series for the word *these* that their code repository releases [25]: fitting their regression to the same sixty pre-ChatGPT monthly points reproduces their published expectation column to $4 \times 10^{-16}$ and their published prediction standard error to $5 \times 10^{-18}$, and returns $\beta = 0.2527$ for December 2025 against the 0.25 their text quotes from rounded values.

Their design has no split-half: the word list is fixed in advance, the threshold is chosen on the 2025 data the 2025 estimate is read from, then carried to the other years, and a simulation with 100,000 documents and 500 marker words keeps the error of that maximum, $|\hat{\beta} - \beta|$, below 0.02 in every case run. The nearer of our two columns to it keeps the candidate lemmas of Section 5.9, chosen on half A, and chooses the threshold within each year on half B, the half the estimate is read from ($T$ = 0.05, 0.2, 0.2 for 2024–2026, against 0.05, 0.04, 0.03 cross-fitted); a run that also chose the candidates on half B would be at least as favourable to itself. Under that rule the Korean abstracts give 9.3%, 48.6% and 81.1% for 2024, 2025 and 2026. Choosing the threshold on the selection half instead and reading it on the measurement half, which is the discipline of Section 5.9, gives 9.3% (SE 0.5), 41.9% (1.9) and 72.1% (1.5), on sets of 6, 56 and 114 lemmas. Their threshold is applied to a word's frequency in 2024, ours to its base-period share as in Section 5.9; with the threshold on the target-year frequency of the selection half instead, the cross-fitted values are 9.3%, 43.5% and 70.6%, so the definition does not carry the result. The second column is the one this paper stands behind; the first is the nearer of the two to the design of [2], for a reader comparing with it, though neither reproduces that design. The difference, 6.7 points for 2025 and 9.0 for 2026, is what choosing the threshold on the measurement half adds on this split: it mixes that optimism with split-to-split variation, and both in-sample winners fail the standard-error rule on the other half (0.040 and 0.034 against 0.025), so it illustrates the cost rather than estimating it; the difference is computed on 16 to 20 thousand abstracts per half with candidate lists of 68 and 154 lemmas, a smaller and noisier setting than their simulation covers; for 2024 the candidate list has 6 lemmas, every threshold from 0.05 upward returns the same set, and the two columns coincide. For 2023 no unit passes the selection criteria and there is no set, as in Section 5.9.

Selecting the words inside the corpus is a step [2] does not take, since their list is fixed in advance from earlier English work, so we ran the whole procedure on pre-ChatGPT target years, with the threshold chosen in sample, which favours the placebo, at all 10 configurations the historical store admits with the three regression points their standard error needs. Those configurations reach one to three years ahead; the four-year horizon of the 2026 figure has no pre-ChatGPT counterpart, as in Section 5.8. With the candidate rule of Section 5.9 (ratio at least 1.5, excess at least 1 pp), in 9 of the 10 no unit passes selection and the estimator is never exercised; the tenth, three years ahead, returns 0.6%. With the rule loosened to any style unit whose share rose, so that a set always exists, the ten values run from −1.4% to 7.1%, the largest at one year ahead and the single three-year value at 0.7%. The placebo therefore bounds the estimator's own noise only up to three years ahead and under a looser candidate rule than the one behind the 2025 and 2026 figures; within that limit those two figures sit well above it, and the 2024 figure does not, so we do not read 2024 as a prevalence.

Where the threshold is chosen out of sample the 2026 maximum falls at $\hat{p}_{\text{human}} = 69.5\%$; the in-sample maximum sits at 94.0%. The substitution $p_{\text{LLM}} = 1$ is, if anything, tighter at the second cell, since $q$ there is 98.9% and $p_{\text{LLM}}$ can be no lower than $q$; what the high cell costs is sensitivity to the extrapolation, because the denominator $1 - \hat{p}_{\text{human}}$ is 0.06 and an error of one point in $\hat{p}_{\text{human}}$ moves the estimate by 3.1 points, against 0.9 at the cross-fitted cell. That is the second reason to prefer the cross-fitted column. The gap floors of 7.8, 20.6 and 33.0 points come from sets chosen to maximise the gap rather than this ratio, so for 2025 and 2026 the two columns are not the same word list. And the standard errors here treat abstracts as independent where Section 4.5 resamples journals: at the cross-fitted 2026 set the journal-cluster interval runs from 68.6 to 75.0%, against 72.1 ± 2.9 from twice the analytic error, so clustering costs little at this set size.

What the estimator gives on English biomedical abstracts in 2025 is 53% over the whole year and 68% for December 2025 [2], with the threshold chosen on that year; the Korean abstracts give 48.6% with the threshold chosen on the measurement half and 41.9% cross-fitted. The two marker lists are built differently and the corpora share no vocabulary, no morphology and only a sliver of subject matter, so this is not a controlled comparison. It says only that a statistic designed on English returns a value of the same order when it is rebuilt on Korean morphology, which is all the data support.

## 6 Discussion

### 6.1 A formal-register signature, not a translated one

One word deserves separate comment before the general account. 구조적 “structural” has the third-largest excess in Table 3 and none of the sixteen model conditions, the Korean open-weight model and the Anthropic model included, reproduces it (Section 5.6). One reading, that the rise is an artefact of the fusion rule of Section 4.1, is ruled out by Appendix E: with the rule switched off the same increase reappears on the bare noun 구조, which goes from +20.6 to +27.1 pp of excess, so the change is in the text and not in the tokenisation. What the rule decides is only whether the increase is booked to the adjectival derivation or absorbed into the noun. The two remaining readings are that it comes from models or writing tools outside the five we could query, or that it is a genuine human register shift running alongside the model-driven one, a drift from nominal toward adjectival phrasing. The data here cannot separate those two, and the second deserves to be taken seriously rather than filed as an exception: if the third-largest excess in Table 3 has a source other than model text, then some part of the smaller excesses may too, and the positive control, not the excess itself, is what licenses reading a given word as a model product. Neither headline statistic rests on 구조적: the single-word bound is attained by 시사하다, which every matched-length 2026 model condition writes above its real rate (27 to 41% against 22%), and 구조적 sits in none of the rare-lemma split-half sets that carry the reported bound, in 2024 because its half-A excess (0.2 pp at ratio 1.1) misses the entry cut and in 2025 and 2026 because its base-period share (2.5%) is above the rarity thresholds chosen on half A (2% and 1%), so excluding it changes that bound by nothing; it does enter the 2026 top-k sets of Table 11 (k = 5, 24.6 pp for all style units), which are not the figure this paper reports; excluding every 적/화/성 derivation moves the lemma-only set bound from 33.0 to 33.8 pp and the all-unit one from 36.3 to 35.0 pp: the lemmas that replace the derivations are slightly stronger, while the two-word units admitted with them are not. Then, the three rising units the length-matched control does not reproduce (구조적, 기제,

보여주다) carry 12.6, 2.8 and 8.0 pp of excess against the 16.1 of 시사하다 alone, and the nine it reproduces above their real rates carry more excess between them than the three. And a human register shift of this size and speed would itself be a finding worth having: 구조적 rises exactly when and where the model-produced words rise (Figure 1) and in no field before 2024, which makes model influence the first hypothesis to test, not an established one; what we cannot claim is that a model typed it. The word is not excluded from the analysis: it is a style unit under our rules, it does not attain the corpus-wide bound in any year (시사하다 does in 2026), but it does attain the field bound for law and public administration in Table 6, and that figure should be read with this caveat.

The Korean words that rose are not translations of the English excess vocabulary. English models over-produce *delve*, *underscore*, *intricate*, *showcase*; the Korean signature is built from Sino-Korean verbs of academic register (시사하다 "imply", 규명하다 "elucidate", 기여하다 "contribute", 조명하다 "shed light on", 확인되다 "be confirmed"), evaluative adjectives (단순하다 "mere", 핵심적 "core", 실질적 "substantive", 구조적 "structural") and discourse frames (이러한 결과는 "these results", 단순한 X를 넘어 "beyond a mere X", 특히 "in particular"). The two lists overlap in function rather than in form: both languages gained words that signal significance and depth, and both lost plain ones. The one direct correspondence is *emphasize*: 강조하다 in Korean and nhấn mạnh in Vietnamese both rose, in line with the 24-of-34-language convergence on "emphasize"-type verbs reported for news text [11]. The rest of the Korean list looks like what a careful non-native writer of Korean produces when asked to sound academic, which is consistent with the view that the preference is induced at the tuning stage and expressed in each language through its own formal register [10].

The register shift can be summarised in one number. Summing the document frequencies of twelve Sino-Korean academic verbs (시사하다, 규명하다, 기여하다, 조명하다, 확인되다, 촉진하다, 강화하다, 통합하다, 기능하다, 작동하다, 제고하다, 도출하다) gives 0.36 expected occurrences per abstract in 2018 and 0.40 in 2022, then 0.54 in 2024, 0.83 in 2025 and 1.11 in 2026, a tripling in four years. The same sum over ten native-Korean expressions (알아보다, 살펴보다, 도움이 되다, 많이, 보다, 해보다, 이루어지다, 나타나다, 밝히다, 찾다) is flat at 1.07 to 1.12 through 2022 and then falls to 1.00, 0.89 and 0.80. The two curves cross in the direction of the Sino-Korean layer, which is the layer Korean uses for formality.

The falling list matters as much as the rising one. 알아보다 "look into", 살펴보다 "take a look at", 도움이 되다 "be helpful", 많이 "a lot" and the framing ~에 관한 연구 "a study on" are the ordinary verbs and adverbs of Korean research prose, unremarkable and slightly colloquial. Models do not produce them (Table 7), and their share fell to between a fifth and a half of the trend. Whatever the prevalence of model use, the observable effect on the language of Korean abstracts is a compression of register: fewer plain expressions, more elevated ones, and a narrowing of the set of verbs used to state a finding.

The one prior measurement of Korean points the other way, and the difference has an explanation. Juzek's news study defines AI-associated words as the lemmas that GPT-4.1 over-produces when it continues news sentences, tracks them in the WMT News Crawl from 2020–2021 to 2023–2024, and finds Korean among the languages where they *fall*, by 22% relative to baseline words, alongside Persian and Japanese; the author offers differences in uptake, corpus composition (the News Crawl is heterogeneous web news without metadata) and language-specific overuse profiles as possible causes, notes that morphologically rich languages add tagging uncertainty, and flags the declines for dedicated follow-ups [11]. Three of our results bear on this. First, the register is different: a list built from news continuations is a news list, and the Korean signature we find is made of Sino-Korean verbs of academic argument (시사하다, 규명하다, 확인되다) and evaluative adjectives that news prose does not use, so the two lists barely overlap and neither study can see the other's shift. Second, the timing is different: the post period of [11] ends in 2024, and in our data Korean shows nothing in 2023 and only turns in the second half of 2024 (Figure 6), so a Korean news series ending in 2024 would be expected to show no rise, whatever the genre. Third, the genres differ in how text is produced: a news article is written by a staff journalist under a desk style and rarely passed through a chatbot, whereas an abstract is the most polished part of a paper, frequently written in two languages, and the part authors most often hand to a model "to polish"; the editing condition of the positive control (Section 5.6) shows that polishing alone installs the rising words. A news decrease and an academic increase in the same language are therefore compatible, and the comparison suggests that measurements of LLM style should be made register by register, with marker lists derived in the register being measured.

The one Korean measurement of scholarly text, by Koo, Kim and Kim [24], is consistent with ours once its endpoint and its units are taken into account. Their KCI series stops in 2024 and reports only modest increases in Korean expressions against sharp ones in the English abstracts of the same journals; Section 5.2 shows why: the Korean onset falls in the second half of 2024, a year after English, so a series ending in 2024 catches its first months only. Their units are inflected forms, which spread one lemma over several rows, a dispersion they themselves name as a possible reason for the weaker Korean signal and which the lemma units of Section 4.1 remove. The 2025 and 2026 values above are what that series would have found had it continued, and the Korean expressions they list as rising (중요한, 이러한, 시사한다, 강조한다, 기여할, 탐구하며) all correspond to marker units of Table 7. Their design also points to a comparison this paper does not make. KCI carries the English abstract of the same articles, and Koo et al. report a sharp 2024 rise of English style words in the same journals whose Korean abstracts had barely moved, which they read as model translation of the English by authors who wrote the Korean themselves; running the pipeline of Section 4 on the English abstracts of our journals would give a cross-language comparison with author, journal and article held fixed, which is stronger than the cross-corpus comparison with the English literature used above. Sections 5.12 and 5.14 make that comparison: the English side of the same articles moves in 2023, the Korean in late 2024, and the Korean shift is present in articles whose English abstract carries none of the period's markers, at 70% of the rate among articles whose English side does carry them, and 30 to 66% of that rate after the English indicator's sensitivity correction.

### 6.2 Marker turnover is consistent with model-generation turnover

Several words peaked in 2024–2025 and then fell back toward trend in 2026 while others kept rising (Section 5.2). The receding group (기여하다 "contribute", 강조하다 "emphasize", 탐구하다 "explore", 중요한 역할 "important role", 기여할 것으로 기대 "is expected to contribute") is the group that `gpt-4o-mini`, a 2024 model, produces most often in the positive control, and the group still rising in 2026 (시사하다, 규명하다, 확인되다, 구조적) is the one the 2026 models produce. This is consistent with a turnover driven by model generations, but it is not proof of one. The receding group was selected as the group with the largest 2024 excess, and selected extremes regress toward the mean in the following year by construction; the positive control that separates the two accounts rests on five models and two prompt styles. The Korean open-weight model of Section 5.6 is the observation that tips the balance: EXAONE 3.5, trained in Korea and released in December 2024, writes the receding group at or above the rates of the 2024 OpenAI model (기여하다 72% against 77%, 탐구하다 72% against 33%, 강조하다 53% against 35%) and the 2026 group at the rates of the 2026 OpenAI model, with drafts 12% longer than the OpenAI arm, which lifts all its rates somewhat but not their pattern; in the five models tested the two vocabularies therefore sort by the year of the model rather than by its provider, a pattern that regression to the mean does not predict; with five models and two prompt styles this is consistent with a generation effect rather than a demonstration of one. Claude Sonnet 5, a 2026 model from a third provider, sorts the same way, with the 2026 group at or above the real 2026 rates and the receding group at 6 to 20% (Section 5.6). The ordering by generation is clean for the receding and the rising groups; it is not clean for the falling vocabulary, where the two 2026 providers differ: at matched length Claude keeps 살펴보다 in 18% of its abstracts and OpenAI in 4%, on either side of the real 2026 rate of 11%, with EXAONE at 12%. Provider variance is therefore not zero, and what sorts by generation is which style words a model adds, not how much of the plain vocabulary it removes. The lexical fingerprint of LLM writing is therefore not a fixed list but a moving one, and a marker list frozen in 2024 would under-count 2026 use. The English full-text follow-up kept its 2024 word list and re-tuned only a frequency threshold on 2025 data [2]; the caution here goes one step further, and it argues for releasing frequency tables rather than word lists, so that the markers can be re-derived as models change.

### 6.3 What the lower bound does and does not say

The bounds of Section 5.9 say that at least the reported share of abstracts, 16% by the single-word statistic and 33% by the set statistic in 2026, carry wording that departs from the pre-LLM trend, and that under the attribution assumption A2 of Section 4.7 that departure is LLM-induced. Without A2 the same number is a floor on a register change whose author is unnamed; with it, a floor on direct LLM processing. Three of this paper's checks are attempts to break A2 rather than to assume it: the placebo years bound the trend error, the annotation and topic-matched comparisons of Section 5.13 make the reading in which the field's subject matter moved unlikely, and the within-article English comparison of Section 5.12 leaves the reading in which the Korean change is a by-product of an English one unable to carry it: the shift persists, reduced but present, where the English side carries no markers. It does not say that those abstracts were written by a model: an abstract drafted by the author, pasted into a chatbot "to polish the Korean", and returned with 시사한다 in place of 알아보았다 counts fully, as does an abstract written from scratch by the model. It also does not distinguish machine text from human writers who have absorbed the style through exposure, a process now documented for spoken English [12]. The within-document test of Section 5.8 does not separate these two processes, and we do not use it to. What it is consistent with is a document-level switch rather than a word-level drift: an abstract that con-tains a rising marker is less likely to contain a falling one, which is what happens when whole abstracts are rewritten into the newer register, by a model or by an author who has adopted that register wholesale, and not what happens when individual words spread through the community one at a time. An author who switched register without any tool would produce the same pattern. What makes a purely human-diffusion account less parsimonious is different evidence: the turnover of Section 6.2 is faster than word-of-mouth register change is usually described as spreading, though we do not model that speed, the markers that rose are those the models of each period produce, and the falling words fell in step with the rising ones rather than after them; none of this identifies the share attributable to direct model processing. Finally, the bound is a floor, not an estimate, and two observations suggest, without identifying it, that the true share sits above it; the mixture estimator of Section 5.15, which reads 72.1% for 2026, is one way of saying how far above, under its own assumption that every processed abstract carries a set word. In English the excess-vocabulary floor of 13.5% for 2024 abstracts sits next to mixture-model estimates of 17 to 35% for computer-science preprints [3, 4], a full-text estimate of 57% of 2025 articles across four publishers [19] and 89% for the full text of late-2025 biomedical papers [2] (53% for their abstracts). And our own length-matched positive controls write their most frequent marker into 41 to 88% of the abstracts they produce, so a corpus processed entirely in that way would still yield a single-word bound below 100. We do not replace the floor by a prevalence, because each conversion assumes something about the processing, and Section 5.10 shows how far the answer moves with that assumption: from 77–97% of abstracts under our matched-length drafting conditions (propagated ranges from 58% to above 100%), through 66–90% on the two headline words under rewriting by the two 2026 API models, to no feasible value under our polishing conditions, which cannot produce the observed vocabulary at any prevalence. Topaz and Bahl are right that a lexical shift is not a calibrated prevalence meter [20]; we report the floor as the headline because it needs no calibration, and the scenario analysis of Section 5.10 as what a calibration would say and how much it depends on what is assumed.

### 6.4 Vietnamese: the same words, one year later

The Vietnamese corpus shows a suggestive but not yet floor-clearing pattern, late and small. Through 2024 nothing rises beyond the negative-control level; in 2025 the excess is carried by nhấn mạnh "emphasize", then chốt "key", linh hoạt "flexible", kỷ nguyên "era", cơ chế "mechanism" and toàn diện "comprehensive", the Vietnamese counterparts of the Korean and English lists. The data admit three explanations and cannot separate them: later adoption of chatbots for Vietnamese academic writing, weaker Vietnamese generation quality in 2023–2024 models that made model text less usable, and the smaller corpus (5,411 panel abstracts in 2025 against 40,000 for Korean), which raises the floor of what can be detected. The 2026 VJOL year, once indexed, will discriminate between these.

### 6.5 Implications for editors, reviewers and teachers

The practical use of these tables is not detection of individual papers, which the method cannot do and which we do not recommend. It is calibration. A reviewer who notices 시사한다, 규명하고자, 구조적, 단순한 X를 넘어 and 이러한 결과는 in one abstract is looking at a pattern that, in 2026, is several times more common than it was in 2022, and the tables give the base rates needed to say so precisely. They say nothing about whether that abstract was produced by a model; the public checker named in the data availability statement marks the same words in a pasted text and carries the same limit. For teachers of academic Korean the falling list is the more useful one: it names the ordinary expressions that students exposed to model output are losing. For journals the field table shows where the change is fastest, and the lower bound gives a defensible number for policy documents in place of English figures.

## 7 Limitations

*Abstracts only.* Everything here is measured on abstracts, the most polished and most frequently machine-edited part of a paper; full-text prevalence in English runs 1.3 to 2.1 times abstract prevalence across the years of [2], and the same direction is likely here. *Coverage.* KCI accredits a specific set of journals and our harvest is a large but not exhaustive sample of them; year sizes differed in the first harvest because of identifier-range walking, which a second harvest corrected; the main text uses the completed corpus, and the correction moved the 2025 single-word bound up by 1.1 pp, the set bounds up by as much as 1.9 pp, and the other single-word headline statistics by at most 0.2 (Appendix H), and the 2026 year covers January to August of a year that is still being indexed. The time axis is the publication month, not the writing date: an abstract published in early 2026 may have been written in 2025, so the year-level bounds lag the underlying writing behaviour by the publication delay, and the agreement between each model generation's vocabulary and the abstracts published in its period (Section 6.2) is a consistency pattern, not a demonstration of temporal precedence. VJOL covers a fraction of Vietnamese journals with coverage that grows over time. *Fields.* Field assignment is by keyword rules on journal names; about 30% of panel journals fall into "other", and interdisciplinary journals are placed by their first matching keyword. *Tokenisation.* The Korean analysis depends on one morphological analyser and on two fusion rules we chose (noun + verbalising suffix, and noun + the adjectival suffixes 적, 화, 성); a different analyser or a different fusion choice would move some units between the style and content classes. Appendix E measures how much this matters: switching the adjectival fusion off leaves the bound identical in every year and exchanges eleven of the thirty units in Table 3 for as many others, so the choice affects which words are named but not the size of the effect. The Vietnamese tagger has lower accuracy than the Korean analyser, and we did not manually verify segmentation. *Positive control.* Three OpenAI models, one Anthropic model and one Korean open-weight model were used. The first three conditions produced abstracts shorter than real ones (313 to 509 characters against about 690) and document frequency grows with length, so their rates are downward-biased and not comparable across conditions; a length-matched condition and an editing condition were added for this reason, and both were repeated with EXAONE 3.5 7.8B on our own hardware. The Korean model's drafts are 12% longer than the OpenAI matched-length arm and its edits 11% shorter than their originals, and a 7.8-billion-parameter model quantised to four bits is not the tool most Korean authors use; HyperCLOVA X, the other widely used Korean model, is available only through an API we did not call; its small open-weight sibling, HyperCLOVA X SEED 1.5B, was tried locally and set aside (Table 20). *Calibration.* The sensitivities of Section 5.10 come from our own prompts and from models available in August 2026; the mix of models and prompts in the field is unknown, no declared-use ground truth exists for KCI abstracts, and the implied shares are conditional on the processing assumed. They bracket the floor and do not replace it, and the rewriting condition shows that the implied share depends on the breadth of the set as much as on the prompt: a processing that reproduces the headline words but not the rare lemmas gives a feasible share on the two-word set and none on the 47-lemma set. *Cross-language comparison.* The within-article comparison of Section 5.12 cannot separate a model that wrote both abstracts from a model that translated one into the other, its English sets are word forms rather than lemmas, and the English abstracts of 2026 are 79% complete. *Base period.* Five base years is shorter than the ten used for English; the negative control shows that it is enough for words with stable pre-2023 frequencies, but slow pre-existing trends (for instance the growth of 종합적으로 "comprehensively" from 2018 onward) are only partly removed. *Annotation.* The style/topic classification of Section 5.13 was made by three language models from a written rubric, not by human linguists, and their agreement is only fair ($\kappa = 0.24$); it is a sensitivity analysis on the composition of the set, not a validated linguistic annotation, and the restricted bounds should be read as such. *English indicator.* The English marker set of Section 5.12 has a sensitivity of its own, measured here only on abstracts written from a title or translated from Korean by three models; the corrected figures of Table 19 span 30 to 66% of the touched rate, so the direction of that comparison is robust and its size is not. *Translation.* The features of Section 5.14 are surface marks, not a translation detector, and they cannot exclude a careful translation that leaves none of them. *Marker selection.* The marker units were fixed on an earlier pass over the same corpus, not pre-registered; the split-half design of Appendix F and the selection-aware bootstrap of Section 5.8 bound the effect of that choice on the headline statistics, but a pre-registered hold-out would have been cleaner. *Causality.* The design attributes excess to LLMs by timing and by the positive control; it cannot exclude that some of the change reflects Korean journals adopting structured or English-modelled abstract conventions in the same years, although the 2022 control, which predates the change, and the fall of plain words, which such conventions would not cause, make this an unlikely full explanation.

## 8 Conclusion

An excess-vocabulary analysis, adapted to morphological units, transfers to Korean and Vietnamese and finds a large LLM signature in Korean scholarly abstracts: nothing in 2023, an onset in the second half of 2024, a steep rise through 2025 that flattens in the second quarter of 2026, several style words at four to eleven times their trend, a matching collapse of plain expressions, a strong negative association between the two vocabularies inside single abstracts (consistent with a document-level register shift rather than word-by-word diffusion, without saying by whom), a marker set that turns over with model generations, a conditional lower bound (under the assumptions of Section 4.7) on LLM-processed abstracts, restricted to units with excess ratio at least 1.5, of 3.5% in 2024, 10.5% in 2025 and 16.1% in 2026 (January to August; 15.9%, 95% CI 14.7 to 17.1, under a journal-cluster bootstrap that reselects the maximum), and a set-based bound, with the set chosen on other journals than it is measured on, of 7.8%, 20.6% and 33.0% (30.8 to 35.4), against placebo levels of at most 2.2 points for the single-word statistic and 2.9 for the set statistic at every extrapolation horizon the pre-LLM data allow. The mixture estimator of Holzwarth, González-Márquez and Kobak [2], rebuilt under the same split-half discipline, gives 41.9% for 2025 and 72.1% for 2026, and 48.6% and 81.1% when its threshold is chosen on the measurement half (Section 5.15). A sensitivity measurement on 1,437 control abstracts from three providers puts the 2026 share at 77–97% in the scenario in which abstracts were processed as model drafting processes them (propagated ranges from 58% to above 100%), at 66–90% on the two headline words in the scenario of rewriting from a draft by the 2026 API models, and shows that light polishing alone cannot produce the observed vocabulary; these are scenarios conditioned on one prompt and one model, not an estimate of prevalence. The result survives placebo target years at every horizon, alternative expectation models, a split-half separation of word choice from measurement, removal of the fusion rule, equal weighting by journal, a strict journal panel, and month-composition matching, and it is present in nine of every ten of the journals with enough abstracts to tell. A preliminary Vietnamese comparison shows the same family of words a year later. The English abstracts of the same articles moved a year before their Korean abstracts, turn their markers over on the same timetable, and show that the Korean shift is not confined to articles whose English abstract carries model vocabulary, at 30 to 66% of that rate once the English indicator's own sensitivity is taken into account. Restricting the marker set to lemmas that three language-model annotators, working independently of one another, all call style leaves 14.7 of the 33.0 points, pairing each 2026 abstract with the closest base-period abstract of the same journal leaves the difference unchanged at 34.1 points, and the surface marks of translated Korean fall while the markers rise, so neither the tested subject-matter controls nor the tested translation routes explain most of the shift. The frequency tables, code and a public dictionary of the most affected Korean expressions are released with this paper so that the measurement can be repeated as the models, and the language, keep changing.

## 9 Declaration of language-model use

This paper was written with heavy use of language models. What follows is the division of labour, put in the body because it bears on how the results should be read. This section was drafted the same way as the rest of the manuscript, and the author has read and approved every sentence of it.

The division of work was as follows. The author conceived the study, chose the question, set and repeatedly redirected the analysis, required the verification passes reported in Appendix J, and is solely responsible for the claims made here. The author does not write code. The analysis code was written and executed, and the English text of this manuscript was drafted, by the Claude Code command-line agent (Anthropic), running in terminal sessions that the author directs in Korean; Table 20 lists every model that took part and its role. Review passes over the manuscript, including those recorded in the version history of Appendix J, were also run with language models on the author's instruction. The classification of marker lemmas as style or topic used three language models and is described where it is used, in Section 5.13. That is a measurement, not a writing aid, and it is the only place where a model acts as a judge. Models are also used as subjects: the control abstracts of Section 5.6, the sensitivity conditions of Section 5.10 and the translation condition of Section 5.14 are all model output, and the numbers those sections report depend on it.

The pipeline was this. Between 13 August 2026 and 4 September 2026 the author opened 55 agent sessions that touched this project (counted up to 10:40 UTC on 4 September 2026 over every session whose transcript mentions the project folder, the sessions that reviewed and assembled this version included; the counting script is in the deposit, the transcripts are not) and typed 1,718 Korean instructions, questions and decisions into them; the agent answered in 52,904 turns and made 29,036 tool calls, harvesting the corpora, writing and running Python and JavaScript, and writing the English of every version as HTML. The models behind the agent changed with availability, Claude Opus 4.8, Claude Opus 5, Claude Fable 5 and Claude Fable 5.1, with Claude Sonnet 5 for delegated sub-tasks. The headline statistics, every table and the 969 tracked values in the running text are inserted by a fill script from the result files, and a consistency check refuses to assemble the paper while a filled value has no source or differs from its result file, a retired claim is still asserted or a tracked number has gone stale. The other measured values in the running text, among them the rates quoted for single words and conditions from Tables 7, 18 and A2 and the monthly and density figures of Sections 5.8 and 5.11, were typed by the agent from those tables and result files and checked by reading, not by machine; values quoted from other papers, model names and dates are typed from their sources. Review passes were of two kinds: adversarial readings by the same agent, prompted to read as a hostile referee, and readings of the PDF by ChatGPT (GPT-5.6 Sol, OpenAI) that the author ran on the author's own account and pasted back to the agent; the findings of both were addressed by the agent, and the version history of Appendix J records the second kind as ChatGPT readings. No one other than the author and one researcher who commented by email had read the manuscript before this version, and the author reads it through Korean summaries and translations that the agent produces. The fractions are 100% of the analysis code, 100% of the build and gate scripts, and 100% of the English text, this section included. The author wrote none of the code and none of the English; what the author wrote is the Korean on the other side of the terminal, the 1,718 instructions, the choice of the question, the rejection of several approaches, the demand for each verification pass recorded in Appendix J, and the approval of every version.

**Table 20:** Language models used in producing this paper, by role. The agent is the Claude Code command-line client (Anthropic); the models behind it changed with availability over the sessions. The roles of generator, translator and annotator are the ones the named sections describe.

| Model | Provider, access | Role | Where |
|---|---|---|---|
| Claude Opus 4.8; Claude Opus 5; Claude Fable 5; Claude Fable 5.1 | Anthropic, through the Claude Code agent | Analysis code, build and gate scripts, the English text of every version, adversarial review passes; Claude Opus 5 also served as one of the three annotators of Section 5.13; no record ties a passage or a script to one of the four | Whole paper; Appendix J |
| Claude Sonnet 5 | Anthropic, Claude Code sub-tasks and non-interactive runs | Delegated sub-tasks of the agent; control generator (edit, rewrite, long, translation); English-abstract generator; annotator | Sections 5.6, 5.10, 5.12, 5.13, 5.14 |
| GPT-5.6 Sol (ChatGPT) | OpenAI, the author's account | Readings of the PDF between versions, run by the author and pasted to the agent (the ChatGPT readings of Appendix J) | Appendix J |
| gpt-5.6-luna | OpenAI API | Control generator (draft, edit, rewrite, long, translation); English-abstract generator; annotator | Sections 5.6, 5.10, 5.12, 5.13, 5.14 |
| gpt-5.6-terra | OpenAI API | Control generator (draft) | Sections 5.6, 5.10 |
| gpt-4o-mini | OpenAI API | Control generator (draft, edit) | Sections 5.6, 5.10 |
| EXAONE 3.5 7.8B Instruct | LG AI Research, run locally with llama.cpp (Q4_K_M) | Control generator (edit, rewrite, long, translation); English-abstract generator | Sections 5.6, 5.10, 5.12, 5.14 |
| EXAONE 4.0 1.2B; HyperCLOVA X SEED 1.5B | LG AI Research; Naver, run locally with llama.cpp | Trial control generation (63 and 40 long-form abstracts), set aside for output quality and used in no table | None |

Generated by the agent: 100% of the analysis code, 100% of the build and gate scripts, 100% of the English text, this section and this table included. The author wrote none of the code and none of the English.

The arithmetic is a separate matter. Every corpus statistic here is produced by the deposited code from the deposited frequency tables and result files, and the package rebuilds every table and every figure from a clean directory, so the counts can be checked without trusting the prose. The prose is not separable in the same way. Which estimator to use, what to make of an annotator agreement of 0.24, whether a topic-matched shift of 0.1 points settles anything: those are readings, they are not in the archive, and they deserve the same doubt as the sentences.

The paper's own English can be measured with the instrument the paper uses on others. Against the fixed 33-word English marker list of Section 5.12 (the second block of Table 15), the running text of this manuscript, this section included, contains 37 marker occurrences in 24,846 words, a rate of 14.9 per ten thousand, against 5.7 for KCI English abstracts of 2018–2022 and 58.5 for the English abstracts written by the models of Section 5.12 (both computed by the same script on 8,000 base-period abstracts and the 257 model abstracts). The tables are left out because they hold the same words as data, and the reference list because its titles are other people's sentences and one of them contains *delving*. Each of the 37 occurrences was checked in context: every one is a mention, quoted as a marker, listed, reported as a frequency or given as the gloss of a Vietnamese word, and none is used in its own right; *delve*, *underscore*, *intricate* and *showcase* appear in this paper only because it lists them. The count says more about that list than about the prose, which this section has already declared to be model-written; the 47-word set of the first block, built from KCI English and admitting ordinary words such as *alongside* and *offers*, is not used for the count, and the manuscript does use several of its members in their own right. A register shift is carried by sen-tence shape and rhythm as much as by vocabulary, and the lexical statistic used throughout this paper is blind to the first two, which limits the corpus results of Section 5.9 exactly as much as it limits this paragraph.

## Data availability

Per-year document-frequency tables for both languages, journal panel lists, field rules, the 1,437 generated Korean positive-control abstracts and 514 generated English control outputs (Section 5.12), the harvesting and analysis code, and the figure and table generators are archived at Zenodo (10.5281/zenodo.22303588 for version 8; all versions under 10.5281/zenodo.22102389) and downloadable now at `https://os.intframe.com/report/ai-style-lexicon/reproducibility.zip` (51 MB). The paper is at `https://os.intframe.com/report/ai-style-lexicon`, the public dictionary at `https://os.intframe.com/report/ai-style-dictionary-ko` and the style checker at `https://os.intframe.com/report/ai-style-check-ko`. Raw abstracts are not redistributed; they can be re-harvested from the public KCI and VJOL OAI-PMH endpoints with the included scripts.

## Appendix A · Korean marker words with bootstrap intervals

**Table A1:** Bootstrap intervals for all 65 units that appear in Tables 3 and 4 or were fixed in the pilot pass. Point estimates and 95% intervals from 1,000 replicates in which whole journals are resampled with replacement once per replicate, each selected journal carrying its abstracts in every base and target year (longitudinal cluster bootstrap), with the base-year fit and the extrapolation recomputed in every replicate, on at most 12,000 abstracts per year. Base years 2018–2022. The units were selected on an earlier pass over the same corpus, so these intervals are not free of selection. Point estimates come from the 12,000-abstract subsample and therefore differ from the corpus-wide values in Tables 3 and 4, which use up to 40,000 abstracts per year, and in two cases the corpus-wide value lies outside the bracket (marked † there); the corpus-wide values are the ones quoted in the text.

| Unit | Ratio 2024 | Ratio 2025 | Ratio 2026 | Excess pp 2025 | Excess pp 2026 |
|---|---|---|---|---|---|
| 시사하다 *suggest, imply* | 1.38 [1.24, 1.57] | 2.48 [2.16, 2.90] | 3.97 [3.44, 4.67] | +7.8 [6.8, 8.8] | +16.1 [14.7, 17.4] |
| 확인되다 *be confirmed* | 1.22 [1.10, 1.39] | 1.90 [1.67, 2.19] | 3.36 [2.88, 4.01] | +5.7 [4.7, 6.7] | +15.2 [13.8, 16.7] |
| 핵심 *core, key* | 1.08 [0.94, 1.22] | 1.85 [1.63, 2.11] | 3.03 [2.68, 3.52] | +5.2 [4.3, 6.1] | +12.6 [11.5, 13.8] |
| 구조적 *structural* | 1.03 [0.85, 1.24] | 2.49 [2.09, 3.07] | 5.71 [4.72, 7.23] | +3.9 [3.2, 4.6] | +12.5 [11.6, 13.5] |
| 규명하다 *elucidate* | 1.09 [0.93, 1.29] | 2.07 [1.75, 2.59] | 4.12 [3.38, 5.29] | +4.1 [3.2, 5.0] | +11.6 [10.5, 12.8] |
| 단순하다 *mere, simple* | 1.31 [1.02, 1.71] | 5.21 [4.07, 6.81] | 8.13 [6.11, 11.57] | +6.4 [5.6, 7.2] | +10.7 [9.8, 11.6] |
| 특히 *in particular* | 1.26 [1.16, 1.37] | 1.70 [1.54, 1.88] | 1.68 [1.51, 1.90] | +10.3 [8.8, 12.1] | +9.9 [8.1, 11.8] |
| 넘다 *go beyond* | 1.29 [1.10, 1.56] | 2.83 [2.36, 3.46] | 4.02 [3.28, 5.04] | +5.4 [4.6, 6.3] | +9.1 [8.1, 10.0] |
| 이러한 결과 *these results* | 1.30 [1.12, 1.52] | 1.84 [1.58, 2.17] | 3.32 [2.79, 4.06] | +3.2 [2.5, 3.9] | +9.0 [7.8, 10.1] |
| 실증적 *empirical* | 1.26 [1.05, 1.55] | 3.44 [2.80, 4.37] | 4.51 [3.59, 5.95] | +5.8 [4.9, 6.7] | +8.2 [7.3, 9.2] |
| 보여주다 *show* | 1.13 [1.01, 1.26] | 1.55 [1.36, 1.75] | 2.02 [1.76, 2.33] | +4.5 [3.3, 5.6] | +8.2 [6.9, 9.5] |
| 검토하다 *examine, review* | 1.07 [0.97, 1.18] | 1.25 [1.11, 1.39] | 1.79 [1.60, 2.02] | +2.2 [1.1, 3.2] | +7.0 [5.8, 8.3] |
| 기능하다 *function as* | 1.67 [1.22, 2.36] | 5.10 [3.62, 7.58] | 9.78 [6.70, 14.67] | +3.0 [2.5, 3.5] | +6.1 [5.4, 6.8] |
| 결합하다 *combine* | 1.01 [0.83, 1.19] | 1.76 [1.46, 2.15] | 3.45 [2.87, 4.32] | +1.7 [1.2, 2.3] | +6.0 [5.3, 6.6] |
| 지니다 *have, bear* | 1.01 [0.89, 1.15] | 1.30 [1.13, 1.50] | 2.02 [1.72, 2.39] | +1.7 [0.9, 2.6] | +5.9 [4.8, 7.0] |
| 제도적 *institutional* | 1.03 [0.84, 1.27] | 2.35 [1.95, 2.96] | 3.43 [2.74, 4.53] | +3.1 [2.5, 3.8] | +5.7 [4.9, 6.4] |
| 통합적 *integrative* | 1.24 [0.98, 1.60] | 3.09 [2.46, 4.10] | 5.29 [3.96, 7.61] | +2.8 [2.3, 3.3] | +5.6 [4.9, 6.3] |
| 형성되다 *be formed* | 1.06 [0.88, 1.30] | 1.42 [1.16, 1.82] | 2.94 [2.32, 3.95] | +1.1 [0.5, 1.8] | +4.9 [4.0, 5.8] |
| 작동하다 *operate* | 1.17 [0.91, 1.55] | 1.99 [1.55, 2.63] | 4.79 [3.64, 6.75] | +1.3 [0.8, 1.7] | +4.9 [4.2, 5.6] |
| 체계적 *systematic* | 1.34 [1.17, 1.56] | 2.13 [1.83, 2.53] | 2.31 [1.95, 2.80] | +4.1 [3.3, 4.9] | +4.8 [4.0, 5.7] |
| 이론적 *theoretical* | 1.07 [0.93, 1.26] | 1.69 [1.43, 2.01] | 2.12 [1.75, 2.64] | +3.0 [2.1, 3.7] | +4.8 [3.8, 5.9] |
| 해석하다 *interpret* | 1.02 [0.88, 1.20] | 1.44 [1.23, 1.71] | 2.29 [1.91, 2.84] | +1.5 [0.9, 2.2] | +4.6 [3.8, 5.5] |
| 확장하다 *extend* | 1.03 [0.88, 1.22] | 1.54 [1.30, 1.83] | 2.47 [2.07, 2.99] | +1.5 [1.0, 2.1] | +4.4 [3.7, 5.2] |
| 작용하다 *act on* | 1.13 [0.97, 1.32] | 1.93 [1.64, 2.31] | 2.08 [1.75, 2.57] | +3.5 [2.7, 4.3] | +4.1 [3.2, 5.0] |
| 형성하다 *form* | 1.05 [0.89, 1.24] | 1.44 [1.23, 1.72] | 2.12 [1.79, 2.59] | +1.5 [0.9, 2.2] | +4.0 [3.2, 4.9] |
| 강화하다 *strengthen* | 1.66 [1.45, 1.91] | 2.26 [1.93, 2.68] | 2.11 [1.75, 2.60] | +4.5 [3.7, 5.2] | +4.0 [3.2, 4.8] |
| 상대적 *relative* | 0.91 [0.79, 1.05] | 1.18 [1.00, 1.39] | 1.95 [1.64, 2.40] | +0.7 [0.0, 1.4] | +3.9 [3.0, 4.8] |
| 메커니즘 *mechanism* | 1.35 [1.05, 1.83] | 2.09 [1.62, 2.81] | 4.23 [3.12, 5.95] | +1.2 [0.8, 1.6] | +3.9 [3.2, 4.6] |
| 통합하다 *integrate* | 1.71 [1.29, 2.42] | 3.13 [2.31, 4.59] | 5.23 [3.68, 8.56] | +1.9 [1.5, 2.4] | +3.8 [3.2, 4.3] |
| 실질적 *substantive* | 1.33 [1.13, 1.56] | 2.62 [2.19, 3.18] | 2.28 [1.87, 2.87] | +4.6 [3.9, 5.3] | +3.7 [2.9, 4.4] |
| 실천적 *practical* | 0.98 [0.79, 1.22] | 2.26 [1.81, 2.98] | 2.65 [2.10, 3.55] | +2.7 [2.1, 3.3] | +3.6 [3.0, 4.3] |
| 구조화 *structuring* | 0.97 [0.77, 1.23] | 1.56 [1.24, 2.05] | 3.15 [2.43, 4.42] | +0.9 [0.5, 1.3] | +3.6 [2.9, 4.2] |
| 결합되다 *be combined* | 1.46 [0.97, 2.35] | 2.97 [1.87, 4.51] | 8.15 [4.66, 11.46] | +1.0 [0.7, 1.4] | +3.4 [2.9, 4.0] |
| 정서적 *emotional* | 0.99 [0.81, 1.27] | 1.99 [1.63, 2.57] | 2.54 [2.01, 3.40] | +2.1 [1.6, 2.7] | +3.4 [2.7, 4.1] |
| 결론적 *conclusive* | 1.25 [1.01, 1.59] | 1.76 [1.41, 2.36] | 2.86 [2.18, 3.98] | +1.4 [0.9, 2.0] | +3.3 [2.6, 4.0] |
| 충분히 *sufficiently* | 0.98 [0.80, 1.20] | 1.43 [1.16, 1.79] | 2.51 [2.05, 3.24] | +0.9 [0.4, 1.4] | +3.2 [2.6, 3.8] |
| 기여하다 *contribute* | 1.80 [1.61, 2.03] | 2.44 [2.16, 2.82] | 1.59 [1.37, 1.90] | +7.0 [6.1, 7.9] | +2.9 [2.0, 3.8] |
| 기제 *mechanism (Sino-Korean)* | 0.97 [0.72, 1.36] | 1.42 [1.05, 2.03] | 3.82 [2.74, 5.92] | +0.4 [0.0, 0.7] | +2.5 [2.0, 3.0] |
| 촉진하다 *promote* | 1.70 [1.38, 2.10] | 2.19 [1.78, 2.76] | 1.96 [1.56, 2.60] | +2.0 [1.5, 2.5] | +1.7 [1.2, 2.3] |
| 조명하다 *shed light on* | 1.47 [1.14, 1.93] | 3.00 [2.33, 4.09] | 2.29 [1.65, 3.46] | +2.6 [2.1, 3.0] | +1.7 [1.1, 2.2] |
| 강조하다 *emphasize* | 1.72 [1.49, 1.97] | 2.19 [1.89, 2.61] | 1.36 [1.13, 1.67] | +5.1 [4.2, 6.0] | +1.5 [0.6, 2.3] |
| 핵심적 *core, key* | 1.25 [0.98, 1.61] | 1.71 [1.29, 2.36] | 2.02 [1.52, 2.79] | +0.9 [0.5, 1.4] | +1.3 [0.9, 1.8] |
| 중요한 역할 *important role* | 3.38 [2.57, 4.85] | 3.63 [2.50, 5.06] | 2.06 [1.30, 2.68] | +1.9 [1.5, 2.4] | +0.7 [0.3, 1.1] |
| 탐구하다 *explore* | 1.90 [1.60, 2.22] | 1.78 [1.48, 2.18] | 1.11 [0.91, 1.39] | +2.1 [1.5, 2.7] | +0.3 [−0.3, 0.9] |
| 노력하다 | 1.25 [0.98, 1.59] | 0.67 [0.50, 0.95] | 0.43 [0.30, 0.68] | −0.5 [−0.9, −0.0] | −0.8 [−1.2, −0.3] |
| 변화되다 | 0.87 [0.68, 1.10] | 0.59 [0.46, 0.80] | 0.46 [0.33, 0.68] | −0.6 [−0.9, −0.2] | −0.8 [−1.2, −0.4] |
| 사료되다 | 1.01 [0.80, 1.34] | 0.66 [0.49, 0.93] | 0.42 [0.28, 0.65] | −0.5 [−0.9, −0.1] | −0.8 [−1.3, −0.3] |
| 일어나다 | 0.93 [0.74, 1.18] | 0.68 [0.52, 0.93] | 0.47 [0.33, 0.70] | −0.5 [−1.0, −0.1] | −0.9 [−1.3, −0.3] |

| Unit | Ratio 2024 | Ratio 2025 | Ratio 2026 | Excess pp 2025 | Excess pp 2026 |
|---|---|---|---|---|---|
| 끼치다 | 0.90 [0.69, 1.18] | 0.53 [0.38, 0.75] | 0.30 [0.20, 0.46] | −0.6 [−1.0, −0.3] | −0.9 [−1.3, −0.6] |
| 나오다 | 0.79 [0.61, 1.05] | 0.55 [0.40, 0.78] | 0.32 [0.22, 0.49] | −0.7 [−1.2, −0.3] | −1.0 [−1.5, −0.6] |
| 소개하다 | 1.03 [0.80, 1.33] | 0.67 [0.51, 0.91] | 0.39 [0.26, 0.60] | −0.6 [−1.1, −0.1] | −1.1 [−1.6, −0.5] |
| 세계적 | 0.93 [0.74, 1.20] | 0.65 [0.50, 0.85] | 0.39 [0.28, 0.54] | −0.7 [−1.2, −0.3] | −1.3 [−1.9, −0.8] |
| 생각하다 | 1.01 [0.84, 1.23] | 0.73 [0.57, 0.93] | 0.43 [0.31, 0.59] | −0.7 [−1.3, −0.1] | −1.4 [−2.1, −0.8] |
| ~에 관한 연구 *a study on* | 0.94 [0.77, 1.15] | 0.47 [0.37, 0.60] | 0.30 [0.24, 0.40] | −1.2 [−1.6, −0.8] | −1.7 [−2.1, −1.2] |
| 느끼다 | 1.03 [0.87, 1.25] | 0.70 [0.56, 0.87] | 0.37 [0.29, 0.50] | −0.8 [−1.3, −0.3] | −1.7 [−2.3, −1.1] |
| 많이 *a lot* | 0.93 [0.78, 1.10] | 0.81 [0.68, 1.02] | 0.37 [0.28, 0.49] | −0.6 [−1.2, 0.1] | −2.0 [−2.7, −1.3] |
| 연구하다 | 1.02 [0.88, 1.20] | 0.72 [0.60, 0.88] | 0.32 [0.25, 0.41] | −0.9 [−1.5, −0.3] | −2.2 [−2.9, −1.6] |
| 도움이 되다 *be helpful* | 1.03 [0.88, 1.22] | 0.57 [0.45, 0.72] | 0.22 [0.17, 0.30] | −1.3 [−2.0, −0.8] | −2.5 [−3.1, −1.9] |
| 알다 | 0.96 [0.85, 1.08] | 0.67 [0.57, 0.77] | 0.40 [0.33, 0.48] | −2.4 [−3.4, −1.4] | −4.1 [−5.2, −3.1] |
| 알아보다 *look into* | 0.82 [0.72, 0.94] | 0.53 [0.44, 0.63] | 0.20 [0.16, 0.26] | −2.9 [−3.8, −2.0] | −4.8 [−5.9, −3.7] |
| 기대하다 | 0.90 [0.80, 1.01] | 0.61 [0.54, 0.70] | 0.33 [0.28, 0.39] | −2.9 [−3.9, −2.1] | −5.4 [−6.3, −4.5] |
| 보다 | 0.88 [0.80, 0.96] | 0.63 [0.56, 0.70] | 0.37 [0.32, 0.42] | −4.1 [−5.1, −3.0] | −6.7 [−8.0, −5.5] |
| 많다 | 0.92 [0.83, 1.01] | 0.66 [0.59, 0.75] | 0.36 [0.31, 0.42] | −4.2 [−5.6, −2.8] | −7.8 [−9.3, −6.3] |
| 진행하다 *carry out* | 0.90 [0.82, 0.98] | 0.63 [0.58, 0.70] | 0.29 [0.26, 0.33] | −4.0 [−4.8, −3.1] | −8.1 [−9.2, −7.0] |
| 살펴보다 *take a look at* | 0.88 [0.83, 0.93] | 0.63 [0.59, 0.68] | 0.41 [0.37, 0.45] | −8.6 [−10.0, −7.1] | −14.2 [−16.0, −12.4] |

## Appendix B · Lexical density

**Table A2:** Lexical density by year. The first two columns count distinct units per abstract directly, with no minimum-count filter, on a fixed sample of 12,000 abstracts per year, so they do not depend on how many abstracts a year contributes; the last two columns sum document frequencies over the units that pass the filter, the quantity that actually enters the excess calculation, and therefore fall in the years with fewer abstracts. Mean abstract length is given for the same fixed sample.

| Year | n (fixed) | Length (chars) | Distinct lemmas | Lemmas, 600–800 chars | n (main) | Filtered lemmas | Filtered bigrams |
|---|---|---|---|---|---|---|---|
| 2018 | 12,000 | 682 | 87.8 | 88.6 | 40,000 | 86.4 | 55.5 |
| 2019 | 12,000 | 681 | 87.4 | 88.5 | 39,419 | 86.0 | 54.9 |
| 2020 | 12,000 | 678 | 87.4 | 88.8 | 40,000 | 86.1 | 55.3 |
| 2021 | 12,000 | 681 | 87.7 | 88.8 | 37,665 | 86.4 | 55.2 |
| 2022 | 12,000 | 677 | 87.4 | 89.1 | 40,000 | 86.1 | 55.6 |
| 2023 | 12,000 | 676 | 87.2 | 89.0 | 27,437 | 85.2 | 51.2 |
| 2024 | 12,000 | 685 | 88.5 | 89.1 | 40,000 | 87.0 | 57.2 |
| 2025 | 12,000 | 683 | 91.9 | 93.1 | 40,000 | 90.7 | 59.6 |
| 2026 | 12,000 | 698 | 97.8 | 98.0 | 32,630 | 95.9 | 58.8 |

**Table A3:** Excess in pp and 2026 excess ratio of eleven marker words before and after density normalisation (Section 5.8, Table 12). Common words absorb the largest part of the density inflation in percentage points (분석하다, expected in 41% of 2025 abstracts, loses 2.5 pp of its 2025 excess); the rare style words and the falling words move little, and the falling words fall slightly further after normalisation.

| Word | 2025 raw | 2025 normalised | 2026 raw | 2026 normalised | Ratio 2026 raw | Ratio 2026 normalised |
|---|---|---|---|---|---|---|
| 시사하다 *suggest, imply* | +7.8 | +7.2 | +16.1 | +13.8 | 4.04 | 3.61 |
| 확인되다 *be confirmed* | +5.1 | +4.5 | +14.1 | +11.8 | 2.91 | 2.60 |
| 구조적 *structural* | +4.4 | +4.1 | +12.6 | +11.0 | 5.96 | 5.32 |
| 규명하다 *elucidate* | +4.0 | +3.6 | +11.6 | +10.0 | 3.92 | 3.50 |
| 단순하다 *mere, simple* | +6.2 | +5.9 | +11.2 | +9.9 | 10.98 | 9.79 |
| 특히 *in particular* | +10.5 | +9.3 | +10.9 | +8.2 | 1.78 | 1.58 |
| 분석하다 *analyse* | +11.0 | +8.5 | +14.4 | +8.3 | 1.34 | 1.20 |
| 중요하다 *be important* | +4.0 | +3.1 | −1.2 | −2.6 | 0.92 | 0.82 |
| 살펴보다 *take a look at* | −7.8 | −8.6 | −13.5 | −14.6 | 0.42 | 0.38 |
| 알아보다 *look into* | −2.7 | −2.9 | −4.4 | −4.5 | 0.24 | 0.21 |
| 진행하다 *carry out* | −4.1 | −4.4 | −8.1 | −8.5 | 0.31 | 0.28 |

Section 5.11 discusses the rise in this quantity as a finding. Here it is treated as a nuisance for the excess statistic: because the analysis counts document frequencies, a general increase in how many distinct units an abstract contains raises every word's frequency a little. The first pair of columns counts distinct units per abstract directly, with no minimum-count filter, on a fixed sample of 12,000 abstracts per year, so the value does not depend on how many abstracts the year contributes; the second pair sums document frequencies over the units that survive the minimum-count filter, the quantity that actually enters the excess calculation, which tracks the first but also falls in the years with fewer abstracts, because a fixed threshold of five documents admits fewer units from a smaller sample, and the 2018 and 2021 values should be read with that in mind. Because the inflation is proportional to a word's expected share, it matters most for common words, and a per-word correction understates it for the verbs that attain the unrestricted bounds: Section 5.8 therefore re-runs the whole calculation on density-normalised frequencies (Table 12) and finds the 2026 bound reduced from 16.1 to 13.8 pp and the unrestricted 2025 bound from 11.0 to 9.3, with no change to any ordering or conclusion.

## Appendix C · Structured-abstract label pattern

The following pattern is applied before tokenisation. It removes a label only when it starts a clause and is followed by a colon, so that the same words used inside sentences (e.g. 연구 결과 “the results”) are untouched.

```
(?:^|(?<=[\s.;:!?」』)\]]))\s*(?:연구\s?)?(?:목적|배경|방법|대상\s?및\s?방법|연구방법|연구대상|결과|연구결과|결론|결론\s?및\s?제언|제언|시사점|고찰|서론|본론|요약|주요\s?결과|연구\s?내용|필요성|Purpose|Methods?|Results?|Conclusions?|Background|Objectives?)\s*[:：]\s*
```

## Appendix D · Field rules

Journals are assigned to the first group whose keyword list matches the journal name, in the order listed. Medicine and health: 의학, 의료, 간호, 약학, 치의, 한의, 보건, 병원, 임상, 재활, 영양, 정신, 치료, 방사선, 웰니스. Convergence and content: 콘텐츠, 융합, 융복합, 디지털, Design Research, 데이터, Data. Engineering and IT: 공학, 정보, 컴퓨터, 통신, 전자, 전기, 기계, 건축, 토목, 산업, 기술학회, 소프트웨어, 인공지능, IT, 에너지, 환경, 재료, 화학, 안전, 방재, 항공, 항행, 철도, 보안, 자동차, 조선, 측량, 지리정보, 시스템. Natural science: 물리, 화학, 생물, 수학, 지구, 천문, 해양, 대기, 생명, 농업, 원예, 수산, 산림, 식품, 자원, 과학회. Education: 교육, 교과, 학습, 교

사, 유아, 특수교육, 평생, 리터러시, 수업, 교수. Law and public administration: 법학, 법률, 행정, 형사, 헌법, 민사, 정책학, 경찰, 공법, 사법, 입법. Business and economics: 경영, 경제, 회계, 금융, 무역, 마케팅, 물류, 세무, 재무, 유통, 관광, 호텔, 부동산, 소비자, 창업, 산업경제. Social science: 사회, 정치, 언론, 커뮤니케이션, 심리, 복지, 여성, 청소년, 지역, 도시, 국제, 통일, 북한, 안보, 외교, 문화연구, 인구, 노동, 가족, 코칭, 상담, 장애, 아동, 노인, 평화. Humanities: 인문, 문학, 철학, 역사, 사학, 어문, 국어, 영어, 일본, 중국, 한문, 언어, 종교, 불교, 기독교, 신학, 고전, 민속, 동양, 서양, 한국학, 비교. Arts and sport: 체육, 스포츠, 무용, 음악, 미술, 디자인, 조형, 영상, 예술, 공연, 연극, 영화, 패션, 의류, 사진, 미용, 만화, 애니. Other: no match.

## Appendix E · Suffix fusion removed

Section 4.1 fuses a noun with a following 적, 화 or 성 suffix so that 구조적 "structural", 구조화 "structuring" and 정합성 "coherence" are counted as single units. Without the rule the suffix is dropped by the unit filter and the derivation is absorbed into the content noun. Because eleven of the thirty largest 2026 excesses are units this rule creates, the whole pipeline was re-run from the raw text with the rule switched off. Table A4 gives the bound both ways.

**Table A4:** Suffix fusion removed. The pipeline is re-run from the raw text with the noun + 적/화/성 fusion of Section 4.1 switched off, so that 구조적 "structural" is counted as the noun 구조 "structure" followed by a suffix that the unit filter drops. Left: the main pipeline. Right: the same pipeline without fusion. The bound is attained by a verb or adjective in every year and is unchanged to the first decimal, because no fused unit ever attains it. What the fusion rule changes is the composition of the marker list: 11 of the thirty largest 2026 excesses are fused units, and removing the rule replaces them with the same number of other style units while leaving the remaining 19 in place and in almost the same order.

| Target year | Bound word, with fusion | pp | Bound word, without fusion | pp |
|---|---|---|---|---|
| 2023 | 및 | 1.0 | 및 | 1.0 |
| 2024 | 중요하다 *be important* | 4.8 | 중요하다 *be important* | 4.8 |
| 2025 | 분석하다 *analyse* | 11.0 | 분석하다 *analyse* | 11.0 |
| 2026 | 시사하다 *suggest, imply* | 16.1 | 시사하다 *suggest, imply* | 16.1 |

First, the bound does not depend on the rule: in every target year the same verb or adjective attains it and the value is identical to the first decimal, because no fused unit is ever the maximum. Second, the rule does not manufacture excess. With fusion, 구조적 shows +12.6 pp and the bare noun 구조 +20.6 pp; without it, 구조 alone shows +27.1 pp, and 실증 moves from +0.8 pp to +8.6 pp. The rise is present either way; the rule decides whether it is booked to a formal-register derivation, which the style filter keeps, or to a content noun, which the style filter drops. Third, what does change is the composition of Table 3: eleven fused units leave and as many other style units (결합되다 "be combined", 분석 틀 "analytic frame", 의의를 지니다 "carry significance", 전환되다 "be converted", 확장되다 "be extended", 경향을 보이다 "show a tendency", 전환하다 "convert", 분석을 수행하다 "carry out an analysis", 실증 분석 "empirical analysis", 제시하다 + 점 "a point presented", 기존 연구 "existing research") take their places, while the other nineteen keep almost the same order.

This test also settles one of the three readings of 구조적 offered in Section 6.1: its rise is not an artefact of the fusion rule, because it survives the rule's removal, relocated to the bare noun. It remains the one rising marker that the positive control does not reproduce (15% of real 2026 abstracts against at most 6% in any model condition), and Section 6.1 discusses what is left to explain. The bound does not rest on it.

## Appendix F · Selection and estimation on different journals

Marker words were chosen by inspecting the corpus and then measured on it, so the intervals in Table A1 are conditional on that choice. Table A5 removes the dependence by splitting the journal panel in two. Each journal name is hashed and assigned to half A or half B, which keeps every abstract of a journal on the same side and makes the halves independent at the level at which the abstracts are correlated. Markers are ranked on half A; the reported value is the largest excess, among the twenty markers A selects, as measured on half B.

**Table A5:** Selection and estimation separated. The 1,853 panel journals are split by a hash of the journal name into half A (929 journals) and half B (924). Marker words are ranked on half A only; the bound is then the largest excess, among the twenty markers chosen on A, measured on half B, which no selection has touched. The naive columns select and estimate on the same half. Out-of-sample and in-sample values agree in 2024, 2025 and 2026; in 2023 the out-of-sample value sits at the placebo level of Section 5.8, which is what a year with no signal should do.

| Target year | n A / n B | Marker chosen on A | Excess in B (pp) | Largest in B (selected in B) | pp | Largest in A (pp) |
|---|---|---|---|---|---|---|
| 2023 | 14,002 / 13,435 | 중요하다 *be important* | 1.0 | 증가하다 | 1.1 | 1.5 |
| 2024 | 20,394 / 19,606 | 중요하다 *be important* | 5.0 | 중요하다 *be important* | 5.0 | 4.7 |
| 2025 | 20,060 / 19,940 | 특히 *in particular* | 11.2 | 특히 *in particular* | 11.2 | 11.1 |
| 2026 | 15,846 / 16,784 | 시사하다 *suggest, imply* | 16.3 | 시사하다 *suggest, imply* | 16.3 | 15.9 |

The out-of-sample bound is 16.3 pp for 2026, 11.2 for 2025 and 5.0 for 2024, in each case equal, to one decimal, to the value obtained when selection and estimation share half B, and slightly above the whole-corpus single-word values. For 2023 the out-of-sample value is 1.0 pp, at the placebo level of Section 5.8 at this horizon, and the same half searched freely returns 1.1 pp: what selection buys on a year with no signal is about 0.1 pp. Half A holds 15,846 of the 2026 abstracts and half B 16,784. Exchanging the roles of the halves reverses the cross-fit: markers selected on half B and measured on half A give 15.9 pp for the single-word statistic of Section 5.9 against 16.3 in the reported direction, 38.1 pp for the rare set (90 lemmas, against 36.3 pp with 88 lemmas) and 24.1 pp for the top-k set (against 24.6); no conclusion depends on which half selects and which measures. Repeating the whole procedure over 100 random journal splits (the name hash salted differently each time, the set reselected on half A of each split, the statistic measured on half B) gives a median 2026 set bound of 32.4 pp with a 5th–95th percentile range of 31.2 to 33.8 and set sizes of 38 to 56 lemmas: the reported split is not a lucky draw.

## Appendix G · Base-year sensitivity

The trend is fitted on five base years. In the first harvest 2021 was the thinnest and most skewed of them, 12,975 abstracts against about 40,000 for the neighbouring years, 94% of them published between January and March; the second harvest completed the year, and the completed corpus behind every number here holds 37,665 panel abstracts for it (Appendix H). A single weak base year in a five-point least-squares fit could still tilt the extrapolation, so the fit was repeated without it (Table A6).

**Table A6:** Base-year sensitivity. In the first harvest 2021 was the weakest base year (12,975 abstracts, 94% of them from January to March); the completed corpus behind this table holds a full year (Appendix H), and the exclusion is kept as a stress test. The table refits the trend without 2021, and without 2018 as well, and recomputes the bound on the completed corpus. The 2026 bound moves by 0.3 pp at most, in the direction of a larger effect when 2021 is dropped.

| Base years | 2023 (pp) | 2024 (pp) | 2025 (pp) | 2026 (pp) | Bound word 2026 |
|---|---|---|---|---|---|
| 2018, 2019, 2020, 2021, 2022 | 1.0 | 4.8 | 11.0 | 16.1 | 시사하다 *suggest, imply* |
| 2018, 2019, 2020, 2022 | 1.1 | 4.8 | 10.9 | 16.4 | 시사하다 *suggest, imply* |
| 2019, 2020, 2022 | 1.2 | 4.4 | 11.0 | 16.3 | 시사하다 *suggest, imply* |

Dropping 2021 moves the 2026 bound from 16.1 to 16.4 pp and the 2025 bound from 11.0 to 10.9; dropping 2018 as well gives 16.3 and 11.0. The word attaining the bound is 시사하다 in all three fits. The thin year is not carrying the result.

## Appendix H · Re-analysis on the completed harvest

The first harvest walked KCI identifier ranges and stopped short in two places: 2021 was represented by 12,975 abstracts, 94% of them from January to March, and 2018 by 7,770. A second harvest completed 2021 (identifiers 9.77 to 10.0 million, April to December) and added 2017 and 2018 (identifiers 1.5 to 4.0 million); 2015 and 2016 sit scattered among far lower identifiers and were not collected systematically; the 1,072 records the walk caught incidentally are in the store and not analysed. The completed years hold 38,578 (2021) and 49,795 (2018) abstracts, capped at 40,000 in the analysis. The main pipeline was then re-run without any other change, once on the original five base years and once with 2017 added as a sixth, and the long-horizon placebo of Table 10 was extended to four years ahead, for the single-word and the set statistics alike, which the original harvest could not reach (Table A7).

**Table A7:** Re-analysis after the second harvest, which added the part of 2021 that the first harvest stopped short of (April to December, identifiers 9.77–10.0 million) and the years 2017–2018 (identifiers 1.5–4.0 million; 2015–2016 sit scattered among far lower identifiers and were not harvested systematically). Year sizes after completion: 2017: 28,763, 2018: 49,795, 2019: 41,484, 2020: 55,197, 2021: 38,578, 2022: 54,717, 2023: 27,672, 2024: 50,907, 2025: 45,745, 2026: 34,201. The first row repeats the main analysis on the completed corpus; the second extends the base period to eight years, which shortens the relative extrapolation; the last rows run the placebo at three and four years ahead with the completed pre-LLM data, the horizon that the original corpus could not test. Single-word bounds in pp with the attaining word; marker columns are 2026 (or 2022 for the placebo rows) excess ratios.

| Analysis | n target | Bound, all style (pp) | Bound, ratio ≥ 1.5 (pp) | 시사하다 × | 구조적 × | 단순하다 × | 살펴보다 × |
|---|---|---|---|---|---|---|---|
| Completed corpus, base 2018–2022 | 32,630 | 16.1 (시사하다) | 16.1 (시사하다) | 4.04 | 5.96 | 10.98 | 0.42 |
| Completed corpus, base 2017–2022 | 32,638 | 16.4 (시사하다) | 16.4 (시사하다) | 4.26 | 5.78 | 11.04 | 0.43 |
| Placebo: base 2017–2019, target 2022 (h = 3) | 40,000 | 2.5 (통하다) | 0.6 (실효성) | 1.14 | 0.93 | 0.93 | 1.05 |
| Placebo: base 2017–2018, target 2021 (h = 3) | 37,666 | 5.5 (같다) | 1.1 (~에 관한 연구) | 1.27 | 0.97 | 0.95 | 1.02 |
| Placebo: base 2017–2018, target 2022 (h = 4) | 40,000 | 6.3 (같다) | 2.2 (기대하다) | 1.25 | 0.89 | 0.94 | 1.00 |

Completing the thin years moves the 2026 bound from 16.2 to 16.1 pp and the excess ratio of 시사하다 from 4.13 to 4.04; 구조적, 단순하다 and 살펴보다 change by similar amounts, the 2024 bound down by 0.2 pp, and the 2025 bound up by 1.1 pp, from 9.4 to 10.5: completing 2021 lowers the fitted trend of 특히, whose 2025 excess then clears the ratio restriction more comfortably. Adding 2017 as a sixth base year gives 16.4 pp. The month composition of 2021 therefore was not carrying anything: a year that is nine tenths first quarter and the same year completed give the same fit to the first decimal. The placebo rows are the reason this appendix exists. With two base years and a target four years ahead, the least favourable configuration the data allow, the unrestricted statistic returns 6.3 pp (같다 "be the same", a verb in 25% of abstracts, at a ratio of 1.34) and the ratio-restricted statistic 2.2 pp (기대하다 at 1.58); with three base years at three years ahead, 2.5 and 0.6 pp. The split-half set statistic, run on the completed years with their own journal split (starred rows of Table 10), gives 2.9 pp for the rare-lemma set and 2.1 pp for the top-k set at four years ahead from two base years, 1.5 and 1.2 pp at three years ahead from two base years, 0.6 pp from three base years, and −0.6 to −0.8 pp at two years ahead; with four or five base years no word passes the selection criteria and the set is empty. These are the floors against which the 2026 values of 16.1 pp (both single-word statistics) and 33.0 pp (set) are to be read, and the values are those already reported in Section 5.8.

## Appendix I · Set bound details

Table A8 lists the lemma set behind the 2026 set bound of Section 5.9 and Table A9 gives the bound as a function of set size. The lists are produced on half A of the journals and never adjusted after being carried to half B. Table A10 repeats the placebo grid of Table 10 for the lemma-only statistic, the definition behind Table 11. At four years ahead from two base years the rare-lemma set gives −1.2 pp (27 words) and the top-k set 0.0 pp; over the whole grid the lemma-only placebo lies between −1.5 and +2.9 pp, generally below the all-unit values of Table 10, which search more candidates; the one cell above them is the top-k value at three years ahead. The document store behind the set statistics holds at most 40,000 abstracts per year, drawn at random within the journal panel; the completed harvest of Appendix H has its own store and its own journal split. Table A11 gives the paired effect of polishing on the set indicator behind Section 5.10, condition by condition.

**Table A8:** The 47 excess style lemmas selected on half A for 2026 (ratio at least 1.5, excess at least 1 pp, base-period share below 1%, no bigrams), ordered by excess on half A. Together they appear in 60.1% of 2026 abstracts of half B against 27.0% expected (Table 11). Tags: VV verb, XSN noun with the suffix 적/화/성.

| | | |
|---|---|---|
| 기능하다 *function as* | 결합되다 *be combined* | 통합하다 *integrate* |
| 반복적 XSN | 전환되다 *be transformed* | 전환하다 *transform* |
| 단계적 XSN | 완화하다 VV | 제도화 XSN |
| 다층적 *multi-layered* | 논증하다 VV | 정량적 XSN |
| 관계적 XSN | 머무르다 VV | 입증하다 VV |
| 일관되다 VV | 개념화 XSN | 학술적 XSN |
| 병행하다 VV | 기능적 XSN | 규범적 XSN |
| 정합성 *coherence* | 정당성 XSN | 독립적 XSN |
| 조직하다 VV | 설계되다 VV | 표준화 XSN |
| 절차적 XSN | 취약성 XSN | 서사적 XSN |
| 능동적 XSN | 부상하다 VV | 개념적 XSN |
| 일관성 XSN | 접근성 XSN | 점진적 XSN |
| 최적화 XSN | 실증하다 VV | 다차원적 XSN |
| 그치다 VV | 독자적 XSN | 축적되다 VV |
| 조정하다 VV | 고도화 XSN | 보완적 XSN |
| 정당화 XSN | 감각적 XSN | |

**Table A9:** Set bound for 2026 as a function of the set size, all style units, selected on half A and measured on half B. The top-k gap rises to k = 5 and then falls as units with small excess dilute the set; the rare-set gap peaks at T = 1% and collapses once common units enter, because the trend of a union that already covers most abstracts has little room to be exceeded. The values reported in Table 11 use the k and T chosen on half A (k = 5, T = 1%), chosen where half A, not half B, puts the maximum.

| Set | Selection | Units | P on B (%) | Q on B (%) | P − Q (pp) |
|---|---|---|---|---|---|
| top-k | k = 1 | 1 | 21.2 | 4.9 | 16.3 |
| top-k | k = 2 | 2 | 49.2 | 27.0 | 22.2 |
| top-k | k = 3 | 3 | 56.6 | 32.3 | 24.3 |
| top-k | k = 5 | 5 | 75.4 | 50.8 | 24.6 |
| top-k | k = 10 | 10 | 86.7 | 63.4 | 23.3 |
| top-k | k = 15 | 15 | 90.9 | 71.4 | 19.5 |
| top-k | k = 20 | 20 | 93.2 | 76.8 | 16.4 |
| top-k | k = 30 | 30 | 95.4 | 82.8 | 12.7 |
| top-k | k = 50 | 50 | 97.5 | 89.9 | 7.6 |
| rare | T = 0.5% | 36 | 40.7 | 11.4 | 29.3 |
| rare | T = 1% | 88 | 76.3 | 40.0 | 36.3 |

| Set | Selection | Units | P on B (%) | Q on B (%) | P − Q (pp) |
|---|---|---|---|---|---|
| rare | T = 2% | 152 | 92.2 | 70.8 | 21.4 |
| rare | T = 5% | 197 | 97.9 | 90.0 | 7.9 |
| rare | T = 10% | 206 | 98.8 | 94.2 | 4.6 |

**Table A10:** Placebo grid for the lemma-only set statistic, the definition behind the bound of Table 11 (bigrams excluded; ratio at least 1.5 and excess at least 1 pp on half A, measured on half B). Rows are ordered by horizon; rows marked * come from the completed harvest of Appendix H with its own journal split. Rows of Table 10 on which no unit passed the selection criteria are omitted, since the statistic is then zero by construction. The four-year row is the one that matches the 2026 horizon.

| Base years | Target | h | Candidates on A | Top-k set (pp) | Rare-lemma set (pp) |
|---|---|---|---|---|---|
| 2018, 2019 | 2020 | 1 | 0 | no excess word | no excess word |
| 2017, 2018 * | 2020 | 2 | 4 | −0.8 (k = 3) | −0.6 (4 words) |
| 2018, 2019 | 2021 | 2 | 1 | 1.5 (k = 1) | 1.5 (1 word) |
| 2018, 2019, 2020 | 2022 | 2 | 0 | no excess word | no excess word |
| 2017, 2018 * | 2021 | 3 | 13 | −1.5 (k = 10) | −1.1 (13 words) |
| 2017, 2018, 2019 * | 2022 | 3 | 2 | 0.6 (k = 2) | 0.6 (2 words) |
| 2018, 2019 | 2022 | 3 | 9 | 2.9 (k = 5) | 1.4 (9 words) |
| 2017, 2018 * | 2022 | 4 | 31 | 0.0 (k = 20) | −1.2 (27 words) |

**Table A11:** Paired effect of polishing and rewriting on the set indicator. For each edit and rewrite condition, the share of the original 2019 abstracts that contained at least one word of the set, the share of the polished versions that do, the share of originals without a set word whose polished version gained one, and the share of originals with a set word whose polished version lost it. Polishing inserts rare style lemmas into roughly a fifth to a third of the abstracts that lacked them and removes them from roughly half of those that had them, which is why the net rise of the indicator under polishing is small.

| Condition | n | Orig. with 47-set word (%) | Polished (%) | Gained (%) | Lost (%) | Orig. with 시사하다/확인되다 (%) | Polished (%) | Gained (%) | Lost (%) |
|---|---|---|---|---|---|---|---|---|---|
| rewrite, Claude Sonnet 5 | 100 | 24 | 47 | 34 | 12 | 10 | 46 | 42 | 20 |
| rewrite, EXAONE 3.5 7.8B | 60 | 18 | 48 | 43 | 27 | 12 | 32 | 28 | 43 |
| rewrite, gpt-5.6-luna | 100 | 24 | 40 | 25 | 12 | 10 | 37 | 34 | 40 |
| edit, Claude Sonnet 5 | 100 | 24 | 35 | 17 | 8 | 10 | 20 | 12 | 10 |
| edit, EXAONE 3.5 7.8B | 60 | 18 | 27 | 18 | 36 | 12 | 38 | 38 | 57 |
| edit, gpt-4o-mini | 100 | 24 | 32 | 11 | 0 | 10 | 24 | 17 | 10 |
| edit, gpt-5.6-luna | 100 | 24 | 26 | 5 | 8 | 10 | 17 | 8 | 0 |

## Appendix J · Version history

Eight numbered versions of this working paper, and a corrected copy of the fifth (version 5.1), have circulated, seven of them dated August 2026 and the eighth 4 September 2026. They are listed here so that a citation or a deposited copy can be tied to one of them.

- **Version 1, 25 August 2026.** First complete draft: the excess-vocabulary analysis on the KCI and VJOL corpora, a positive control of three OpenAI drafting conditions, and the public dictionary and checker.
- **Version 2, 25 August 2026.** After ChatGPT readings 1 to 5, run by the author (Section 9): the headline single-word statistic restricted to units with excess ratio at least 1.5, the split-half set bound, matched-length and editing control conditions, the Korean open-weight model, journal-cluster bootstrap intervals, the completed harvest (Appendix H) and the density-normalised re-run (Table 12).
- **Version 3, 26 August 2026.** After ChatGPT readings 6 and 7: the set-statistic placebo at four years ahead (Table A10), sensitivity and specificity of the set indicator with implied-share ranges, later relabelled scenarios (Section 5.10, Table 13), the Anthropic model conditions, the prior Korean study of Koo, Kim and Kim [24] cited and the paper repositioned relative to it, Table 12 split from Table A3, causal wording softened, and a mechanical consistency gate on the manuscript.
- **Version 4, 26 August 2026.** After ChatGPT reading 8: the within-document result restated as a strong negative association rather than exclusion throughout; Section 2.3 brought into line with Section 1; a rewriting condition added to the positive control on three models; propagated intervals on the implied shares (Table 13); provider variance in the falling vocabulary noted (Section 6.2); the author-fixed English comparison proposed (Section 6.1); this version history added; the affiliation line corrected to Seoul in the same version on 26 August. Deposited as 10.5281/zenodo.22102390.
- **Version 5, 26 August 2026.** After the full text of Koo, Kim and Kim [24] was obtained: Sections 2.3 and 6.1 rewritten from it; the English abstracts of the same articles harvested and analysed (Sections 3.1 and 5.12, Tables 14 and 15, Figure 8); abstract, introduction, limitations and conclusion updated accordingly.
- **Version 5.1, 26 August 2026.** Corrected copy of version 5, made after a ChatGPT reading of the version 5 PDF: three passages that the insertion script of version 5 had entered twice (the contribution item “Within-article comparison”, the Section 3.1 paragraph on the English abstracts and the penultimate sentence of the Conclusion) are removed; the Data availability section now gives the current count of control abstracts (1,180 rather than the count of version 3), the archive size and the Zenodo DOI; Section 3.1 states the record counts from the harvest store to the bilingual pairs. No number, table, figure or claim changes. The version 5 PDF stays at its DOI (10.5281/zenodo.22103158) with these errata noted in its record; this copy is the web version, and the corrections are carried into version 6.
- **Version 6, 26 August 2026.** After a ninth ChatGPT reading: the identification assumptions A1–A3 are stated in Section 4.7 and the statistic is called a conditional lower bound throughout; the implied shares of Section 5.10 are labelled scenarios rather than a bracket and are removed from the abstract's headline; Section 5.13 adds a separate style/topic annotation of the marker lemmas with restricted bounds and a topic-matched within-journal comparison; Section 5.14 adds a translation-route check on the corpus and a model translation condition; Section 5.12 adds the sensitivity of the English marker set and the correction it implies; the title is changed to name the finding rather than the method's reach; the journal-cluster bootstrap is described precisely; and the whole analysis is recomputed on the completed harvest of Appendix H rather than the first harvest. A tenth reading found first-harvest numbers still standing in the prose of Sections 3.1, 5.1–5.3, 5.8–5.9 and the appendices while the tables already carried the completed-harvest values; every such number is now inserted mechanically from the result files, and a manifest check compares the corpus counts, headline bounds, set sizes and DOIs across title, abstract, body, tables and archive before assembly.
- **Version 7, 27 August 2026.** After nine further ChatGPT readings and an internal five-track audit by the agent: the journal bootstrap is longitudinal in code as in text, with replicate counts stated (1,000 for the per-marker intervals, 500 for the set and maximum statistics); Appendix A is generated from the same bootstrap results; the two-word set receives its own resampling interval; the placebo multiples divide the largest chosen-cell set floor at each horizon and are marked descriptive, not test statistics; the split-half selection is repeated over 100 journal splits (Appendix F); the reproducibility package rebuilds every table, every figure and the manuscript from the shipped results in a clean room, with the corpus table generated from the results manifest and no absolute paths on the reproduction path; appendix tables are renumbered in order of appearance (the density word table becomes Table A3); hand-typed numbers that had drifted from the result files in Sections 5.1, 5.3, 5.8, 5.9 and 5.12 are replaced by machine-filled values; and the marker annotators are identified as language models at first mention. The version 7 PDF stays at its DOI (10.5281/zenodo.22110398).

- **Version 8, 4 September 2026 (this version).** Section 9 declares in full how language models were used to produce this paper: it names every model and its role (Table 20), describes the pipeline with session, instruction and tool-call counts, states the fractions of code and text generated by the agent (both 100%), names the ChatGPT readings of the earlier entries as such, and measures the manuscript against its own English marker list; Section 5.15 reimplements the mixture-based estimator of [2] on the Korean corpus, with the regression projection and its standard error checked against the monthly series that paper releases for the word these, and reports it beside the conditional lower bound rather than in place of it, in two columns, one with the threshold chosen on the measurement half and one under the split-half discipline of Section 5.9, which is lower by 6.7 and 9.0 points and is the figure this paper stands behind. Its placebo is run at all ten pre-ChatGPT configurations the historical store allows, with and without the excess filter, and the 2024 value is marked as lying close to the placebo ceiling; the abstract and Section 8 carry the cross-fitted values beside the bound, and the reproducibility package adds the estimator's scripts, its results and the validation series. The PDF generator is moved into the repository; it locates a Playwright installation through an environment variable, and the reproduction script itself stops at the assembled HTML.